\documentclass{article}

\usepackage{microtype}
\usepackage{graphicx}
\usepackage{subcaption}
\usepackage{booktabs} % for professional tables
\usepackage{hyperref}
\usepackage{fontawesome5}
\usepackage{marvosym}

\usepackage[accepted]{icml2026}

\usepackage{amsmath}
\usepackage{amssymb}
\usepackage{mathtools}
\usepackage{amsthm}
\usepackage{amsfonts}       % blackboard math symbols
\usepackage{xcolor}         % colors
\usepackage{color}
\usepackage{lipsum}
\usepackage{multirow,rotating}
\usepackage{enumitem}
\usepackage{xspace}
\usepackage{quoting}
\usepackage{csquotes}
\usepackage[most]{tcolorbox}
\usepackage{colortbl}

\usepackage{mdframed}
\newmdenv[
  leftline=false,
  rightline=false,
  topline=false,
  bottomline=false,
  linecolor=gray,
  linewidth=2pt,
  innerrightmargin=5pt,
  innerleftmargin=2pt,
  innertopmargin=5pt,
  innerbottommargin=0pt,
  leftmargin=4pt,
  rightmargin=5pt,
  skipabove=\parskip,
  skipbelow=0
]{myquote}

\usepackage[capitalize,noabbrev]{cleveref}

\theoremstyle{plain}

\theoremstyle{definition}

\theoremstyle{remark}

\usepackage[disable,textsize=tiny]{todonotes}
\icmltitlerunning{How Far Ahead Do LLMs Plan? Uncovering the Latent Horizon in Chain-of-Thought Reasoning}

\begin{document}

\twocolumn[
  \icmltitle{How Far Ahead Do LLMs Plan?\texorpdfstring{\\}{ }Uncovering the Latent Horizon in Chain-of-Thought Reasoning}

  % It is OKAY to include author information, even for blind submissions: the
  % style file will automatically remove it for you unless you've provided
  % the [accepted] option to the icml2026 package.

  % List of affiliations: The first argument should be a (short) identifier you
  % will use later to specify author affiliations Academic affiliations
  % should list Department, University, City, Region, Country Industry
  % affiliations should list Company, City, Region, Country

  % You can specify symbols, otherwise they are numbered in order. Ideally, you
  % should not use this facility. Affiliations will be numbered in order of
  % appearance and this is the preferred way.
  \icmlsetsymbol{equal}{*}

  \begin{icmlauthorlist}
    \icmlauthor{Liyan Xu}{wechat}
    \icmlauthor{Mo Yu}{wechat}
    \icmlauthor{Fandong Meng}{wechat}
    \icmlauthor{Jie Zhou}{wechat}
  \end{icmlauthorlist}

  \icmlaffiliation{wechat}{WeChat AI, Tencent Inc}

  \icmlcorrespondingauthor{Liyan Xu}{liyanlxu@tencent.com}
  \icmlcorrespondingauthor{Jie Zhou}{withtomzhou@tencent.com}

  % You may provide any keywords that you find helpful for describing your
  % paper; these are used to populate the "keywords" metadata in the PDF but
  % will not be shown in the document
  \icmlkeywords{Large Language Models, LLM, Chain-of-Thought, CoT, Probing, Planning, Uncertainty Estimation}

  \vskip 0.3in
]

% this must go after the closing bracket ] following \twocolumn[ ...

% This command actually creates the footnote in the first column listing the
% affiliations and the copyright notice. The command takes one argument, which
% is text to display at the start of the footnote. The \icmlEqualContribution
% command is standard text for equal contribution. Remove it (just {}) if you
% do not need this facility.

% Use ONE of the following lines. DO NOT remove the command.
% If you have no special notice, KEEP empty braces:
\printAffiliationsAndNotice{}  % no special notice (required even if empty)
% Or, if applicable, use the standard equal contribution text:
% \printAffiliationsAndNotice{\icmlEqualContribution}

\newcommand\probe{Tele-Lens\xspace}

\begin{abstract}
Chain-of-thought (CoT) reasoning has become a central mechanism for eliciting multi-step reasoning in Large Language Models (LLMs). Yet recent evidence presents a tension: hidden states appear to already encode future reasoning before CoT fully unfolds, while explicit steps still remain crucial for tasks requiring compositional computation.
To deepen the understanding between LLM's internal states and its verbalized reasoning trajectories, we investigate the latent planning strength of LLMs, through our probing method, Tele-Lens, applying to hidden states across diverse task domains.
Our empirical results indicate that LLMs exhibit a \emph{myopic} horizon, primarily conducting incremental transitions without precise global planning.
Leveraging this characteristic, we propose a hypothesis on enhancing uncertainty estimation of CoT, which we validate that a sparse set of \emph{pivot} positions can effectively represent the uncertainty of the entire path.
We further underscore the significance of exploiting CoT dynamics, and demonstrate that automatic recognition of CoT bypass can be achieved without performance degradation.
Our code, data and models are released at \href{https://github.com/lxucs/tele-lens}{https://github.com/lxucs/tele-lens}.
\end{abstract}

\section{Introduction}
\label{sec:intro}

Chain-of-Thought (CoT) \cite{cot21,cot} has fundamentally reshaped problem-solving in natural language processing, marking a shift from traditional pattern-matching approaches, e.g. encoder-based classification \cite{bert,roberta}, toward prompt-based reasoning articulated explicitly in natural language \cite{zhou2023large,dong-etal-2024-survey,sahoo2025systematic}.
The capacity of CoT is further amplified through extensive thinking emanated from reinforcement learning, characterized by recent models such as DeepSeek-R1 \cite{r1}.

While CoT is widely perceived as the de facto reasoning paradigm, however, recent studies on Large Language Models (LLMs) have revealed complementary, and at times seemingly conflicting perspectives.
On the one hand, LLMs have been shown to exhibit \textbf{internal planning on the reasoning trace prior to the explicit emergence of CoT}.
\cite{pal-etal-2023-future} reveals that early hidden states already encode information about subsequent generations; \cite{sheng2026on} finds that reasoning capability is largely pre-planned before generation begins; \cite{dong2025emergent} observes that early hidden states can reliably predict reasoning steps and key attributes with high correlation. Similarly, numerous studies have shown that the initial stages of CoT can effectively anticipate the final answers \cite{azaria-mitchell-2023-internal,gottesman-geva-2024-estimating,afzal-etal-2025-knowing}.

The internal planning of LLMs appear to diminish the significance of CoT generation, raising the question of whether the thinking process is just echoing pre-determined paths already encoded in prior internal states. On the other hand, theoretical analyses advocate that \textbf{CoT is indispensable due to the limited expressivity of Transformers} bounded by its architectures \cite{bhattamishra-etal-2023-simplicity,merrill-sabharwal-2023-parallelism,li2024chain}, and only intermediate steps of CoT can derive length generalization \cite{anil2022exploring,xiao2025generalizing} and compositional reasoning \cite{wies2023subtask,abbe2024how,zubic2025limits}. Therefore, the  manifestation of pre-calculated trajectories appear \emph{unlikely} via internal planning before the onset of CoT.

Nonetheless, the relationship between the model’s internal representations and its verbalized reasoning tokens largely remains opaque. In this work, we investigate the \textbf{\emph{internal dynamics of CoT}}, and target the following questions concerning the latent planning horizon:
\begin{myquote}
\begin{enumerate}[label=Q\arabic*.,noitemsep,nolistsep,leftmargin=*]
\item \emph{To what extent do hidden states encode a global plan for the reasoning roadmap, as opposed to supporting rather local, incremental state transitions?}
\item \emph{How does the planning horizon further influence other characteristics of CoT reasoning?}
\end{enumerate}
\end{myquote}

We believe answering these questions is important not only for deepening our understanding of CoT dynamics, but also for providing critical insights into broader model-thinking design scenarios. For example, recent works leverage LLM internal states to enable early exiting from CoT reasoning, thereby mitigating the \emph{overthinking} problem \cite{yong2026think,yang2026dynamic}. Recently, both GPT-5\footnote{\href{https://openai.com/index/gpt-5-system-card}{https://openai.com/index/gpt-5-system-card}} and Claude Code\footnote{\href{https://platform.claude.com/docs/en/build-with-claude/adaptive-thinking}{https://platform.claude.com/docs/en/build-with-claude/adaptive-thinking}} introduce forms of \emph{adaptive thinking} that route user requests to models of different sizes or allocate varying levels of reasoning effort. Such mechanisms fundamentally rely on how strong the model can ``sense'' or ``see through'' the input complexity through internal planning.

Towards this objective, we derive empirical insights by examining the synergy between explicit CoT steps and its latent planning horizon. Building on the observations, we then highlight the significance of leveraging CoT dynamics on estimating CoT's uncertainty and necessity.

To answer the first question, \cref{sec:horizon} presents a series of probing experiments designed for dissecting LLM hidden states, aiming to evaluate the internal planning strengths with respect to future reasoning trajectories.
We first introduce our probing method, termed \textbf{\probe}, which employs a trained low-rank adapter \cite{bottleneck-adapter} that transforms each hidden state within CoT steps to predict \underline{Tele}ological information along multiple dimensions, including subsequent tokens, final answers, reasoning lengths, etc. 
Importantly, unlike prior works that primarily address single-domain tasks, we conduct probing experiments across 12 diverse datasets spanning different classes and domains, ranging from straightforward knowledge question answering to classic hard problems, e.g. Parity (counting the number of digits as even or odd), a canonical challenge for Transformers \cite{chiang-cholak-2022-overcoming,hahn-rofin-2024-sensitive}.

By empirical results, we observe sharply contrasting behaviors across probing dimensions and task domains, as detailed in \cref{ssec:results}. For instance, while early hidden states may encode predictive signals of the final answer for relatively simple tasks, they can behave nearly randomly on compositional tasks, where they only begin to reliably capture the precise answer within the final one or two reasoning steps before CoT completion.

Overall, our probing results bring a unified view of the prior complementary beliefs from previous works: LLMs exhibit a \textbf{\emph{myopic planning horizon}}, in which hidden states primarily support immediate, local transitions rather than long-range, global trajectories. However, for simpler tasks that fall within the LLM's single-step pattern-matching capacity, early hidden states may capture a coarse gist of the final answer—albeit in a more heuristic manner, rather than through a precise, pre-planned reasoning process.

Addressing Q2, leveraging latent CoT signals, we first focus on \textbf{uncertainty calibration} over CoT, where an effective confidence metric, e.g. the rollout perplexity or entropy, should assign high scores to correct reasoning paths and low scores to uncertain ones \cite{huang2024surveyuncertaintyestimationllms,chen2024inside,bakman-etal-2025-reconsidering}.
We propose a hypothesis followed by empirical validation: the uncertainty of CoT follows a \textbf{Wooden Barrel principle}. Just as the capacity of a barrel is determined not by its average stave height but by its shortest stave, the reliability of a reasoning chain is governed by a small number of \emph{pivot} positions. Intuitively, as the model's latent planning is \emph{myopic}, most CoT tokens are high-confident local transitions that may dilute the underlying uncertainty of the entire trace.
We therefore speculate that focusing on a small set of \emph{pivot} positions instead of global aggregates could enable more precise uncertainty estimation.
Empirical results in \cref{ssec:uncertainty} find that even a simple strategy of top-$k$ selection can effectively enhance the accuracy of estimation across all three general uncertainty metrics, yielding up to 6\% absolute improvement.

Beyond uncertainty estimation, we also present a \textsl{proof-of-concept} that the CoT planning patterns can be leveraged to recognize whether CoT is necessary to derive the final answer, achieving automatic \textbf{CoT bypass} that directly outputs the answer with minimal performance degradation. Experiments in \cref{ssec:necessity} demonstrate that our proposed strategy using Qwen3-32B can realize up to 16.2\% CoT bypass with only a negligible 0.03 overall accuracy drop.

Back to Q2, our proposed strategies on CoT uncertainty and necessity estimation further underscore the significance of analyzing CoT dynamics, which encode hidden yet valuable information.
We hope that this work on uncovering the latent planning horizon could advance the understanding of CoT synergy, and spur more identification of hidden signals to be exploited in broader model-thinking scenarios.

\section{CoT Planning Horizon}
\label{sec:horizon}

This section delineates the detailed experimental setup and findings on diving into the latent planning capacity. \cref{ssec:probe} introduces our probing method, \probe, followed by a description of the data setup and model configurations. Empirical results are reported in \cref{ssec:results}.

\subsection{\probe}
\label{ssec:probe}

To enable probing across multiple dimensions, including subsequent token prediction, our method is designed to support prediction over the full LLM vocabulary. To this end, we adopt a transformation-based approach to probe various \underline{Tele}ological information upon the CoT trace, dubbed \probe.
It follows the paradigm of Logit Lens \cite{nostalgebraist2021logitlens,belrose2025elicitinglatentpredictionstransformers}, originally for examining layer-wise interpretability in Transformers, which bridges the hidden states from intermediate Transformers layers to the final LM head directly, thereby enabling whole-vocabulary prediction.
For \probe, to mitigate overfitting and computational overhead, we adopt a bottleneck low-rank adapter \cite{bottleneck-adapter} with added nonlinearity for hidden state transformation, more formally described as follows.

Concretely, for an LLM rollout, we denote the response tokens in its thinking process up to the final answer as $T = \{t_1, t_2, .., t_n\}$, representing a reasoning trajectory of length $n$ (throughout this paper, we use the terms ``thinking'' and ``CoT'' interchangeably).
The hidden state corresponding to token $t_i$ at the $k$-th Transformers layer is then denoted as $H_i^k \in \mathbb{R}^{d}$, with $d$ being the LLM hidden size.
The corresponding transformed hidden state $\widetilde{H}_i^k \in \mathbb{R}^{d}$ by applying the bottleneck adapter, and its predicted probability distribution $\mathcal{P}_i^k$ over the LLM vocabulary $\mathcal{V}$, are defined as:
\begin{align}
    \widetilde{H}_i^k = \operatorname{GeLU} \Bigl( \bigl( H_i^k + &\operatorname{Emb}^k(\delta) \bigl) \, A^k \Bigl) \, B^k \label{eq:adapter} \\ 
    \mathcal{P}_i^k(\mathcal{V} \mid t_i,A^k,B^k,\operatorname{Emb}^k, \delta) &= \operatorname{Softmax} \bigl( \widetilde{H}_i^k L \bigl) \label{eq:head}
\end{align}
where $A^k \in \mathbb{R}^{d \times r}$, $B^k \in \mathbb{R}^{r \times d}$ and $\operatorname{Emb}^k \in \mathbb{R}^{m \times d}$ are the learnable parameters of the adapter for the $k$-th Transformers layer, typically with a low rank $r < d$. Particularly, $\operatorname{Emb}^k$ is an \emph{optional} embedding matrix, taking an offset $\delta = 1,2,..,m$ to inject the target predicting position up to $m$ offset.
$L \in \mathbb{R}^{d \times |\mathcal{V}|}$ is the LM head matrix that will keep frozen during adapter training.

For each token $t_i$ in the reasoning path, we take its hidden state of each Transformers layer and probe along three teleological dimensions:
\begin{itemize}[noitemsep,nolistsep]
    \item \textbf{Subsequent tokens}: we use solely $H_i^k$ to predict its $m$ following tokens $\{t_{i+\delta} \mid \delta=1,2,..,m\}$. Each offset $\delta$ is injected respectively as in \cref{eq:adapter}.
    \item \textbf{Reasoning length}: we use $H_i^k$ to predict the total length of the thinking. Instead of applying LM head by \cref{eq:head}, we take $\widetilde{H}_i^k$ followed by a single regression layer to yield a number prediction.
    \item \textbf{Final answer}: we use $H_i^k$ to predict the final answer directly, with $\operatorname{Emb}^k$ removed in \cref{eq:adapter}. Each answer should be uniquely identifiable by a token in $\mathcal{V}$, thus this suits only for tasks with a fixed answer space.
\end{itemize}

If LLMs maintain a precise global plan early in the reasoning process, then the strength of that plan should, in principle, be measurable by these probing targets, before explicit CoT fully unfolds.
For final-answer probing particularly, as answer tokens constitute the natural continuation of the reasoning process, they can reflect how early, and to what extent, future outcome information is internally represented throughout CoT generation.

\subsection{Tasks and Datasets}
\label{ssec:tasks}

As previous works on CoT analysis mainly focus on specific domains of interest, the findings can be complementary that reflects different perspectives and angles, as discussed in \cref{sec:intro}.
Towards more comprehensive empirical insights, we broaden the scope of domains and include 12 diverse tasks, which we categorize into three types as below. Concrete examples of these tasks are provided in Appendix~\ref{app:task-example}.

\vspace{-0.4em}
\paragraph{Explicit Compositional Tasks}
\label{ssec:type1}

These tasks require explicit multi-step procedures to resolve, involving a high degree of structural modularity. Notably, as suggested by both prior empirical studies and theoretical analyses, Transformers often struggles to efficiently perform function composition within a single forward pass \cite{dziri2023faith,merrill-sabharwal-2023-parallelism,zubic2025limits}. Consequently, such tasks usually require intermediate CoT steps to derive the final answer. We include the following three tasks, for which data generation is fully controllable.

\begin{itemize}[noitemsep,nolistsep,leftmargin=1em]
\item \textbf{Parity}: a classic task often seen in Transformers' expressivity and learnability analysis \cite{chiang-cholak-2022-overcoming,bhattamishra-etal-2023-simplicity,hahn-rofin-2024-sensitive}. Given a sequence of digits, the task essentially asks whether the total count of a target digit is even or odd.
\item \textbf{Cycle}: we adopt a task introduced by \citet{abbe2024how}, in which the input consists of a list of directed edges, forming either a single full-sized cycle or two half-sized cycles. The task requires determining whether there exists a path between two specified vertices, or equivalently, whether they fall into the same cycle.
\item \textbf{Subsum}: an algorithmic task, Max Subsequence Sum, adopted in prior Transformers studies \cite{dziri2023faith}. For a list of $n$ numbers, the task computes the maximum sum of its subsequences, which admits an $O(n)$ dynamic programming solution. We query the least significant digit of the maximum sum for a fixed answer space.
\end{itemize}

\vspace{-0.5em}
\paragraph{Implicit Compositional Tasks}
\label{ssec:type2}

These tasks typically require multiple reasoning steps as well, but in a more nuanced and implicit manner embedded in the problem semantics, such as mathematical or logical reasoning.
For math-related tasks, we adopt three following datasets: \textbf{GSM8K} \cite{gsm8k}, \textbf{MATH} \cite{hendrycks2021measuring}, \textbf{AIME} \cite{aime}.
To enable a fixed answer space tailored for final-answer probing, we adapt each problem into a multi-choice format, by prompting GPT-4.1 to generate plausible yet misleading distractor options. Details on the multi-choice conversion are provided in Appendix~\ref{app:data-existing}.

For logical reasoning, we include the following two datasets that evaluate soft reasoning framed in natural language: \textbf{MuSR} \cite{sprague2024musr}, \textbf{Zebra} \cite{lin2025zebralogic}.

\vspace{-0.5em}
\paragraph{Knowledge and Semantic Tasks}
\label{ssec:type3}

These tasks primarily focus on knowledge-intensive queries grounded in the provided semantic context, without a particular focus on intense reasoning. This category comprises four datasets: \textbf{CSQA} (CommonsenseQA, \citet{talmor-etal-2019-commonsenseqa}), \textbf{MMLU} \cite{mmlu}, \textbf{QuALITY} \cite{pang-etal-2022-quality}, \textbf{GPQA} \cite{rein2024gpqa}.
Brief descriptions of all existing datasets are further provided in Appendix~\ref{app:dataset-desc}.

As all 12 tasks have a fixed answer space, with most of them being multi-choice questions, each answer is uniquely identifiable by a token in the vocabulary. Accordingly, the label set for the final-answer probing is constituted by 20 tokens in total, as detailed in Appendix~\ref{app:data-existing}.

\subsection{LLM Backbones}
\label{ssec:llm}

To obtain response rollouts and corresponding hidden states, we consider two types of LLM backbones described below.

% \vspace{-0.3em}
\paragraph{Off-the-Shelf LLM}
As our probing experiments require access to both model weights and reliable CoT outputs, we employ the open-source Qwen3 series with native support of both thinking and non-thinking modes. We use \textbf{Qwen3-32B} as the primary backbone to ensure robust performance while maintaining manageable computational cost.

% \vspace{-0.5em}
\paragraph{In-Domain LLM}
In addition to open-source LLMs with readily available thinking modes, we also employ an in-domain LLM trained with task-specific supervision, for two key reasons. First, a task-aware model exhibits more stable and decisive reasoning, thereby serving as an ``upper bound'' on internal planning capacity. Second, this setup helps reduce potential confounding factors inherent to general-purpose LLMs tied to specific model families.

Our In-Domain LLM learns task-aware CoT via reinforcement learning with GRPO \cite{shao2024deepseekmath}. We intentionally train from \textbf{Qwen2.5-7B-Instruct}, which does not have thinking mode natively, allowing for a cleaner bootstrap of CoT behavior on these tasks. We introduce our detailed GRPO training settings in Appendix~\ref{app:rl}.

\subsection{Experimental Settings}
\label{ssec:setting}

\paragraph{Dataset Construction}
We construct our probing datasets with train/dev/test splits across the 12 tasks, which contain up to 4000 / 100 / 500 problems per task, respectively. For the three tasks—Parity, Cycle, and Subsum—the problems are obtained via data generation. For other tasks, problems are sampled from their original datasets.
Details of our data generation and sampling, as well as further statistics are provided in Appendix~\ref{app:data-generation} and \ref{app:stats}.

% \vspace{-0.5em}
\paragraph{Training and Hyperparameters}
For each probing dimension, we train a dedicated \probe adapter for each Transformers layer of a LLM backbone, using a rank of $r=256$.
Each training run is conducted for approximately 5K steps, with early stopping on the dev set. More hyperparameters for adapter training are provided in Appendix~\ref{app:hyper}.

\begin{figure*}[!t]
\centering
\includegraphics[width=\textwidth]{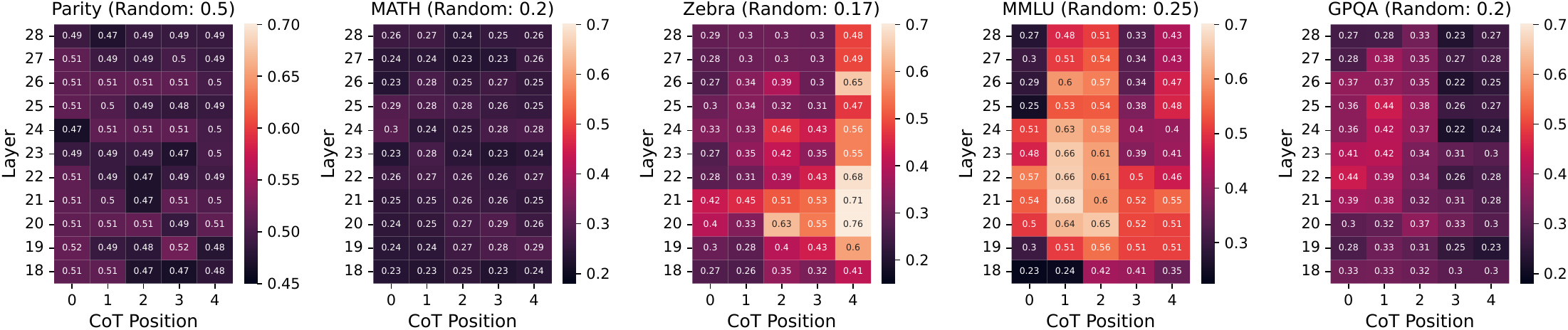}
\caption{Results for the final-answer probing: average accuracy of In-Domain LLM for the first five tokens within CoT trajectories, measured across selected Transformers layers and tasks. The full figure across all tasks is presented in \cref{fig:probing-answer-by-layer-full} (see Appendix~\ref{app:probe-results}).}
\label{fig:probing-answer-by-layer}
% \vspace{-0.5em}
\end{figure*}

\begin{figure*}[!th]
    \centering
    \begin{subfigure}{{0.48\textwidth}}
        \centering
        \includegraphics[width=\linewidth]{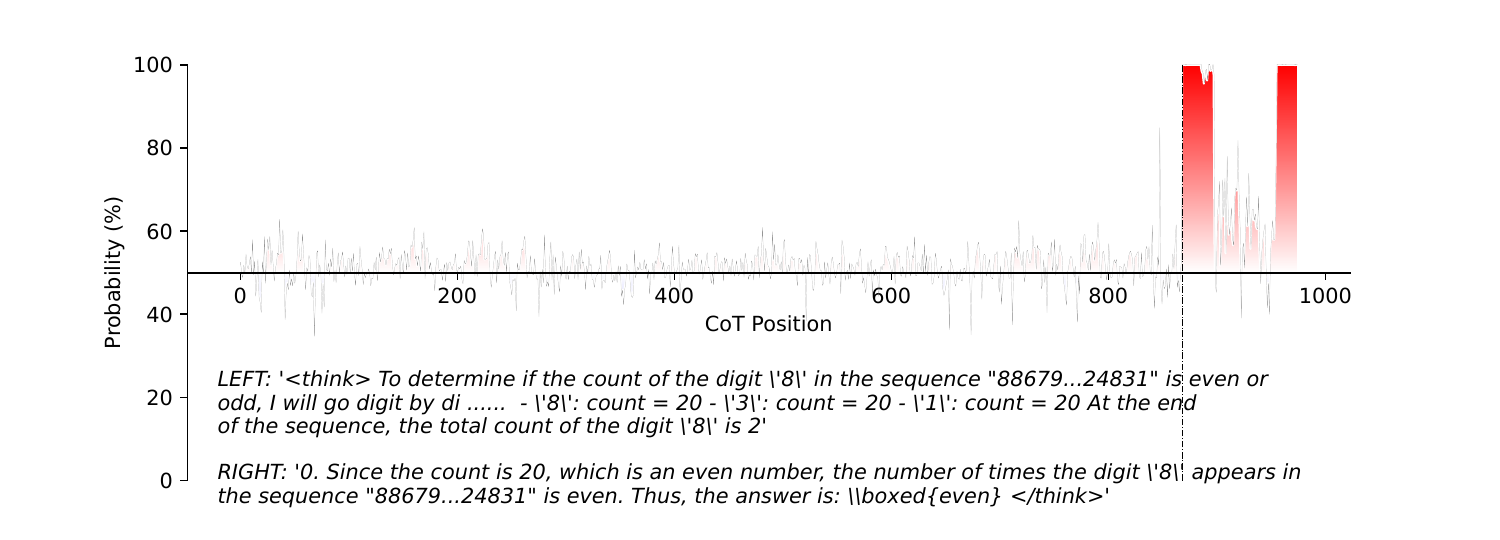}
        \caption{Parity example with In-Domain LLM.}
    \end{subfigure}
    \hspace{1em} % Horizontal space between images
    \begin{subfigure}{0.48\textwidth}
        \centering
        \includegraphics[width=\linewidth]{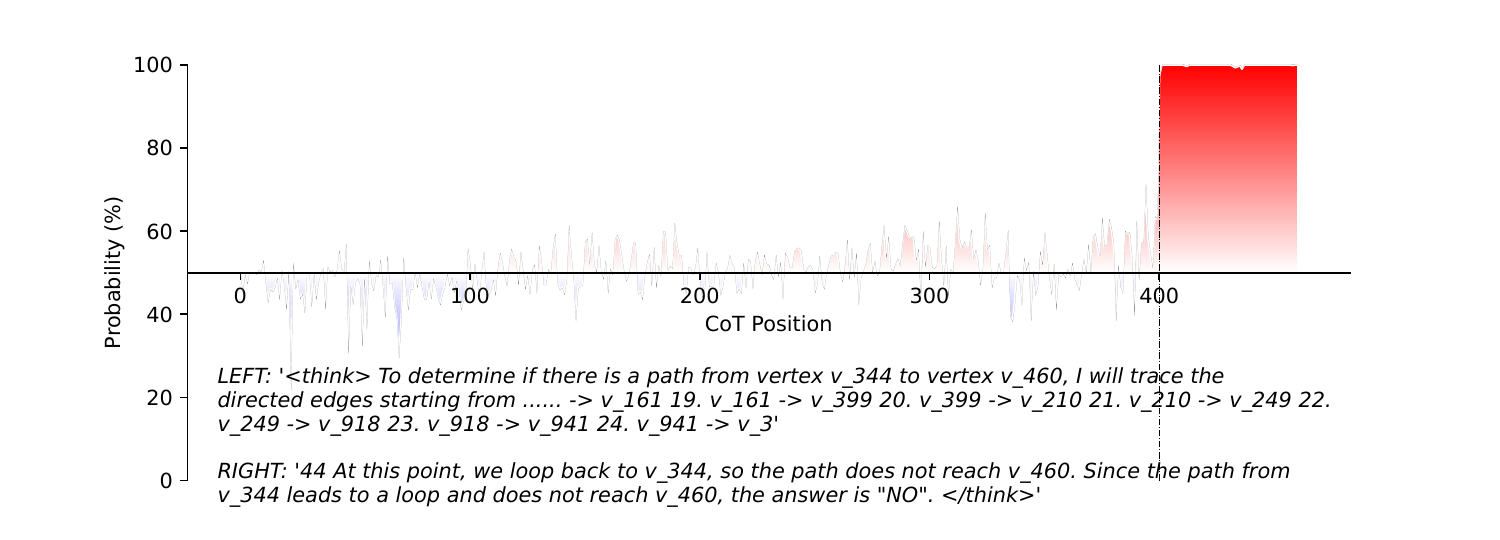}
        \caption{Cycle example with In-Domain LLM.}
    \end{subfigure}

    \caption{Examples of final-answer probing accuracy along full CoT trajectories with In-Domain LLM (random guessing is at 50\%). The vertical dashed line indicates the position at which accuracy first spikes. ``\textsl{LEFT}'' and ``\textsl{RIGHT}'' at the bottom illustrate the reasoning details right before and after the accuracy spike, respectively. Similar examples with Off-the-Shelf LLM are provided in \cref{fig:full-32b}.}
    \label{fig:full}
\vspace{-0.5em}
\end{figure*}

\subsection{Empirical Results}
\label{ssec:results}

We first report the performance of LLM backbones to characterize the 12 tasks, evaluating off-the-shelf Qwen3 and our trained In-Domain LLM. Full results are provided in \cref{tab:perf} (Appendix~\ref{app:llm-eval}), from which we draw the observations:
\begin{itemize}[leftmargin=1em]
\item For those compositional tasks requiring explicit multi-step reasoning, direct answering without CoT can only achieves near-random performance (e.g. Parity, Cycle), corroborating prior findings on the expressivity limits of Transformers \cite{chiang-cholak-2022-overcoming,merrill-sabharwal-2023-parallelism,zubic2025limits}. For other tasks, CoT generally yields substantial improvement as well.
\item Owing to differences in model generation and scale, our in-domain LLM underperforms the naive Qwen3 models on certain datasets. Despite this, it achieves the best performance on three compositional tasks and attains overall performance comparable to Qwen3, while producing substantially shorter CoT trajectories (approximately 1K+ characters per CoT, compared to 10K+ for Qwen3). These results validate the training effectiveness in inducing \textbf{more stable and decisive reasoning paths}. A qualitative CoT comparison is provided in Appendix~\ref{app:cot-example}. The latent planning horizon of In-Domain LLM is thus viewed as an ``upper bound'' for these tasks.
\end{itemize}

\vspace{-0.5em}
With \probe adapters trained and evaluated on the collected CoT trajectories, we present the empirical observations for each probing dimension as follows.

\subsubsection{Planning for Final Answers}
\label{ssec:result-answer}

\cref{fig:probing-answer-by-layer} first presents the average probing accuracy with In-Domain LLM along the initial CoT positions (full results across all tasks in \cref{fig:probing-answer-by-layer-full}). The overall trend with Off-the-Shelf Qwen3 is also similar as in \cref{fig:probing-answer-by-layer-full-32b}.
At first glance, it is clear that different Transformers layers exhibit varying predictive capacities. Notably, the highest performance does not occur at the final layer, but rather at layers between the middle and the last, which is consistent with findings of prior works that intermediate layers encode richer semantic information \cite{NEURIPS2019_159c1ffe,gari-soler-apidianaki-2021-lets,skean2025layer}.
For analyses in this section, we focus on results by layer 48 (64 total) for Off-the-Shelf Qwen3 and layer 21 (28 total) for In-Domain LLM.

It is worth noting that although \cref{fig:probing-answer-by-layer} plots only the initial positions, the probing is performed over the complete CoT trajectories. We further present the accuracy evolution along the full individual reasoning traces in \cref{fig:full}, \cref{fig:full-32b} and  \cref{fig:full-id-more}.
Overall, we draw two key findings as below.

\begin{myquote}
$\blacktriangleright$ \textbf{\emph{For precise future planning, LLMs exhibit a myopic horizon rather than long-term foresight.}}
\end{myquote}

To illustrate this, we focus on explicit compositional tasks, where their initial final-answer planning is near random, shown by Parity, Cycle and Subsum in \cref{fig:probing-answer-by-layer-full}\&\ref{fig:probing-answer-by-layer-full-32b}. Analysis of the full planning dynamics, depicted by the two examples in \cref{fig:full}, reveals that the precise final-answer only emerges one step before the reasoning completion, such that the probability of final answers remain flat before the final spike in the end: the final answer for Parity is planned only after the counting of all digits, and for Cycle, it is planned only after observing a complete path or cycle. As answer tokens are natural continuations of the CoT reasoning process, they act as anchors that reflect short-sighted internal planning, which fails to capture multi-step foresight.

To demonstrate quantitatively, we parse the CoT trajectories and obtain the final-answer probability at each critical intermediate steps. For Parity, we report the probabilities of CoT positions right after counting each digit. As shown in \cref{tab:myopic}, the probing only converges at the final counting position, exceeding 90\%, while for preceding positions, the accuracy hovers around random guessing as 50\%.

\begin{figure*}[!t]
\centering
\includegraphics[width=\textwidth]{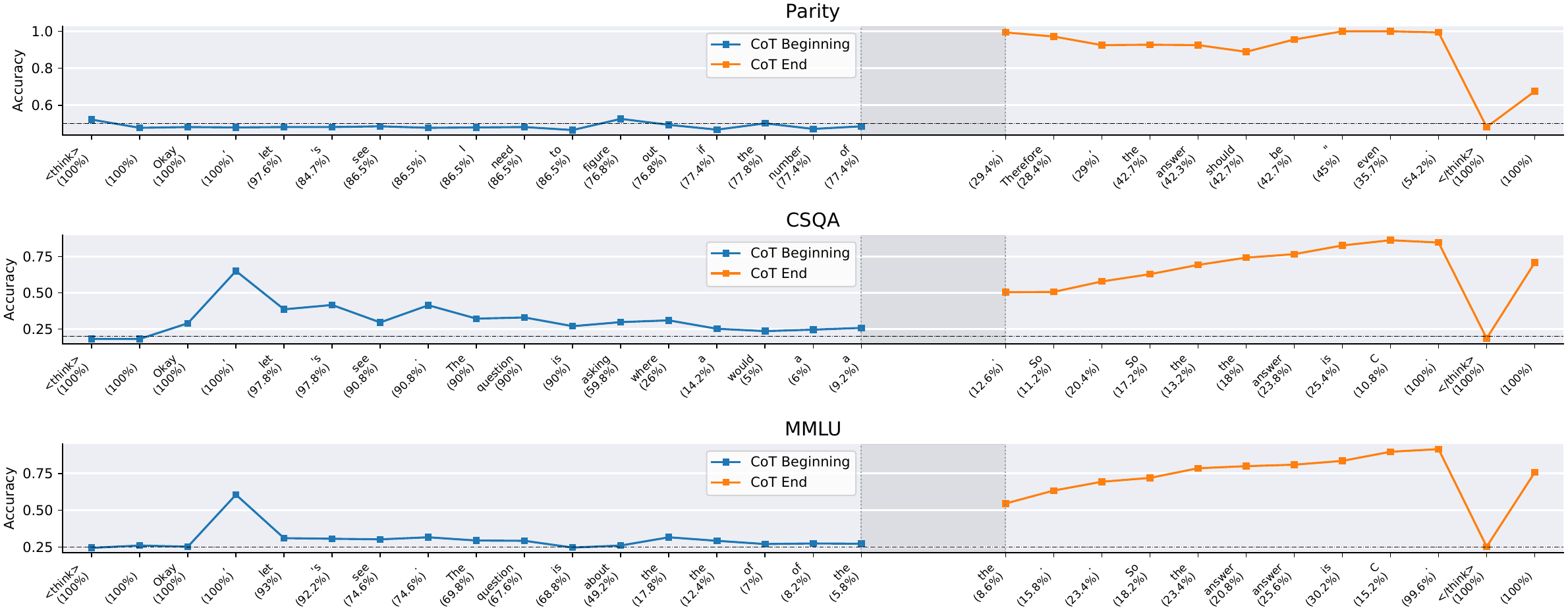}
\caption{Average final-answer probing accuracy on three tasks with Off-the-Shelf LLM (Qwen3-32B) along CoT positions. The most frequent token at each position is annotated with its occurrence frequency. The notably earlier accuracy spikes are especially pronounced for Knowledge and Semantic tasks, but largely remain flat for Compositional tasks. The full results across all tasks are shown in \cref{fig:spike-32b-full-a} for Off-the-Shelf LLM and \cref{fig:spike-32b-full-b} for In-Domain LLM (Appendix~\ref{app:probe-results}) .}
\label{fig:spike-32b}
% \vspace{-0.5em}
\end{figure*}

\begin{figure*}[!t]
\centering
\includegraphics[width=\textwidth]{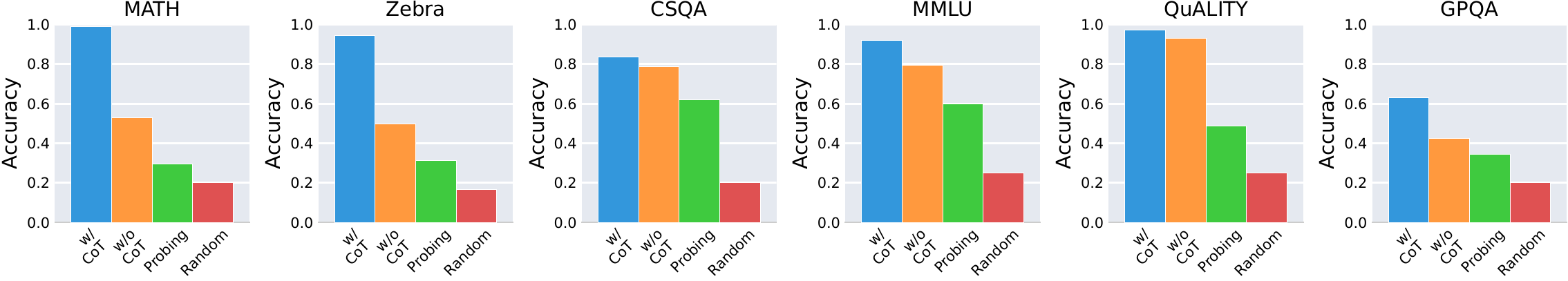}
\caption{Task accuracy comparison for Off-the-Shelf LLM (Qwen3-32B) under four settings: using thinking mode (\textbf{w/ CoT}); using non-thinking mode (\textbf{w/o CoT}); the best probing accuracy among initial CoT positions (\textbf{Probing}); the random-guess baseline (\textbf{Random}). The coarse signals of early final-answer planning are shown inferior to the direct prediction counterpart without CoT involved. Full results across all tasks are provided in \cref{fig:acc-32b-full}. Similar comparisons for In-Domain LLM is provided in \cref{fig:acc-full}.}
\label{fig:acc-32b}
\vspace{-0.5em}
\end{figure*}

For simpler tasks, the probing also only stabilizes near the reasoning end. More illustrations are provided in \cref{fig:full-id-more}.

\begin{table}[!t]
\small
\centering
\caption{Probing probabilities for Parity at CoT positions immediately following the counting of each of the last four digits in the sequence (random as 0.5). Position \textbf{0} denotes the highest probing probability after all digits have been counted (the upper bound).}
\begin{tabular}{l|ccccc}
\toprule
& -4 & -3 & -2 & -1 & \bf 0 \\
\midrule
In-Domain LLM & 0.49 & 0.51 & 0.51 & 0.97 & \bf 0.99 \\
Off-the-Shelf LLM & 0.50 & 0.52 & 0.51 & 0.94 & \bf 0.97 \\
\bottomrule
\end{tabular}
\label{tab:myopic}
\vspace{-1em}
\end{table}

\begin{myquote}
$\blacktriangleright$ \textbf{\emph{LLMs can exhibit coarse signals for final answers in early stages of CoT, but reflecting only a pattern-matching gist, rather than precise reasoning plans.}}
\end{myquote}

As shown in \cref{fig:probing-answer-by-layer}, LLMs can sometimes ``sense'' the gist of the answer early on, particularly for those emphasizing semantic understanding rather than explicit multi-step reasoning.
To illustrate with more clarity, \cref{fig:spike-32b} depicts the probing dynamics on CSQA and MMLU, which focus more on semantics and knowledge, in which an early spike in probing accuracy is notably evident, in contrast to tasks like Parity.
By the full evaluation results presented in \cref{fig:spike-32b-full-a} and \cref{fig:spike-32b-full-b}, it appears that early hidden states do possess certain information predictive of the final answers, just as observed in prior works \cite{azaria-mitchell-2023-internal,gottesman-geva-2024-estimating,afzal-etal-2025-knowing}.

However, our in-depth analysis suggests that these coarse predictive signals primarily reflect a vague perceptual cue, but not resulting from exercising a pre-planned reasoning path. We proceed to compare the performance of this early coarse signal, with that of true reasoning via CoT, as well as direct answering without CoT; the results are presented in \cref{fig:acc-32b} (full results in  \cref{fig:acc-32b-full}).
Across almost all tasks, early final-answer planning yields lower task accuracy than both standard reasoning with CoT and direct answering without CoT.
Therefore, even with comparable reasoning budgets, early planning remains less effective than direct answering. The performance gap further widens substantially when CoT is applied, strongly indicating that such coarse signals \textbf{do not arise from precise plans in latent space}.

\subsubsection{Planning for Reasoning Path}
\label{ssec:result-subsequent}

Empirical results on probing subsequent tokens further advocate a myopic planning horizon of LLMs detailed below.

\begin{myquote}
$\blacktriangleright$ \textbf{\emph{LLM hidden states generally encode limited foresight over subsequent reasoning paths.}}
\end{myquote}

\begin{figure}[!t]
\centering
\includegraphics[width=\columnwidth]{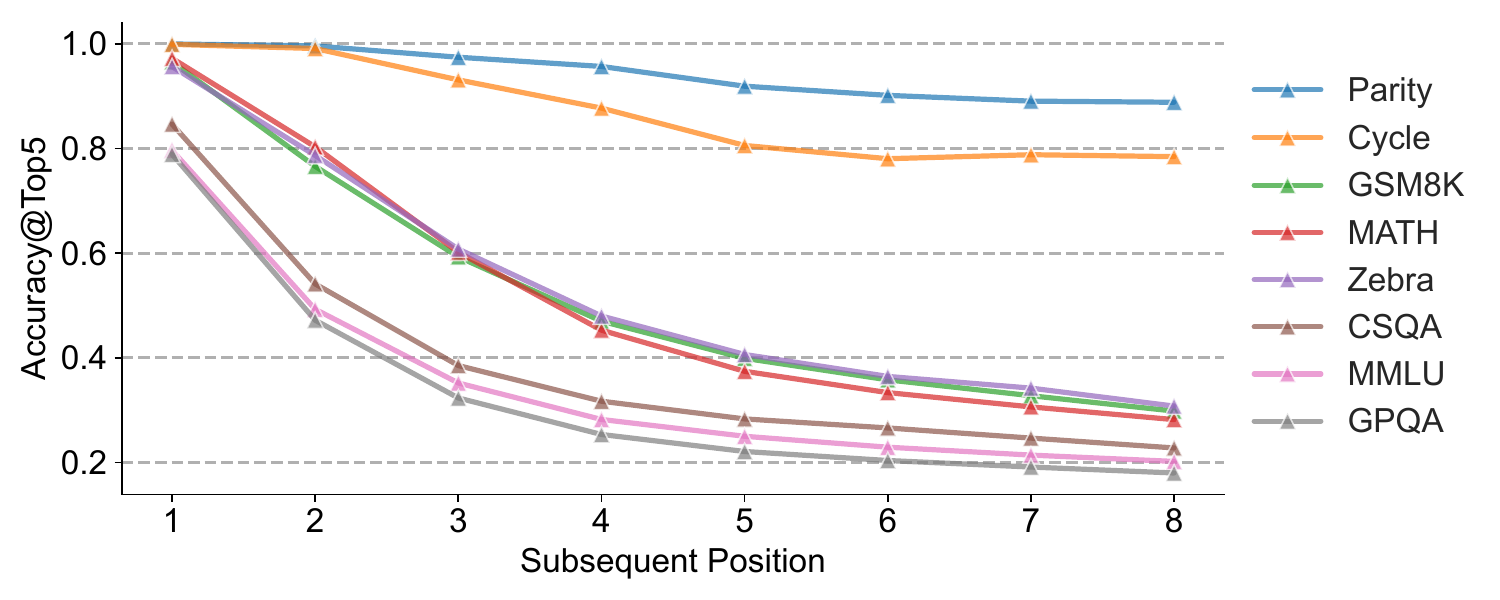}
\caption{Top-5 accuracy for subsequent token prediction, using the last Transformers layer of In-Domain LLM. Full results across layers and tasks are presented in \cref{fig:hop-full} and \cref{fig:hop-full-32b}.}
\label{fig:hop-drop}
\vspace{-1em}
\end{figure}

\begin{figure*}[!t]
\centering
\includegraphics[width=0.85\textwidth]{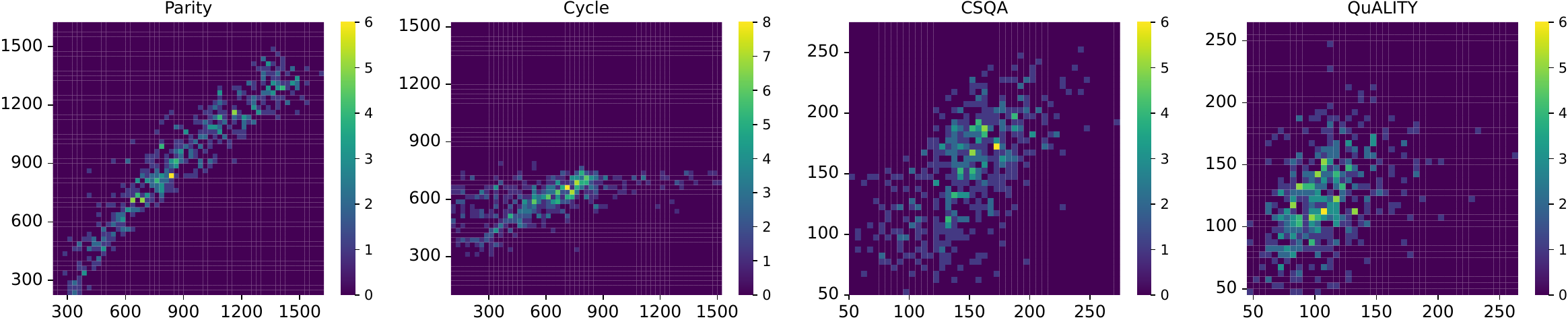}
\caption{Heatmap of the predicted reasoning length (\emph{y}-axis) using initial CoT hidden states against the actual reasoning length (\emph{x}-axis). The rather unreliable predictions on those tasks without structural solutions (e.g. CSQA) suggest that a precise global plan does not emerge early in CoT, even for the task-aware In-Domain LLM. Full results across tasks are provided in \cref{fig:length-full} and \ref{fig:length-full-32b}.}
\label{fig:length}
\vspace{-0.4em}
\end{figure*}

\begin{figure}[!ht]
    \centering
    \begin{subfigure}{{0.44\columnwidth}}
        \centering
        \includegraphics[width=\linewidth]{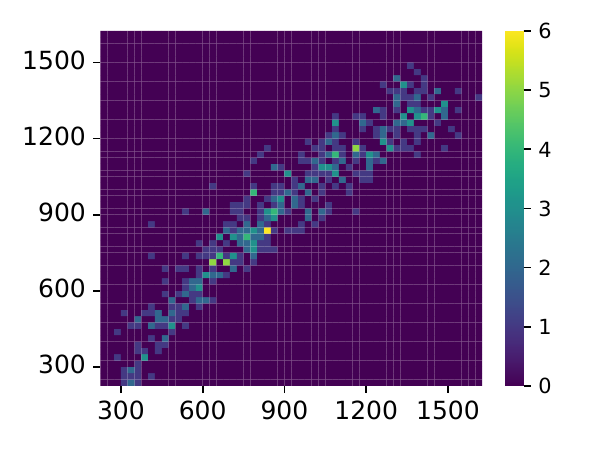}
        \caption{Reasoning length predictions for Parity.}
    \end{subfigure}
    \hspace{0.2em} % Horizontal space between images
    \begin{subfigure}{0.44\columnwidth}
        \centering
        \includegraphics[width=\linewidth]{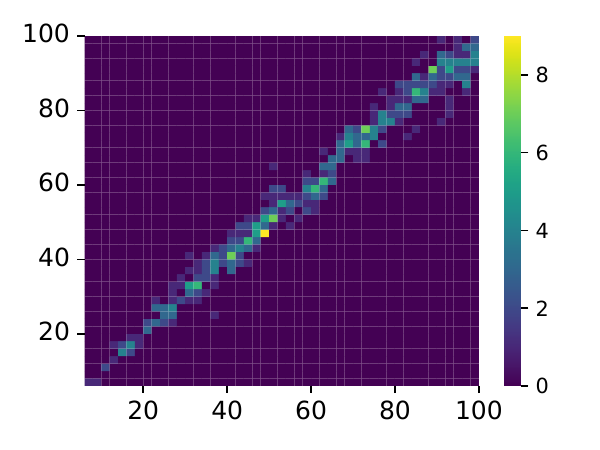}
        \caption{Input sequence length predictions for Parity.}
    \end{subfigure}

    \caption{Task-specific factors can confound reasoning length predictions. For Parity, the total length is typically proportional to the input sequence length, which can be perceivable by LLMs.}
    \label{fig:length-parity}
\vspace{-0.5em}
\end{figure}

For each hidden state, we assess subsequent token prediction performance up to its 8-th following token along the CoT trajectory. As LLM generation is a sampling process over a latent distribution, we measure by Top-5 Accuracy, deeming a prediction correct if the true subsequent token appears within the top-5 predictions.
\cref{fig:hop-drop} presents the evaluation results with In-Domain LLM, which show a clear overall decline in accuracy as the subsequent token position advances, specially for tasks dominated by semantic understanding and factual knowledge (e.g., MMLU and GPQA).

\cref{fig:hop-drop} also suggests that LLM does plan the subsequent path to a certain extent, with Top-5 accuracy exceeding 50\% for the next two steps. However, a more long-term planning is only limited to tasks with structural modularity, such as Parity or Cycle, whose reasoning trajectories exhibit discernible patterns (see the CoT example in \cref{fig:example-id}).
In general, hidden states lack a clear vision over subsequent reasoning.
Beyond In-Domain LLM as an ``upper bound'', a similar trend is also observed for Off-the-Shelf LLM, albeit with much lower accuracy across all tasks, especially with significant drop on structural tasks, as illustrated by the comparison between \cref{fig:hop-full} and \cref{fig:hop-full-32b}.

\subsubsection{Planning for Global Steps}
\label{ssec:result-step}

Reasoning length probing again indicates a lack of global planning prior to the emergence of CoT, as discussed below.

\begin{myquote}
$\blacktriangleright$ \textbf{\emph{LLMs do not grasp the reasoning length precisely, though task-specific heuristics may offer shortcuts.}}
\end{myquote}

In general, if LLMs possessed a global reasoning view in sight, early hidden states would be predictive of the total length across input domains.
However, our empirical results suggest that the initial CoT hidden states hardly have reliable internal clock for global reasoning length, for both In-Domain and Off-the-Shelf LLMs, as illustrated by the unstable and often low correlations across most tasks, shown by the heatmaps in \cref{fig:length} (full plots in \cref{fig:length-full} and \ref{fig:length-full-32b}).

On closer inspection, two tasks appear to be exceptions, Parity and Subsum, which exhibit high correlation with the true reasoning lengths, as in \cref{fig:length-full}. However, interpreting this as evidence of robust CoT planning on these tasks can be misleading.
We highlight the attribution of task-specific confounding factors, illustrated in \cref{fig:length-parity}: for both tasks, reasoning paths are typically in proportional to the input sequence length, which is readily observable by LLMs and thus could serve as a shortcut signal in probing.
In contrast, for Cycle as in \cref{fig:length}, such shortcut does not apply, as its reasoning length scales with the path between two vertices rather than the input length (example in \cref{fig:example-cycle-id}), which LLMs have difficulty estimating directly. The discrepancy between Parity/Subsum and Cycle further underscores the limited presence of actual global planning in LLMs.

% \vspace{-2em}
\section{Leveraging CoT Dynamics}
\label{sec:utility}

Given the \emph{myopic} planning horizon observed in our probing experiments, we highlight the significance of exploiting such CoT dynamics, and demonstrate how these planning characteristics can be leveraged to estimate both the uncertainty and the necessity of CoT.

\subsection{CoT Uncertainty Estimation}
\label{ssec:uncertainty}
For language models, general metrics such as perplexity or entropy are standard to estimate the inference confidence. A well-calibrated uncertainty metric should ideally assign high scores to correct outputs and lower scores to uncertain ones.
In our studies, we target metrics that utilize internal signals within CoT trajectories, focusing on the three general uncertainty metrics described below.

(i) \emph{Perplexity}, equivalent to the average Negative Log-Likelihood of the sequence (NLL) for uncertainty. For a sequence $X$ with $N$ tokens $\{x_1, x_2, .., x_N\}$, NLL is:
\begin{equation}
    \text{NLL}(X) = -\frac{1}{N} \sum_{i=1}^{N} \log P(x_i \mid x_{<i})
\end{equation}

(ii) Average \emph{entropy} H across tokens in the sequence, with $\mathcal{V}$ being the model vocabulary:
\begin{equation}
    \text{H}(X) = \frac{1}{N} \sum_{i=1}^{N} \left( -\sum_{w \in \mathcal{V}} P(w | x_{<i}) \log P(w | x_{<i}) \right)
\end{equation}

(iii) \emph{Self-Certainty} (SC) \cite{kang2025scalable}, defined on the vocabulary distribution as below:
\begin{equation}
    \text{SC}(X) = \frac{-1}{N|\mathcal{V}|} \sum_{i=1}^{N} \sum_{w \in \mathcal{V}} \log \left( |\mathcal{V}| \cdot P \left( w \mid x_{<i} \right) \right)
\end{equation}

Note that the three metrics all aggregate over entire CoT positions in the sequence.
Intuitively, however, tokens that steer the reasoning process within CoT are often sparse; the majority of tokens function as ``syntactic fillers'' necessary for linguistic coherence.
These filler tokens are usually high-confidence local transitions, as evidenced by the density distributions presented in \cref{fig:ent-density}, aligned with prior findings that most tokens in LLM are of low entropy \cite{wang2025beyond,li2026making}.

\paragraph{Wooden Barrel Principle}
Building on the local planning cues in \cref{sec:horizon}, we speculate that the internal signals localized around a few critical tokens are more informative than trajectory-wide aggregates, where a conventional global averaging across all generated tokens may \textbf{dilute the sensitivity} of the uncertainty estimation.
We thus posit such principle: just like a barrel's capacity is determined by its shortest stave, analogously, we hypothesize that the uncertainty of a reasoning chain is governed a subset of critical logical leaps, which we term \emph{reasoning pivots}. 

We conduct empirical validation and demonstrate that focusing on these pivot positions, even through a simple top-$k$ selection strategy, could yield cleaner signals for uncertainty calibration.
Our validation utilizes two orthogonal sources of latent signals, as described below.

% \vspace{-1em}
\paragraph{Latent Signals by \probe}
Before we proceed with general metrics of uncertainty estimation, we first demonstrate that latent signals from a sparse subset of tokens are indeed effective to characterize the uncertainty of the whole reasoning trajectory.
Motivated by \cref{fig:full}, where specific positions exhibit significant accuracy spikes during final-answer probing, we propose utilizing the \emph{entropy} from \probe to identify pivot positions: along a CoT path, we select top-$k$ positions with the lowest final-answer entropy, as a proxy to indicate the confidence level of the entire path.

Accordingly, we conduct preliminary experiments with In-Domain LLM using the last Transformers layer: after the top-$k$ positions are selected, we obtain their average of final-answer entropy as a new uncertainty metric. The results, measured by standard AUROC, are presented in \cref{tab:id-uncertainty}.

\begin{table}[tbp!]
\centering
\caption{Uncertainty estimation results (AUROC) with In-Domain LLM, using latent signals from final-answer probing via \probe (\cref{ssec:uncertainty}); values closer to 1 indicate better calibration. Using a subset of 5 positions along the CoT can better capture the uncertainty of the full path, with 9\% substantial improvement over the best baseline. Full results across all tasks are shown in \cref{tab:id-uncertainty-full}.}
\resizebox{\columnwidth}{!}{
\begin{tabular}{l|cccc|c}
\toprule
& GSM8K & Zebra & MMLU & GPQA & \bf Avg.\\
\midrule
Perplexity & 0.70 & 0.58 & 0.53 & 0.50 & 0.57 \\
Entropy & 0.72 & 0.60 & 0.52 & 0.50 & 0.58 \\
Self-Certainty & 0.76 & 0.67 & 0.53 & 0.51 & 0.60 \\
\cmidrule{1-6}
\probe (Top-5) & \bf 0.87 & \bf 0.77 & \bf 0.73 & \bf 0.56 & \bf 0.69 \\
\probe (Top-10) & 0.81 & 0.75 & 0.72 & 0.56 & 0.68 \\
\probe (Top-20) & 0.82 & 0.67 & 0.65 & 0.51 & 0.63 \\
\probe (Top-50) & 0.78 & 0.69 & 0.56 & 0.47 & 0.64 \\
\bottomrule
\end{tabular}}
\label{tab:id-uncertainty}
\vspace{-1.5ex}
\end{table}

Comparing against conventional baselines over the full path, our top-$k$ selection strategy upon \probe signals achieves up to 9\% absolute improvement upon the best baseline. Notably, the best estimation is obtained with $k=5$ pivot tokens, demonstrating that latent signals from only a few positions can be a strong indicator of the whole path.

\begin{table*}[tbp!]
\centering
\caption{Uncertainty estimation results (AUROC) with Qwen3-32B using the last Transformers layer, applying our top-$k$ strategy upon each general metric. Note that the average CoT length across inputs exceeds 7K tokens, while our simple strategy that selects top-$100$ positions is able to yield steady improvement. Full results across tasks with both 8B and 32B models are provided in \cref{tab:qwen-uncertainty-full}.}
\resizebox{0.76\textwidth}{!}{
\begin{tabular}{l|cccccccc|c}
\toprule
& GSM8K & MATH & MuSR & Zebra & CSQA & MMLU & QuALITY & GPQA & \bf Avg.\\
\midrule
Perplexity & 0.71 & \bf 0.93 & 0.48 & 0.74 & 0.68 & 0.76 & 0.78 & 0.69 & 0.72 \\
\quad w/ 100 Pivots & \bf 0.81 & 0.92 & \bf 0.50 & \bf 0.90 & \bf 0.74 & \bf 0.81 & \bf 0.82 & \bf 0.73 & \bf 0.78 \\
\cmidrule{1-10}
Entropy & 0.71 & \bf 0.92 & 0.47 & 0.77 & 0.68 & 0.77 & 0.77 & 0.68 & 0.72 \\
\quad w/ 100 Pivots & \bf 0.81 & 0.70 & \bf 0.49 & \bf 0.90 & \bf 0.74 & \bf 0.83 & \bf 0.82 & \bf 0.74 & \bf 0.75 \\
\cmidrule{1-10}
Self-Certainty & 0.45 & 0.82 & \bf 0.47 & 0.92 & 0.51 & 0.67 & 0.64 & 0.68 & 0.65 \\
\quad w/ 100 Pivots & \bf 0.55 & \bf 0.90 & \bf 0.47 & \bf 0.93 & \bf 0.59 & \bf 0.74 & \bf 0.70 & \bf 0.70 & \bf 0.70 \\
\bottomrule
\end{tabular}}
\label{tab:qwen-uncertainty}
% \vspace{-0.5ex}
\end{table*}

\vspace{-0.5em}
\paragraph{Latent Signals by General Metrics}
We next extend our validation to general scenarios without involving signals from a dedicated prober, lifting the constraint of a fixed answer space. We consider the three general metrics derived solely from predicted next-token logits over the model's vocabulary.
For the generalizability of our findings, we conduct experiments with Off-the-Shelf LLMs, using both Qwen3-8B and Qwen3-32B.
Specifically, we select the top-$k$ positions along a thinking path with the highest entropy / self-certainty, or with the lowest log-likelihood, respectively, representing the \textbf{most uncertain local steps} (the \emph{shortest staves}). We use the average among these positions of each corresponding metric as the final estimation.

As shown in \cref{tab:qwen-uncertainty}, for each LLM, applying the top-$k$ selection brings no negative impact using the selected $k$ values. The improvement is especially pronounced with Qwen3-32B: $k=100$ consistently drives 3+\% absolute improvement across all metrics, reaching up to 6\%, thereby supporting the efficacy of our hypothesis.

\begin{figure}[!ht]
\centering
\includegraphics[width=0.8\columnwidth]{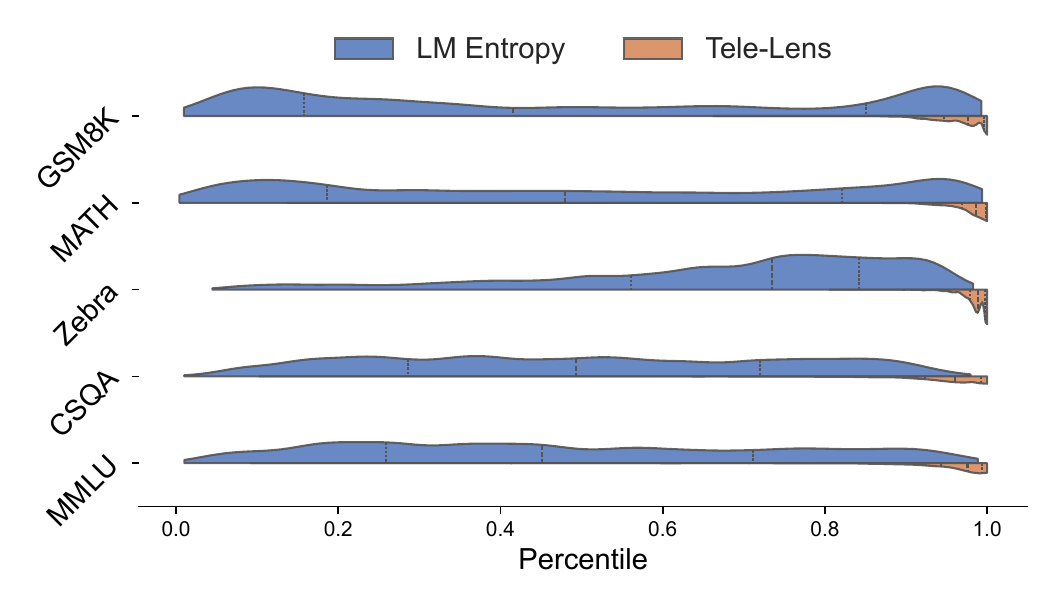}
\caption{Spatial density distribution of selected \emph{pivot} positions with In-Domain LLM along CoT paths. Using \probe, the selected positions tend to concentrate near CoT completion, whereas positions selected by general LM entropy typically are distributed across the entire CoT trajectory. Integrating multiple sources of latent signals may spur further improvement.}
\label{fig:spatial}
% \vspace{-1em}
\end{figure}

Furthermore, we highlight the potential of exploiting richer latent signals and strategies for more effective calibration. On the one hand, the signals by \probe are highly informative, suggesting that additional training can be beneficial, consistent with prior findings that raw confidence metrics from LLMs alone may be sub-optimal \cite{kapoor2024large}. On the other hand, the spatial distributions of the selected \emph{pivots} differ substantially between signals derived from \probe and those based on general metrics, as illustrated in \cref{fig:spatial}. These divergent distributions indicate that latent signals from multiple sources may provide complementary information, and their integration can be promising to further improve the identification of critical positions, leading to more robust uncertainty calibration.

\subsection{CoT Necessity Estimation}
\label{ssec:necessity}

We next conduct a study to estimate the necessity of CoT reasoning. In the broader context of improving model thinking efficiency, prior work has proposed various approaches for reducing CoT overhead, such as early exiting \cite{yong2026think,yang2026dynamic} and CoT length steering \cite{li2026efficient}.
In this study, we investigate from a different perspective: assessing the input complexity to determine whether CoT reasoning is necessary in the first place.

Motivated by the observations in \cref{fig:spike-32b}, we leverage the early answer gist revealed by final-answer probing, where the probing accuracy may spike at the initial CoT positions for easier tasks. 
Our experiments show that these early signals can be effective recognizing whether a full CoT generation is required to accurately derive the final answer.
By selectively bypassing CoT generation in non-essential cases, we can achieve a reduction in computational load with negligible performance degradation.

For each rollout specifically, we first generate its initial five CoT tokens and assess the normalized final-answer entropy $\bar{\mathrm{H}}$ over logit distribution $\mathbf{p}$ across $C=20$ probing classes:
\begin{equation}
    \bar{\mathrm{H}}(\mathbf{p}) = (-\sum_{i=1}^{C} p_i \log p_i) / \log C
\end{equation}
As $\bar{\mathrm{H}}$ lies in the range $[0, 1]$, we adopt a threshold-based strategy: for initial positions, if any of their normalized entropy falls below a predefined threshold, representing a confident answer gist, we halt the corresponding CoT generation and directly output the answer by disabling the LLM's thinking mode, bypassing a full generation.
The evaluation results are reported in \cref{tab:reduction}.

With the threshold set to $0.1$, our objective is robustly accomplished for both In-Domain and Off-the-Shelf LLMs: the aforementioned heuristic automatically recognizes inputs for which CoT is necessary to derive the final answer, such as Parity, while bypassing CoT generation on easier tasks, such as CSQA. For instance, Qwen3-32B attains 16.2\% / 12.4\% thinking reduction for CSQA / MMLU almost ``for free'', with only 0.03\% overall accuracy degradation.

\begin{table}[!t]
\centering
\caption{Evaluation results for CoT bypass, varying thresholds of normalized entropy from final-answer probing. The bypass ratio for each task is reported. \textbf{Avg.}: average bypass ratio; \textbf{Perf.}: average accuracy change. Full results are provided in \cref{tab:reduction-full}.}
\resizebox{\columnwidth}{!}{
\begin{tabular}{l|cccc|cc}
\toprule
& Parity & CSQA & MMLU & GPQA & \bf Avg. & \bf Perf.\\
\midrule
\multicolumn{7}{c}{In-Domain LLM}\\
\cmidrule{1-7}
Th=0.1 & 0\% & 40.2\% & 30.4\% & 7\% & 13.3\% & -0.47 \\
Th=0.2 & 0\% & 65\% & 45\% & 12\% & 21.6\% & -1.42 \\
\midrule
\multicolumn{7}{c}{Off-the-Shelf LLM (Qwen3-32B)}\\
\cmidrule{1-7}
Th=0.1 & 0\% & 16.2\% & 12.4\% & 1.2\% & 2.8\% & -0.03 \\
Th=0.2 & 0\% & 28.8\% & 20.2\% & 3.2\% & 6.2\% & -0.37 \\
\bottomrule
\end{tabular}}
\label{tab:reduction}
\vspace{-1.5ex}
\end{table}

As our necessity estimation relies on a fixed answer space to obtain the metric, we present it primarily as a \textsl{proof-of-concept}. Nevertheless, we underscore the significance of exploiting such useful latent signals, which can benefit various scenarios, such as to facilitate model training \cite{huang-etal-2025-enhancing} and CoT compression \cite{singh2026llmsencodefunctionalimportance,li2026making} and steering \cite{li2026efficient}.
We provide further discussions and Related Works in Appendix~\ref{app:related}.

\section{Conclusion}

In this work, we investigate the internal planning capacity of LLMs and uncover a myopic planning horizon during CoT generation. For analysis, we design a series of probing experiments using our proposed method, \probe, and we highlight the exploitation of such latent signals, demonstrating on both CoT uncertainty and necessity estimation.

\section*{Impact Statement}

This paper presents work whose goal is to advance the understanding of internal dynamics of Large Language Models, in particular the latent planning horizon and the according utilization. There may be potential societal consequences of this work, none which we feel must be specifically highlighted here.

\bibliography{icml2026}

@inproceedings{bert,
    title = "{BERT}: Pre-training of Deep Bidirectional Transformers for Language Understanding",
    author = "Devlin, Jacob  and
      Chang, Ming-Wei  and
      Lee, Kenton  and
      Toutanova, Kristina",
    editor = "Burstein, Jill  and
      Doran, Christy  and
      Solorio, Thamar",
    booktitle = "Proceedings of the 2019 Conference of the North {A}merican Chapter of the Association for Computational Linguistics: Human Language Technologies, Volume 1 (Long and Short Papers)",
    month = jun,
    year = "2019",
    address = "Minneapolis, Minnesota",
    publisher = "Association for Computational Linguistics",
    url = "https://aclanthology.org/N19-1423/",
    doi = "10.18653/v1/N19-1423",
    pages = "4171--4186",
}

@article{r1,
   title={DeepSeek-R1 incentivizes reasoning in LLMs through reinforcement learning},
   volume={645},
   ISSN={1476-4687},
   url={http://dx.doi.org/10.1038/s41586-025-09422-z},
   DOI={10.1038/s41586-025-09422-z},
   number={8081},
   journal={Nature},
   publisher={Springer Science and Business Media LLC},
   author={DeepSeek-AI},
   year={2025},
   month=sep, pages={633–638} }

@inproceedings{
dziri2023faith,
title={Faith and Fate: Limits of Transformers on Compositionality},
author={Nouha Dziri and Ximing Lu and Melanie Sclar and Xiang Lorraine Li and Liwei Jiang and Bill Yuchen Lin and Sean Welleck and Peter West and Chandra Bhagavatula and Ronan Le Bras and Jena D. Hwang and Soumya Sanyal and Xiang Ren and Allyson Ettinger and Zaid Harchaoui and Yejin Choi},
booktitle={Thirty-seventh Conference on Neural Information Processing Systems},
year={2023},
url={https://openreview.net/forum?id=Fkckkr3ya8}
}

@inproceedings{
wies2023subtask,
title={Sub-Task Decomposition Enables Learning in Sequence to Sequence Tasks},
author={Noam Wies and Yoav Levine and Amnon Shashua},
booktitle={The Eleventh International Conference on Learning Representations },
year={2023},
url={https://openreview.net/forum?id=BrJATVZDWEH}
}

@inproceedings{bhattamishra-etal-2023-simplicity,
    title = "Simplicity Bias in Transformers and their Ability to Learn Sparse {B}oolean Functions",
    author = "Bhattamishra, Satwik  and
      Patel, Arkil  and
      Kanade, Varun  and
      Blunsom, Phil",
    editor = "Rogers, Anna  and
      Boyd-Graber, Jordan  and
      Okazaki, Naoaki",
    booktitle = "Proceedings of the 61st Annual Meeting of the Association for Computational Linguistics (Volume 1: Long Papers)",
    month = jul,
    year = "2023",
    address = "Toronto, Canada",
    publisher = "Association for Computational Linguistics",
    url = "https://aclanthology.org/2023.acl-long.317/",
    doi = "10.18653/v1/2023.acl-long.317",
    pages = "5767--5791",
}

@inproceedings{
anil2022exploring,
title={Exploring Length Generalization in Large Language Models},
author={Cem Anil and Yuhuai Wu and Anders Johan Andreassen and Aitor Lewkowycz and Vedant Misra and Vinay Venkatesh Ramasesh and Ambrose Slone and Guy Gur-Ari and Ethan Dyer and Behnam Neyshabur},
booktitle={Advances in Neural Information Processing Systems},
editor={Alice H. Oh and Alekh Agarwal and Danielle Belgrave and Kyunghyun Cho},
year={2022},
url={https://openreview.net/forum?id=zSkYVeX7bC4}
}

@inproceedings{
abbe2024how,
title={How Far Can Transformers Reason? The Globality Barrier and Inductive Scratchpad},
author={Emmanuel Abbe and Samy Bengio and Aryo Lotfi and Colin Sandon and Omid Saremi},
booktitle={The Thirty-eighth Annual Conference on Neural Information Processing Systems},
year={2024},
url={https://openreview.net/forum?id=FoGwiFXzuN}
}

@inproceedings{
li2024chain,
title={Chain of Thought Empowers Transformers to Solve Inherently Serial Problems},
author={Zhiyuan Li and Hong Liu and Denny Zhou and Tengyu Ma},
booktitle={The Twelfth International Conference on Learning Representations},
year={2024},
url={https://openreview.net/forum?id=3EWTEy9MTM}
}

@inproceedings{
zubic2025limits,
title={Limits of Deep Learning: Sequence Modeling through the Lens of Complexity Theory},
author={Nikola Zubic and Federico Sold{\`a} and Aurelio Sulser and Davide Scaramuzza},
booktitle={The Thirteenth International Conference on Learning Representations},
year={2025},
url={https://openreview.net/forum?id=DhdqML3FdM}
}

@inproceedings{
xiao2025generalizing,
title={Generalizing Reasoning Problems to Longer Lengths},
author={Changnan Xiao and Bing Liu},
booktitle={The Thirteenth International Conference on Learning Representations},
year={2025},
url={https://openreview.net/forum?id=zpENPcQSj1}
}

@inproceedings{
wang2025beyond,
title={Beyond the 80/20 Rule: High-Entropy Minority Tokens Drive Effective Reinforcement Learning for {LLM} Reasoning},
author={Shenzhi Wang and Le Yu and Chang Gao and Chujie Zheng and Shixuan Liu and Rui Lu and Kai Dang and Xiong-Hui Chen and Jianxin Yang and Zhenru Zhang and Yuqiong Liu and An Yang and Andrew Zhao and Yang Yue and Shiji Song and Bowen Yu and Gao Huang and Junyang Lin},
booktitle={The Thirty-ninth Annual Conference on Neural Information Processing Systems},
year={2025},
url={https://openreview.net/forum?id=yfcpdY4gMP}
}

@inproceedings{
liu2025mind,
title={Mind Your Step (by Step): Chain-of-Thought can Reduce Performance on Tasks where Thinking Makes Humans Worse},
author={Ryan Liu and Jiayi Geng and Addison J. Wu and Ilia Sucholutsky and Tania Lombrozo and Thomas L. Griffiths},
booktitle={Forty-second International Conference on Machine Learning},
year={2025},
url={https://openreview.net/forum?id=J3gzdbYZxS}
}

@inproceedings{
sprague2025to,
title={To CoT or not to CoT? Chain-of-thought helps mainly on math and symbolic reasoning},
author={Zayne Rea Sprague and Fangcong Yin and Juan Diego Rodriguez and Dongwei Jiang and Manya Wadhwa and Prasann Singhal and Xinyu Zhao and Xi Ye and Kyle Mahowald and Greg Durrett},
booktitle={The Thirteenth International Conference on Learning Representations},
year={2025},
url={https://openreview.net/forum?id=w6nlcS8Kkn}
}

@inproceedings{
li2023emergent,
title={Emergent World Representations: Exploring a Sequence Model Trained on a Synthetic Task},
author={Kenneth Li and Aspen K Hopkins and David Bau and Fernanda Vi{\'e}gas and Hanspeter Pfister and Martin Wattenberg},
booktitle={The Eleventh International Conference on Learning Representations },
year={2023},
url={https://openreview.net/forum?id=DeG07_TcZvT}
}

@misc{nostalgebraist2021logitlens,
  title={logit lens on non-gpt2 models + extensions},
  author={nostalgebraist},
  year={2021},
  url={https://colab.research.google.com/drive/1MjdfK2srcerLrAJDRaJQKO0sUiZ-hQtA}
}

@misc{belrose2025elicitinglatentpredictionstransformers,
      title={Eliciting Latent Predictions from Transformers with the Tuned Lens}, 
      author={Nora Belrose and Igor Ostrovsky and Lev McKinney and Zach Furman and Logan Smith and Danny Halawi and Stella Biderman and Jacob Steinhardt},
      year={2023},
      eprint={2303.08112},
      archivePrefix={arXiv},
      primaryClass={cs.LG},
      url={https://arxiv.org/abs/2303.08112}, 
}

@inproceedings{pal-etal-2023-future,
    title = "Future Lens: Anticipating Subsequent Tokens from a Single Hidden State",
    author = "Pal, Koyena  and
      Sun, Jiuding  and
      Yuan, Andrew  and
      Wallace, Byron  and
      Bau, David",
    editor = "Jiang, Jing  and
      Reitter, David  and
      Deng, Shumin",
    booktitle = "Proceedings of the 27th Conference on Computational Natural Language Learning (CoNLL)",
    month = dec,
    year = "2023",
    address = "Singapore",
    publisher = "Association for Computational Linguistics",
    url = "https://aclanthology.org/2023.conll-1.37/",
    doi = "10.18653/v1/2023.conll-1.37",
    pages = "548--560",
}

@inproceedings{
dong2025emergent,
title={Emergent Response Planning in {LLM}s},
author={Zhichen Dong and Zhanhui Zhou and Zhixuan Liu and Chao Yang and Chaochao Lu},
booktitle={Forty-second International Conference on Machine Learning},
year={2025},
url={https://openreview.net/forum?id=Ce79P8ULPY}
}

@inproceedings{huang-etal-2025-enhancing,
    title = "Enhancing Chain-of-Thought Reasoning with Critical Representation Fine-tuning",
    author = "Huang, Chenxi  and
      Yan, Shaotian  and
      Xie, Liang  and
      Lin, Binbin  and
      Fan, Sinan  and
      Xin, Yue  and
      Cai, Deng  and
      Shen, Chen  and
      Ye, Jieping",
    editor = "Che, Wanxiang  and
      Nabende, Joyce  and
      Shutova, Ekaterina  and
      Pilehvar, Mohammad Taher",
    booktitle = "Proceedings of the 63rd Annual Meeting of the Association for Computational Linguistics (Volume 1: Long Papers)",
    month = jul,
    year = "2025",
    address = "Vienna, Austria",
    publisher = "Association for Computational Linguistics",
    url = "https://aclanthology.org/2025.acl-long.1129/",
    doi = "10.18653/v1/2025.acl-long.1129",
    pages = "23173--23195",
    ISBN = "979-8-89176-251-0",
}

@inproceedings{
kang2025scalable,
title={Scalable Best-of-N Selection for Large Language Models via Self-Certainty},
author={Zhewei Kang and Xuandong Zhao and Dawn Song},
booktitle={The Thirty-ninth Annual Conference on Neural Information Processing Systems},
year={2025},
url={https://openreview.net/forum?id=29FRqmVQK8}
}

@inproceedings{
chen2024inside,
title={{INSIDE}: {LLM}s' Internal States Retain the Power of Hallucination Detection},
author={Chao Chen and Kai Liu and Ze Chen and Yi Gu and Yue Wu and Mingyuan Tao and Zhihang Fu and Jieping Ye},
booktitle={The Twelfth International Conference on Learning Representations},
year={2024},
url={https://openreview.net/forum?id=Zj12nzlQbz}
}

@inproceedings{
lin2025zebralogic,
title={ZebraLogic: On the Scaling Limits of {LLM}s for Logical Reasoning},
author={Bill Yuchen Lin and Ronan Le Bras and Kyle Richardson and Ashish Sabharwal and Radha Poovendran and Peter Clark and Yejin Choi},
booktitle={Forty-second International Conference on Machine Learning},
year={2025},
url={https://openreview.net/forum?id=sTAJ9QyA6l}
}

@inproceedings{pang-etal-2022-quality,
    title = "{Q}u{ALITY}: Question Answering with Long Input Texts, Yes!",
    author = "Pang, Richard Yuanzhe  and
      Parrish, Alicia  and
      Joshi, Nitish  and
      Nangia, Nikita  and
      Phang, Jason  and
      Chen, Angelica  and
      Padmakumar, Vishakh  and
      Ma, Johnny  and
      Thompson, Jana  and
      He, He  and
      Bowman, Samuel",
    editor = "Carpuat, Marine  and
      de Marneffe, Marie-Catherine  and
      Meza Ruiz, Ivan Vladimir",
    booktitle = "Proceedings of the 2022 Conference of the North American Chapter of the Association for Computational Linguistics: Human Language Technologies",
    month = jul,
    year = "2022",
    address = "Seattle, United States",
    publisher = "Association for Computational Linguistics",
    url = "https://aclanthology.org/2022.naacl-main.391/",
    doi = "10.18653/v1/2022.naacl-main.391",
    pages = "5336--5358",
}

@inproceedings{
mmlu,
title={Measuring Massive Multitask Language Understanding},
author={Dan Hendrycks and Collin Burns and Steven Basart and Andy Zou and Mantas Mazeika and Dawn Song and Jacob Steinhardt},
booktitle={International Conference on Learning Representations},
year={2021},
url={https://openreview.net/forum?id=d7KBjmI3GmQ}
}

@inproceedings{
hendrycks2021measuring,
title={Measuring Mathematical Problem Solving With the {MATH} Dataset},
author={Dan Hendrycks and Collin Burns and Saurav Kadavath and Akul Arora and Steven Basart and Eric Tang and Dawn Song and Jacob Steinhardt},
booktitle={Thirty-fifth Conference on Neural Information Processing Systems Datasets and Benchmarks Track (Round 2)},
year={2021},
url={https://openreview.net/forum?id=7Bywt2mQsCe}
}

@inproceedings{
prm800k,
title={Let's Verify Step by Step},
author={Hunter Lightman and Vineet Kosaraju and Yuri Burda and Harrison Edwards and Bowen Baker and Teddy Lee and Jan Leike and John Schulman and Ilya Sutskever and Karl Cobbe},
booktitle={The Twelfth International Conference on Learning Representations},
year={2024},
url={https://openreview.net/forum?id=v8L0pN6EOi}
}

@misc{gsm8k,
      title={Training Verifiers to Solve Math Word Problems}, 
      author={Karl Cobbe and Vineet Kosaraju and Mohammad Bavarian and Mark Chen and Heewoo Jun and Lukasz Kaiser and Matthias Plappert and Jerry Tworek and Jacob Hilton and Reiichiro Nakano and Christopher Hesse and John Schulman},
      year={2021},
      eprint={2110.14168},
      archivePrefix={arXiv},
      primaryClass={cs.LG},
      url={https://arxiv.org/abs/2110.14168}, 
}

@inproceedings{talmor-etal-2019-commonsenseqa,
    title = "{C}ommonsense{QA}: A Question Answering Challenge Targeting Commonsense Knowledge",
    author = "Talmor, Alon  and
      Herzig, Jonathan  and
      Lourie, Nicholas  and
      Berant, Jonathan",
    editor = "Burstein, Jill  and
      Doran, Christy  and
      Solorio, Thamar",
    booktitle = "Proceedings of the 2019 Conference of the North {A}merican Chapter of the Association for Computational Linguistics: Human Language Technologies, Volume 1 (Long and Short Papers)",
    month = jun,
    year = "2019",
    publisher = "Association for Computational Linguistics",
    url = "https://aclanthology.org/N19-1421/",
    pages = "4149--4158",
}

@inproceedings{
sprague2024musr,
title={Mu{SR}: Testing the Limits of Chain-of-thought with Multistep Soft Reasoning},
author={Zayne Rea Sprague and Xi Ye and Kaj Bostrom and Swarat Chaudhuri and Greg Durrett},
booktitle={The Twelfth International Conference on Learning Representations},
year={2024},
url={https://openreview.net/forum?id=jenyYQzue1}
}

@inproceedings{
rein2024gpqa,
title={{GPQA}: A Graduate-Level Google-Proof Q\&A Benchmark},
author={David Rein and Betty Li Hou and Asa Cooper Stickland and Jackson Petty and Richard Yuanzhe Pang and Julien Dirani and Julian Michael and Samuel R. Bowman},
booktitle={First Conference on Language Modeling},
year={2024},
url={https://openreview.net/forum?id=Ti67584b98}
}

@misc{aime,
      title={{AIME} Problems and Solutions},
      author={{AIME}},
      year={2025},
      url={https://artofproblemsolving.com/wiki/index.php/AIME_Problems_and_Solutions}
}

@InProceedings{bottleneck-adapter,
  title = 	 {Parameter-Efficient Transfer Learning for {NLP}},
  author =       {Houlsby, Neil and Giurgiu, Andrei and Jastrzebski, Stanislaw and Morrone, Bruna and De Laroussilhe, Quentin and Gesmundo, Andrea and Attariyan, Mona and Gelly, Sylvain},
  booktitle = 	 {Proceedings of the 36th International Conference on Machine Learning},
  pages = 	 {2790--2799},
  year = 	 {2019},
  editor = 	 {Chaudhuri, Kamalika and Salakhutdinov, Ruslan},
  volume = 	 {97},
  series = 	 {Proceedings of Machine Learning Research},
  month = 	 {09--15 Jun},
  publisher =    {PMLR},
  url = 	 {https://proceedings.mlr.press/v97/houlsby19a.html},
}

@misc{shao2024deepseekmath,
      title={DeepSeekMath: Pushing the Limits of Mathematical Reasoning in Open Language Models}, 
      author={Zhihong Shao and Peiyi Wang and Qihao Zhu and Runxin Xu and Junxiao Song and Xiao Bi and Haowei Zhang and Mingchuan Zhang and Y. K. Li and Y. Wu and Daya Guo},
      year={2024},
      eprint={2402.03300},
      archivePrefix={arXiv},
      primaryClass={cs.CL},
      url={https://arxiv.org/abs/2402.03300}, 
}

@misc{chen2025deephiddencognitionfacilitates,
      title={Deep Hidden Cognition Facilitates Reliable Chain-of-Thought Reasoning}, 
      author={Zijun Chen and Wenbo Hu and Richang Hong},
      year={2026},
      month={January}, 
booktitle={Proceedings of the AAAI Conference on Artificial Intelligence}, 
url={https://arxiv.org/abs/2507.10007}, 
}

@misc{zhang2025reasoningmeetslaws,
      title={When Reasoning Meets Its Laws}, 
      author={Junyu Zhang and Yifan Sun and Tianang Leng and Jingyan Shen and Liu Ziyin and Paul Pu Liang and Huan Zhang},
      year={2025},
      eprint={2512.17901},
      archivePrefix={arXiv},
      primaryClass={cs.AI},
      url={https://arxiv.org/abs/2512.17901}, 
}

@misc{singh2026llmsencodefunctionalimportance,
      title={Do LLMs Encode Functional Importance of Reasoning Tokens?}, 
      author={Janvijay Singh and Dilek Hakkani-Tür},
      year={2026},
      eprint={2601.03066},
      archivePrefix={arXiv},
      primaryClass={cs.CL},
      url={https://arxiv.org/abs/2601.03066}, 
}

@inproceedings{
bigelow2025forking,
title={Forking Paths in Neural Text Generation},
author={Eric J Bigelow and Ari Holtzman and Hidenori Tanaka and Tomer Ullman},
booktitle={The Thirteenth International Conference on Learning Representations},
year={2025},
url={https://openreview.net/forum?id=8RCmNLeeXx}
}

@inproceedings{afzal-etal-2025-knowing,
    title = "Knowing Before Saying: {LLM} Representations Encode Information About Chain-of-Thought Success Before Completion",
    author = "Afzal, Anum  and
      Matthes, Florian  and
      Chechik, Gal  and
      Ziser, Yftah",
    editor = "Che, Wanxiang  and
      Nabende, Joyce  and
      Shutova, Ekaterina  and
      Pilehvar, Mohammad Taher",
    booktitle = "Findings of the Association for Computational Linguistics: ACL 2025",
    month = jul,
    year = "2025",
    address = "Vienna, Austria",
    publisher = "Association for Computational Linguistics",
    url = "https://aclanthology.org/2025.findings-acl.662/",
    doi = "10.18653/v1/2025.findings-acl.662",
    pages = "12791--12806",
    ISBN = "979-8-89176-256-5",
}

@inproceedings{azaria-mitchell-2023-internal,
    title = "The Internal State of an {LLM} Knows When It{'}s Lying",
    author = "Azaria, Amos  and
      Mitchell, Tom",
    editor = "Bouamor, Houda  and
      Pino, Juan  and
      Bali, Kalika",
    booktitle = "Findings of the Association for Computational Linguistics: EMNLP 2023",
    month = dec,
    year = "2023",
    address = "Singapore",
    publisher = "Association for Computational Linguistics",
    url = "https://aclanthology.org/2023.findings-emnlp.68/",
    doi = "10.18653/v1/2023.findings-emnlp.68",
    pages = "967--976",
}

@inproceedings{gottesman-geva-2024-estimating,
    title = "Estimating Knowledge in Large Language Models Without Generating a Single Token",
    author = "Gottesman, Daniela  and
      Geva, Mor",
    editor = "Al-Onaizan, Yaser  and
      Bansal, Mohit  and
      Chen, Yun-Nung",
    booktitle = "Proceedings of the 2024 Conference on Empirical Methods in Natural Language Processing",
    month = nov,
    year = "2024",
    address = "Miami, Florida, USA",
    publisher = "Association for Computational Linguistics",
    url = "https://aclanthology.org/2024.emnlp-main.232/",
    doi = "10.18653/v1/2024.emnlp-main.232",
    pages = "3994--4019"
}

@inproceedings{liu-etal-2024-enhancing-language,
    title = "Enhancing Language Model Factuality via Activation-Based Confidence Calibration and Guided Decoding",
    author = "Liu, Xin  and
      Fatahi Bayat, Farima  and
      Wang, Lu",
    editor = "Al-Onaizan, Yaser  and
      Bansal, Mohit  and
      Chen, Yun-Nung",
    booktitle = "Proceedings of the 2024 Conference on Empirical Methods in Natural Language Processing",
    month = nov,
    year = "2024",
    address = "Miami, Florida, USA",
    publisher = "Association for Computational Linguistics",
    url = "https://aclanthology.org/2024.emnlp-main.583/",
    doi = "10.18653/v1/2024.emnlp-main.583",
    pages = "10436--10448",
}

@inproceedings{
cot,
title={Chain of Thought Prompting Elicits Reasoning in Large Language Models},
author={Jason Wei and Xuezhi Wang and Dale Schuurmans and Maarten Bosma and brian ichter and Fei Xia and Ed H. Chi and Quoc V Le and Denny Zhou},
booktitle={Advances in Neural Information Processing Systems},
editor={Alice H. Oh and Alekh Agarwal and Danielle Belgrave and Kyunghyun Cho},
year={2022},
url={https://openreview.net/forum?id=_VjQlMeSB_J}
}

@misc{cot21,
      title={Show Your Work: Scratchpads for Intermediate Computation with Language Models}, 
      author={Maxwell Nye and Anders Johan Andreassen and Guy Gur-Ari and Henryk Michalewski and Jacob Austin and David Bieber and David Dohan and Aitor Lewkowycz and Maarten Bosma and David Luan and Charles Sutton and Augustus Odena},
      year={2021},
      eprint={2112.00114},
      archivePrefix={arXiv},
      primaryClass={cs.LG},
      url={https://arxiv.org/abs/2112.00114}, 
}

@inproceedings{
wang2025planning,
title={Planning in Natural Language Improves {LLM} Search for Code Generation},
author={Evan Z Wang and Federico Cassano and Catherine Wu and Yunfeng Bai and William Song and Vaskar Nath and Ziwen Han and Sean M. Hendryx and Summer Yue and Hugh Zhang},
booktitle={The Thirteenth International Conference on Learning Representations},
year={2025},
url={https://openreview.net/forum?id=48WAZhwHHw}
}

@inproceedings{wang-etal-2023-plan,
    title = "Plan-and-Solve Prompting: Improving Zero-Shot Chain-of-Thought Reasoning by Large Language Models",
    author = "Wang, Lei  and
      Xu, Wanyu  and
      Lan, Yihuai  and
      Hu, Zhiqiang  and
      Lan, Yunshi  and
      Lee, Roy Ka-Wei  and
      Lim, Ee-Peng",
    editor = "Rogers, Anna  and
      Boyd-Graber, Jordan  and
      Okazaki, Naoaki",
    booktitle = "Proceedings of the 61st Annual Meeting of the Association for Computational Linguistics (Volume 1: Long Papers)",
    month = jul,
    year = "2023",
    address = "Toronto, Canada",
    publisher = "Association for Computational Linguistics",
    url = "https://aclanthology.org/2023.acl-long.147/",
    doi = "10.18653/v1/2023.acl-long.147",
    pages = "2609--2634",
}

@misc{roberta,
      title={RoBERTa: A Robustly Optimized BERT Pretraining Approach}, 
      author={Yinhan Liu and Myle Ott and Naman Goyal and Jingfei Du and Mandar Joshi and Danqi Chen and Omer Levy and Mike Lewis and Luke Zettlemoyer and Veselin Stoyanov},
      year={2019},
      eprint={1907.11692},
      archivePrefix={arXiv},
      primaryClass={cs.CL},
      url={https://arxiv.org/abs/1907.11692}, 
}

@inproceedings{
zhou2023large,
title={Large Language Models are Human-Level Prompt Engineers},
author={Yongchao Zhou and Andrei Ioan Muresanu and Ziwen Han and Keiran Paster and Silviu Pitis and Harris Chan and Jimmy Ba},
booktitle={The Eleventh International Conference on Learning Representations },
year={2023},
url={https://openreview.net/forum?id=92gvk82DE-}
}

@misc{sahoo2025systematic,
      title={A Systematic Survey of Prompt Engineering in Large Language Models: Techniques and Applications}, 
      author={Pranab Sahoo and Ayush Kumar Singh and Sriparna Saha and Vinija Jain and Samrat Mondal and Aman Chadha},
      year={2025},
      eprint={2402.07927},
      archivePrefix={arXiv},
      primaryClass={cs.AI},
      url={https://arxiv.org/abs/2402.07927}, 
}

@inproceedings{dong-etal-2024-survey,
    title = "A Survey on In-context Learning",
    author = "Dong, Qingxiu  and
      Li, Lei  and
      Dai, Damai  and
      Zheng, Ce  and
      Ma, Jingyuan  and
      Li, Rui  and
      Xia, Heming  and
      Xu, Jingjing  and
      Wu, Zhiyong  and
      Chang, Baobao  and
      Sun, Xu  and
      Li, Lei  and
      Sui, Zhifang",
    editor = "Al-Onaizan, Yaser  and
      Bansal, Mohit  and
      Chen, Yun-Nung",
    booktitle = "Proceedings of the 2024 Conference on Empirical Methods in Natural Language Processing",
    month = nov,
    year = "2024",
    address = "Miami, Florida, USA",
    publisher = "Association for Computational Linguistics",
    url = "https://aclanthology.org/2024.emnlp-main.64/",
    doi = "10.18653/v1/2024.emnlp-main.64",
    pages = "1107--1128",
}

@article{merrill-sabharwal-2023-parallelism,
    title = "The Parallelism Tradeoff: Limitations of Log-Precision Transformers",
    author = "Merrill, William  and
      Sabharwal, Ashish",
    journal = "Transactions of the Association for Computational Linguistics",
    volume = "11",
    year = "2023",
    address = "Cambridge, MA",
    publisher = "MIT Press",
    url = "https://aclanthology.org/2023.tacl-1.31/",
    doi = "10.1162/tacl_a_00562",
    pages = "531--545",
}

@inproceedings{hahn-rofin-2024-sensitive,
    title = "Why are Sensitive Functions Hard for Transformers?",
    author = "Hahn, Michael  and
      Rofin, Mark",
    editor = "Ku, Lun-Wei  and
      Martins, Andre  and
      Srikumar, Vivek",
    booktitle = "Proceedings of the 62nd Annual Meeting of the Association for Computational Linguistics (Volume 1: Long Papers)",
    month = aug,
    year = "2024",
    address = "Bangkok, Thailand",
    publisher = "Association for Computational Linguistics",
    url = "https://aclanthology.org/2024.acl-long.800/",
    doi = "10.18653/v1/2024.acl-long.800",
    pages = "14973--15008",
}

@inproceedings{chiang-cholak-2022-overcoming,
    title = "Overcoming a Theoretical Limitation of Self-Attention",
    author = "Chiang, David  and
      Cholak, Peter",
    editor = "Muresan, Smaranda  and
      Nakov, Preslav  and
      Villavicencio, Aline",
    booktitle = "Proceedings of the 60th Annual Meeting of the Association for Computational Linguistics (Volume 1: Long Papers)",
    month = may,
    year = "2022",
    address = "Dublin, Ireland",
    publisher = "Association for Computational Linguistics",
    url = "https://aclanthology.org/2022.acl-long.527/",
    doi = "10.18653/v1/2022.acl-long.527",
    pages = "7654--7664",
}

@misc{huang2024surveyuncertaintyestimationllms,
      title={A Survey of Uncertainty Estimation in LLMs: Theory Meets Practice}, 
      author={Hsiu-Yuan Huang and Yutong Yang and Zhaoxi Zhang and Sanwoo Lee and Yunfang Wu},
      year={2024},
      eprint={2410.15326},
      archivePrefix={arXiv},
      primaryClass={cs.CL},
      url={https://arxiv.org/abs/2410.15326}, 
}

@inproceedings{bakman-etal-2025-reconsidering,
    title = "Reconsidering {LLM} Uncertainty Estimation Methods in the Wild",
    author = "Bakman, Yavuz Faruk  and
      Yaldiz, Duygu Nur  and
      Kang, Sungmin  and
      Zhang, Tuo  and
      Buyukates, Baturalp  and
      Avestimehr, Salman  and
      Karimireddy, Sai Praneeth",
    editor = "Che, Wanxiang  and
      Nabende, Joyce  and
      Shutova, Ekaterina  and
      Pilehvar, Mohammad Taher",
    booktitle = "Proceedings of the 63rd Annual Meeting of the Association for Computational Linguistics (Volume 1: Long Papers)",
    month = jul,
    year = "2025",
    address = "Vienna, Austria",
    publisher = "Association for Computational Linguistics",
    url = "https://aclanthology.org/2025.acl-long.1429/",
    doi = "10.18653/v1/2025.acl-long.1429",
    pages = "29531--29556",
    ISBN = "979-8-89176-251-0",
}

@inproceedings{NEURIPS2019_159c1ffe,
 author = {Reif, Emily and Yuan, Ann and Wattenberg, Martin and Viegas, Fernanda B and Coenen, Andy and Pearce, Adam and Kim, Been},
 booktitle = {Advances in Neural Information Processing Systems},
 editor = {H. Wallach and H. Larochelle and A. Beygelzimer and F. d\textquotesingle Alch\'{e}-Buc and E. Fox and R. Garnett},
 pages = {},
 publisher = {Curran Associates, Inc.},
 title = {Visualizing and Measuring the Geometry of BERT},
 url = {https://proceedings.neurips.cc/paper_files/paper/2019/file/159c1ffe5b61b41b3c4d8f4c2150f6c4-Paper.pdf},
 volume = {32},
 year = {2019}
}

@article{gari-soler-apidianaki-2021-lets,
    title = "Let{'}s Play Mono-Poly: {BERT} Can Reveal Words' Polysemy Level and Partitionability into Senses",
    author = "Gar{\'i} Soler, Aina  and
      Apidianaki, Marianna",
    editor = "Roark, Brian  and
      Nenkova, Ani",
    journal = "Transactions of the Association for Computational Linguistics",
    volume = "9",
    year = "2021",
    address = "Cambridge, MA",
    publisher = "MIT Press",
    url = "https://aclanthology.org/2021.tacl-1.50/",
    doi = "10.1162/tacl_a_00400",
    pages = "825--844",
}

@inproceedings{
skean2025layer,
title={Layer by Layer: Uncovering Hidden Representations in Language Models},
author={Oscar Skean and Md Rifat Arefin and Dan Zhao and Niket Nikul Patel and Jalal Naghiyev and Yann LeCun and Ravid Shwartz-Ziv},
booktitle={Forty-second International Conference on Machine Learning},
year={2025},
url={https://openreview.net/forum?id=WGXb7UdvTX}
}

@inproceedings{
li2026making,
title={Making Slow Thinking Faster: Compressing {LLM} Chain-of-Thought via Step Entropy},
author={Zeju Li and Jianyuan Zhong and Ziyang Zheng and Xiangyu Wen and Zhijian Xu and Yingying Cheng and Fan Zhang and Qiang Xu},
booktitle={The Fourteenth International Conference on Learning Representations},
year={2026},
url={https://openreview.net/forum?id=cGLqQfS5wH}
}

@misc{dapo,
      title={DAPO: An Open-Source LLM Reinforcement Learning System at Scale}, 
      author={Qiying Yu and Zheng Zhang and Ruofei Zhu and Yufeng Yuan and Xiaochen Zuo and Yu Yue and Weinan Dai and Tiantian Fan and Gaohong Liu and Lingjun Liu and Xin Liu and Haibin Lin and Zhiqi Lin and Bole Ma and Guangming Sheng and Yuxuan Tong and Chi Zhang and Mofan Zhang and Wang Zhang and Hang Zhu and Jinhua Zhu and Jiaze Chen and Jiangjie Chen and Chengyi Wang and Hongli Yu and Yuxuan Song and Xiangpeng Wei and Hao Zhou and Jingjing Liu and Wei-Ying Ma and Ya-Qin Zhang and Lin Yan and Mu Qiao and Yonghui Wu and Mingxuan Wang},
      year={2025},
      eprint={2503.14476},
      archivePrefix={arXiv},
      primaryClass={cs.LG},
      url={https://arxiv.org/abs/2503.14476}, 
}

@inproceedings{
ton2025understanding,
title={Understanding Chain-of-Thought in {LLM}s through Information Theory},
author={Jean-Francois Ton and Muhammad Faaiz Taufiq and Yang Liu},
booktitle={Forty-second International Conference on Machine Learning},
year={2025},
url={https://openreview.net/forum?id=IjOWms0hrf}
}

@inproceedings{
sheng2026on,
title={On Reasoning Strength Planning in Large Reasoning Models},
author={Leheng Sheng and An Zhang and Zijian Wu and Weixiang Zhao and Changshuo Shen and Yi Zhang and Xiang Wang and Tat-Seng Chua},
booktitle={The Thirty-ninth Annual Conference on Neural Information Processing Systems},
year={2025},
url={https://openreview.net/forum?id=H26A7cl91u}
}

@misc{shao2025continuous,
      title={Continuous Autoregressive Language Models}, 
      author={Chenze Shao and Darren Li and Fandong Meng and Jie Zhou},
      year={2025},
      eprint={2510.27688},
      archivePrefix={arXiv},
      primaryClass={cs.CL},
      url={https://arxiv.org/abs/2510.27688}, 
}

@misc{wang2026latent,
      title={Latent Chain-of-Thought as Planning: Decoupling Reasoning from Verbalization}, 
      author={Jiecong Wang and Hao Peng and Chunyang Liu},
      year={2026},
      eprint={2601.21358},
      archivePrefix={arXiv},
      primaryClass={cs.AI},
      url={https://arxiv.org/abs/2601.21358}, 
}

@inproceedings{
patel2022mapping,
title={Mapping Language Models to Grounded Conceptual Spaces},
author={Roma Patel and Ellie Pavlick},
booktitle={International Conference on Learning Representations},
year={2022},
url={https://openreview.net/forum?id=gJcEM8sxHK}
}

@inproceedings{
yong2026think,
title={Think or Not? Exploring Thinking Efficiency in Large Reasoning Models via an Information-Theoretic Lens},
author={Xixian Yong and Xiao Zhou and Yingying Zhang and Jinlin Li and Yefeng Zheng and Xian Wu},
booktitle={The Thirty-ninth Annual Conference on Neural Information Processing Systems},
year={2025},
url={https://openreview.net/forum?id=DpOSndSOZz}
}

@inproceedings{
yang2026dynamic,
title={Dynamic Early Exit in Reasoning Models},
author={Chenxu Yang and Qingyi Si and Yongjie Duan and Zheliang Zhu and Chenyu Zhu and Qiaowei Li and Minghui Chen and Zheng Lin and Weiping Wang},
booktitle={The Fourteenth International Conference on Learning Representations},
year={2026},
url={https://openreview.net/forum?id=NpU7ZXafRi}
}

@inproceedings{
kapoor2024large,
title={Large Language Models Must Be Taught to Know What They Don{\textquoteright}t Know},
author={Sanyam Kapoor and Nate Gruver and Manley Roberts and Katherine M. Collins and Arka Pal and Umang Bhatt and Adrian Weller and Samuel Dooley and Micah Goldblum and Andrew Gordon Wilson},
booktitle={The Thirty-eighth Annual Conference on Neural Information Processing Systems},
year={2024},
url={https://openreview.net/forum?id=QzvWyggrYB}
}

@inproceedings{
li2026efficient,
title={Efficient Reasoning with Balanced Thinking},
author={Yulin Li and Tengyao Tu and Li Ding and Junjie Wang and Hui-Ling Zhen and Yixin Chen and Yong Li and Zhuotao Tian},
booktitle={The Fourteenth International Conference on Learning Representations},
year={2026},
url={https://openreview.net/forum?id=cJseWJJ5IM}
}

@inproceedings{
ghandeharioun2024patchscopes,
title={Patchscopes: A Unifying Framework for Inspecting Hidden Representations of Language Models},
author={Asma Ghandeharioun and Avi Caciularu and Adam Pearce and Lucas Dixon and Mor Geva},
booktitle={Forty-first International Conference on Machine Learning},
year={2024},
url={https://openreview.net/forum?id=5uwBzcn885}
}
\bibliographystyle{icml2026}

%\newpage
\clearpage
\appendix

% \onecolumn
% \section{Related Works}
% \label{app:related}

\section{Tasks and Datasets}
\label{app:task}

As described in \cref{ssec:tasks}, our probing experiments span 12 diverse tasks of different types for a comprehensive view of empirical insights. This section further provides concrete examples, data processing details and statistics.

\subsection{Task Examples}
\label{app:task-example}

\begin{tcolorbox}[breakable,
title=Task Example for: Parity]

\small{\texttt{[Settings]:
\begin{itemize}[noitemsep,nolistsep]
\item Sequence length: 41
\item Target digit to count: 2
\item Answer: even
\end{itemize}
}}
\tcbline

Determine whether the number of ``2'' in the following digit sequence is even or odd; please output only your decision by either ``even'' or ``odd''.
\vspace{\baselineskip}

91223822122515222430601862928242722242251
\end{tcolorbox}

\begin{tcolorbox}[breakable,
title=Task Example for: Cycle]

\small{\texttt{[Settings]:
\begin{itemize}[noitemsep,nolistsep]
\item Number of edges: 16
\item Answer: NO
\end{itemize}
}}
\tcbline

\textbf{Task}

Given the following directed graph represented as a list of edges (from\_vertex $\rightarrow$ to\_vertex), along with two target vertices, you need to determine whether there exists a path from the first target vertex to the second.
\vspace{\baselineskip}

\textbf{Edges}

v\-453 $\rightarrow$ v\-561\\
v\-666 $\rightarrow$ v\-34\\
v\-34 $\rightarrow$ v\-791\\
v\-791 $\rightarrow$ v\-17\\
v\-416 $\rightarrow$ v\-0\\
v\-658 $\rightarrow$ v\-666\\
v\-0 $\rightarrow$ v\-74\\
v\-254 $\rightarrow$ v\-427\\
v\-427 $\rightarrow$ v\-520\\
v\-561 $\rightarrow$ v\-254\\
v\-74 $\rightarrow$ v\-453\\
v\-520 $\rightarrow$ v\-416\\
v\-664 $\rightarrow$ v\-464\\
v\-17 $\rightarrow$ v\-664\\
v\-640 $\rightarrow$ v\-658\\
v\-464 $\rightarrow$ v\-640\\

\textbf{Target}

v\_34, v\_561\\

\textbf{Output}

Please output only ``YES'' if a path exists, or ``NO'' if it does not.
\end{tcolorbox}

\begin{tcolorbox}[breakable,
title=Task Example for: Subsum]

\small{\texttt{[Settings]:
\begin{itemize}[noitemsep,nolistsep]
\item Sequence length: 29
\item Max subsequence sum: 84
\item Answer: 4
\end{itemize}
}}
\tcbline

Given the following sequence of numbers, determine the least significant digit of the maximum sum of its subsequences, such that no two numbers in the subsequence are adjacent in the original sequence. Please output only the according least significant digit directly.
\vspace{\baselineskip}

[2, 4, 6, 6, 1, 8, 5, 5, 4, 6, 6, 6, 6, 8, 1, 8, 9, 1, 9, 9, 4, 1, 9, 5, 4, 2, 4, 3, 2]
\end{tcolorbox}

\begin{tcolorbox}[breakable,
title=Task Example for: GSM8K (Multi-Choice)]

\small{\texttt{[Settings]:
\begin{itemize}[noitemsep,nolistsep]
\item Answer: D
\end{itemize}
}}
\tcbline

\textbf{Task}

Given the following problem along with its options, determine the best option as the answer. Please only output your selected answer option by the letter (e.g., A, B, C).
\vspace{\baselineskip}

\textbf{Problem}

Rob, Royce, and Pedro are contractors getting ready to put a new roof on three homes. If the three homes will need 250 cases of shingles, with the first house needing 1/2 of the second, and the third needing double the first. How many cases of shingles will the third house need?
\vspace{\baselineskip}

Options:\\
A. 125\\
B. 200\\
C. 50\\
D. 100\\
E. 83
\end{tcolorbox}

\begin{tcolorbox}[breakable,
title=Task Example for: MATH (Multi-Choice)]

\small{\texttt{[Settings]:
\begin{itemize}[noitemsep,nolistsep]
\item Answer: B
\end{itemize}
}}
\tcbline

\textbf{Task}

Given the following problem along with its options, determine the best option as the answer. Please only output your selected answer option by the letter (e.g., A, B, C).
\vspace{\baselineskip}

\textbf{Problem}

If each point of the circle $x^2 + y^2 = 25$ is reflected in the point $(4,1),$ the set of image points satisfies the equation
\[x^2 + ay^2 + bx + cy + d = 0.\]Compute the ordered quadruple $(a,b,c,d)$ of real numbers.
\vspace{\baselineskip}

Options:\\
A. (1,16,4,43)\\
B. (1,-16,-4,43)\\
C. (1,-8,-2,17)\\
D. (1,-16,4,43)\\
E. (1,-16,-4,-43)
\end{tcolorbox}

\begin{tcolorbox}[breakable,
title=Task Example for: AIME (Multi-Choice)]

\small{\texttt{[Settings]:
\begin{itemize}[noitemsep,nolistsep]
\item Answer: B
\end{itemize}
}}
\tcbline

\textbf{Task}

Given the following problem along with its options, determine the best option as the answer. Please only output your selected answer option by the letter (e.g., A, B, C).
\vspace{\baselineskip}

\textbf{Problem}

The set of points in 3-dimensional coordinate space that lie in the plane $x+y+z=75$ whose coordinates satisfy the inequalities $x-yz<y-zx<z-xy$ forms three disjoint convex regions. Exactly one of those regions has finite area. The area of this finite region can be expressed in the form $a\sqrt{b}$, where $a$ and $b$ are positive integers and $b$ is not divisible by the square of any prime. Find $a+b$.
\vspace{\baselineskip}

Options:\\
A. 524\\
B. 510\\
C. 498\\
D. 504\\
E. 496
\end{tcolorbox}

\begin{tcolorbox}[breakable,
title=Task Example for: MuSR]

\small{\texttt{[Settings]:
\begin{itemize}[noitemsep,nolistsep]
\item Answer: A
\end{itemize}
}}
\tcbline

\textbf{Task}

Given the following article, along with a related question and its answer options, please determine the best answer option for this question.
\vspace{\baselineskip}

\textbf{Article}

In my latest tenure at a bustling educational institution, three staff members, Emily, Robert, and Alice, consistently caught my attention amidst the sea of educators and support personnel. As the school manager, my role was to distribute tasks, specifically Teaching and Admin work, in a way that capitalized on each individual's unique strengths, thereby streamlining the school's operations. These assignments, as crucial as they were intricate, were akin to the individual notes in a symphony, each playing a vital role in the harmony of the institution.\\
Alice was a unique blend of complexities, as I carefully observed her interactions with the staff. Her proclivity for administrative tasks was evident, a much-needed quality in the heaving sea of paperwork the school generated. Alice often took the responsibility of capturing the minutes during our staff meetings and backed up Robert’s teachings with her painstaking administrative work...\\
...
\vspace{\baselineskip}

\textbf{Question}

Given the story, how would you uniquely allocate each person to make sure both tasks are accomplished efficiently?

A. Teaching: Emily, Admin work: Alice and Robert

B. Teaching: Alice, Admin work: Emily and Robert

C. Teaching: Robert, Admin work: Alice and Emily
\vspace{\baselineskip}

\textbf{Output}

Please only output your selected answer option by ``A/B/C/...''.
\end{tcolorbox}

\begin{tcolorbox}[breakable,
title=Task Example for: Zebra]

\small{\texttt{[Settings]:
\begin{itemize}[noitemsep,nolistsep]
\item Answer: A
\end{itemize}
}}
\tcbline

\textbf{Task}

Given the following problem along with its options, determine the best option as the answer. Please only output your selected answer option by the letter (e.g., A, B, C).
\vspace{\baselineskip}

\textbf{Problem}

There are 5 houses, numbered 1 to 5 from left to right, as seen from across the street. Each house is occupied by a different person. Each house has a unique attribute for each of the following characteristics:\\
 - Each person has a unique name: Peter, Alice, Arnold, Bob, Eric\\
 - Each person has a unique hobby: photography, cooking, knitting, gardening, painting\\
 - Each person has a unique favorite drink: root beer, milk, water, coffee, tea
\vspace{\baselineskip}

Rules:\\
1. Eric is the coffee drinker.\\
2. The tea drinker is the person who paints as a hobby.\\
3. The person who enjoys knitting is not in the fourth house.\\
4. Peter is not in the fourth house.\\
5. Eric is somewhere to the right of the root beer lover.\\
6. Arnold is the person who loves cooking.\\
7. The one who only drinks water is somewhere to the right of the person who enjoys gardening.\\
8. There is one house between Bob and the person who paints as a hobby.\\
9. The person who enjoys gardening is directly left of the root beer lover.\\
10. The photography enthusiast is the one who only drinks water.
\vspace{\baselineskip}

Question:\\
What is Drink of the person who lives in House 1?\\
A. milk\\
B. tea\\
C. root beer\\
D. coffee\\
E. water
\end{tcolorbox}

\begin{tcolorbox}[breakable,
title=Task Example for: CSQA]

\small{\texttt{[Settings]:
\begin{itemize}[noitemsep,nolistsep]
\item Answer: B
\end{itemize}
}}
\tcbline

\textbf{Task}

Given the following commonsense question, please determine its best answer option.
\vspace{\baselineskip}

\textbf{Question}

The drought was dangerous for the trees, they were more likely to what?\\
A. fire\\
B. burn\\
C. covered in snow\\
D. wall in\\
E. grow tall
\vspace{\baselineskip}

\textbf{Output}

Please only output your selected answer option by ``A/B/C/...''.
\end{tcolorbox}

\begin{tcolorbox}[breakable,
title=Task Example for: MMLU]

\small{\texttt{[Settings]:
\begin{itemize}[noitemsep,nolistsep]
\item Answer: D
\end{itemize}
}}
\tcbline

\textbf{Task}

Given the following question and its options, determine the best option as the answer. Please only output your selected answer option by ``A/B/C/D''.
\vspace{\baselineskip}

\textbf{Question}

A manager's competitor sent a defamatory letter to the manager accusing him of professional incompetence and calling him one of the worst businessmen in town. It was addressed to the manager. He read it, put it in a private drawer, and did not read it again. Later, he tried to sue the competitor for defamation as a result of the letter. Will the court likely grant the defendant's motion to dismiss, and on what grounds? Base your answer on the common law definition of defamation.
\vspace{\baselineskip}

A. No, it will not dismiss because a plaintiff in a defamatory action has an absolute right to a jury trial to prove defamation.\\
B. Yes, it will dismiss on the basis that the language is not damaging to the manager's reputation.\\
C. No, it will not dismiss because the circumstances show that all of the elements of defamation are all present.\\
D. Yes, it will dismiss on the basis that the publication is made to the manager alone.
\end{tcolorbox}

\begin{tcolorbox}[breakable,
title=Task Example for: QuALITY]

\small{\texttt{[Settings]:
\begin{itemize}[noitemsep,nolistsep]
\item Answer: B
\end{itemize}
}}
\tcbline

\textbf{Task}

Given the following snippets from an article or a story, along with a related question and its answer options, you need to determine the best option based on the information provided by these snippets.

These snippets may not necessarily always contain the supported information to answer the question; in that case try to give a best guess.
\vspace{\baselineskip}

\textbf{Snippets}

For his earlier errors, Coleman had first received a suspended sentence, then two terminal sentences to be fixed by the warden. My predecessors had given him first a few weeks, then a few months of sleep in Dreamland. Coleman's eyes didn't frighten me; I focused right on the pupils. ``That was a pretty foul trick, Councilman. Did you hope to somehow frighten me out of executing this sentence by what you told me this morning?''\\
I couldn't follow his reasoning. Just how making me think my life was only a Dream such as I imposed on my own prisoners could help him, I couldn't see...\\
...
\vspace{\baselineskip}

\textbf{Question}

What were Coleman’s motivations in visiting the warden?\\
A. Providing the warden with his annual raise announcement\\
B. Scaring him into believing his life was a dream\\
C. Gathering information to bring down the warden’s compound\\
D. Persuading the warden to step down from his position\\

\textbf{Output}

Please only output your selected answer option by ``A/B/C/D''.
\end{tcolorbox}

\begin{tcolorbox}[breakable,
title=Task Example for: GPQA]

\small{\texttt{[Settings]:
\begin{itemize}[noitemsep,nolistsep]
\item Answer: B
\end{itemize}
}}
\tcbline

\textbf{Task}

Given the following exam question, please determine its best answer option.
\vspace{\baselineskip}

\textbf{Question}

Observations of structures located at a distance of about 2.1 gigaparsecs (2.1 Gpc) are being carried out. The detected absorption line energy equivalent is about 3.9 micro electron volts ($3.9 * 10^{-6}$ eV).
\vspace{\baselineskip}

What is most likely to be observed with this absorption line in the Milky Way?\\
A. Warm atomic interstellar medium.\\
B. Cold atomic interstellar medium.\\
C. Warm molecular interstellar medium.\\
D. Cold molecular interstellar medium.
\vspace{\baselineskip}

\textbf{Output}

Please only output your selected answer option by ``A/B/C/D''.
\end{tcolorbox}

\subsection{Dataset Descriptions}
\label{app:dataset-desc}

A brief description of each existing dataset adopted in our experiments is provided below.

\paragraph{Implicit Compositional Tasks}
Three mathematical tasks and two logical reasoning tasks are included:

\begin{itemize}[noitemsep,nolistsep,leftmargin=1em]
\item \textbf{GSM8K}: a dataset focusing on middle school-level problems of various difficulties \cite{gsm8k}.
\item \textbf{MATH}: a dataset introduced by \citet{hendrycks2021measuring} with high school competition-level math problems. We follow the MATH-500 test split by \citet{prm800k}.
\item \textbf{AIME}: 30 math competition problems from AIME'25 (2025 American Invitational Mathematics Examination) \cite{aime}.
\item \textbf{MuSR}: a \underline{Mu}ltistep Soft Reasoning dataset to evaluate logical deduction with natural language rules over long, text-based narratives \cite{sprague2024musr}.
\item \textbf{Zebra}: the benchmark ZebraLogic \cite{lin2025zebralogic} designed to evaluate symbolic reasoning and constraint satisfaction abilities within a natural language context.
\end{itemize}

\paragraph{Knowledge and Semantic Tasks}
Four knowledge-intensive benchmarks focusing on semantic understanding rather than explicit reasoning are included:
\begin{itemize}[noitemsep,nolistsep,leftmargin=1em]
\item \textbf{CSQA}: CommonsenseQA \cite{talmor-etal-2019-commonsenseqa}, targeting commonsense reasoning based on world knowledge.
\item \textbf{MMLU}: a broad-spectrum dataset to evaluate knowledge from 57 subjects, covering various STEM and social science domains \cite{mmlu}.
\item \textbf{QuALITY}: a narrative question answering dataset \cite{pang-etal-2022-quality}. To reduce computational overhead, we frame the context in the form of relevant snippets retrieved via dense retrieval (a RAG setting with a max 2K context length), rather than using full documents.
\item \textbf{GPQA}: a challenging dataset designed to test expert-level knowledge of multiple domains, such as biology, physics, chemistry, etc. \cite{rein2024gpqa}.
\end{itemize}

\subsection{Data Preparation For Existing Datasets}
\label{app:data-existing}

Among the 12 tasks used in our probing experiments, three mathematics tasks (MATH, GSM8K, and AIME) are originally evaluated via free-form generation, without a fixed answer space. To enable final-answer probing, we use the following prompt to convert each problem into a multiple-choice format using GPT-4.1.

\begin{tcolorbox}[breakable,
title=Prompt for Multi-Choice Conversion]
\small
\textbf{Task}

Given the following problem and its correct answer solution, you need to generate four plausible but incorrect answer options, which will serve as misleading distractors to construct multiple-choice questions.
\vspace{\baselineskip}

\textbf{Problem}

\{problem\}
\vspace{\baselineskip}

\textbf{Solution} [Optional]

\{solution\}
\vspace{\baselineskip}

\textbf{Correct Answer}

\{answer\}
\vspace{\baselineskip}

\textbf{Output}

Please first think about four misleading wrong answer options, starting with ``\#\#\# Think''.\\
Then, starting with ``\#\#\# Options'', provide each option per line, where each line is directly the according answer option in a similar format of the correct answer, without adding any prefix or explanation.
\end{tcolorbox}

For other existing datasets originally in a multiple-choice format, we also shuffle the order of options for each question, intended to mitigate potential memorization effects and positional bias by LLMs.

With all 12 tasks having a fixed answer space, the label set for the final-answer probing consists of 20 tokens in total:

\[\begin{Bmatrix}
\text{A, B, C, D, E, F, YES, NO, even, odd,}\\
\text{0, 1, 2, 3, 4, 5, 6, 7, 8, 9}
\end{Bmatrix}\]

\subsection{Data Generation and Sampling}
\label{app:data-generation}

The data generation process for the three explicit compositional tasks is fully controllable. For probing, we generate problems and their corresponding labels for each task using the following procedure.

\paragraph{Parity}
We generate random digit sequences from length 5 to 100. For each sequence, the target digit to count is randomly selected from $\{1,2,7,8\}$. The label is then determined by the parity of its count.

\paragraph{Cycle}
For each input, we first set the number of edges, an even number randomly determined from 4 to 100. Two instances are yielded at each time, one forming a full cycle using all edges, and another forming two equal-sized cycles each using half of the edges. We then randomly assign vertex names from a pool of 1000 candidates to each cycle, ensuring diversity in both vertex identities and edge orderings. 
For the target vertices to determine whether there exists a path in between, we randomly select two vertices in the first case (there is always a path due to cycling), and randomly select one vertex from each cycle for the second case (no path exists).

\paragraph{Subsum}
We generate random lists of integers (each in the range from 1 to 9), with lengths ranging from 2 to 50. The labels are obtained by applying the dynamic programming solution to each list.

With our data generation process, the resulting labels for each task are evenly distributed.
Our code for data generation and the resulting datasets will be publicly released.

\paragraph{Data Sampling}
For other tasks, we sample from their original test sets, such as our test split always consists of problems drawn from the original test sets. For our train/dev splits, we keep sampling remaining problems from the test sets. If a dataset does not contain a sufficient number of test instances, we additionally sample from their original training and dev sets, if they are available.

Note that AIME’25 contains only 30 problems in total. Following the above procedure, all 30 problems are included in our test split, with no problems added to our train or dev split for AIME.

\subsection{Dataset Construction and Statistics}
\label{app:stats}

With our train/dev/test splits in place, we perform inference by each of the two LLM backbones, collecting their corresponding rollouts and hidden states.
We retain hidden states of all tokens along CoT trajectories within a maximum length of 16,384 for the test split.
For the train/dev splits, to maintain a manageable storage cost, we keep hidden states of sampled 5\% / 10\% CoT tokens for Off-the-Shelf LLM and In-Domain LLM respectively.

The resulting train/dev/test dataset has 2.4M / 81K / 11M hidden states for Off-the-Shelf LLM, and 2.5M / 57K / 2.7M hidden states for In-Domain LLM, respectively (for each Transformers layer).

Each hidden state is labeled according to the corresponding rollout outcomes for each teleological dimension, e.g. the tokens ID for the $i$-th subsequent token, the token ID of the final predicted answer, the total CoT length, etc.
Given the ample number of instances in our dataset, our \probe adapters can automatically learn latent features that discriminate among different labels.

\subsection{Hyperparameters}
\label{app:hyper}

The hidden size is $d=5120$ for Off-the-Shelf LLM (Qwen3-32B), and $d=3584$ for In-Domain LLM that is trained upon Qwen2.5-7B-Instruct. We use $r=256$ for both models, which is shown sufficient by prior works on probing \cite{dong2025emergent}. For training, we adopt learning rate $1 \times 10^{-3}$, batch size $8000$, a linear decay learning rate schedule, and early stopping on dev set, with approximately $5000$ max training steps; we do not enable weight decay or warmup period.
The training is conducted on Nvidia V100 GPUs, and only the parameters of \probe are updated; the LM head is kept frozen during training. 

\section{In-Domain LLM}
\label{app:rl}

\subsection{Training Details}

As described in \cref{ssec:llm}, we conduct reinforcement learning with GRPO \cite{shao2024deepseekmath} on Qwen2.5-7B-Instruct to obtain an In-domain LLM on the 12 tasks.
During training, we add the format instruction in the LLM system prompt, and use a reward signal based solely on format validation and answer correctness (each with score $1$).
The training set comprises 48K problems, sampled from original training sets of those existing datasets, as well as auto-generated problems for explicit compositional tasks.

\begin{tcolorbox}[breakable,
title=System Prompt for: In-Domain LLM]
\fontsize{7}{\baselineskip}\texttt{You are a helpful assistant. Now the user asks you to solve a reasoning problem. You need to first think about the solving process in the mind and then provide the user with the answer. The thinking process is enclosed within $<$think$>$ $<$/think$>$ tags, i.e., $<$think$>$ thinking process here $<$/think$>$ final answer.}
\end{tcolorbox}

For training, we use a rollout size $16$, batch size $320$ with mini batch size $80$, capping the max response length as $4096$. We adopt a cosine learning rate schedule with initial learning rate $1 \times 10^{-6}$ and $10$ warmup steps. Following DAPO \cite{dapo}, we adopt the clip-higher strategy to encourage exploration, setting the upper clip ratio as $0.3$ and the lower clip ratio as $0.2$. The training converges within 800 steps, among which Parity is the slowest to converge.

\subsection{Evaluation}
\label{app:llm-eval}

Evaluation of our adopted LLM backbones across 12 tasks, including our In-Domain LLM trained by reinforcement learning, are provided in \cref{tab:perf}.

\subsection{In-Domain CoT Examples}
\label{app:cot-example}

\begin{figure}[!t]
\centering
\includegraphics[width=\columnwidth]{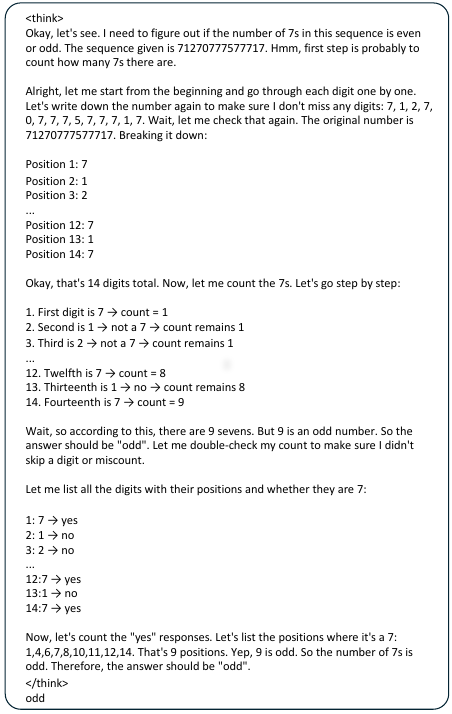}
\caption{Example of Parity Response with Off-the-Shelf LLM (Qwen3-32B). Full evaluation is discussed in \cref{ssec:results}.}
\label{fig:example-32b}
% \vspace{-0.6em}
\end{figure}

\begin{figure}[!t]
\centering
\includegraphics[width=0.9\columnwidth]{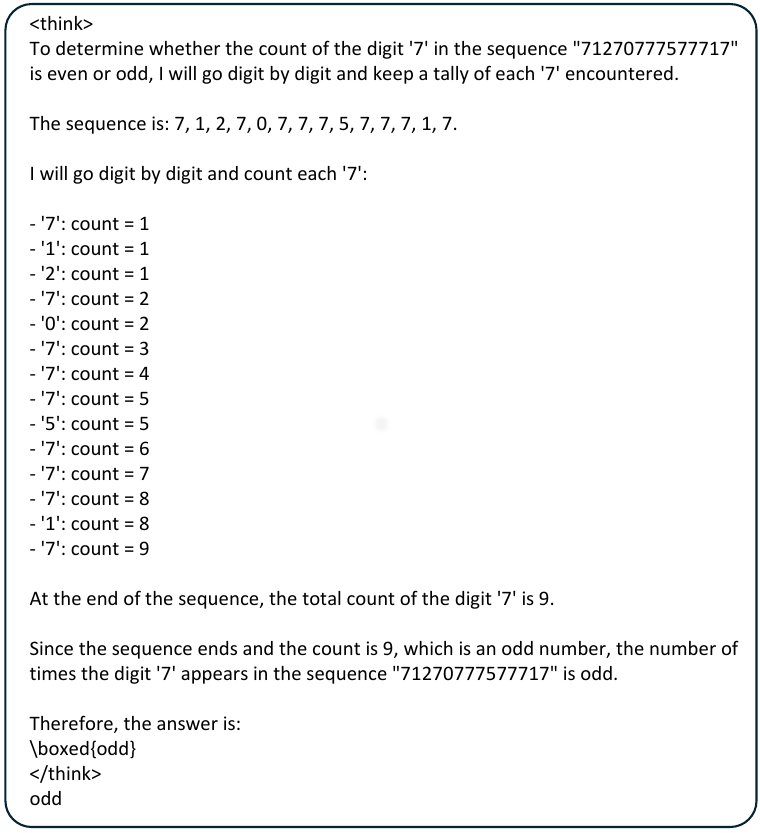}
\caption{Example of Parity Response with our In-Domain LLM trained via GRPO. The resulting reasoning trajectory is much shorter with predictable patterns, as discussed in \cref{ssec:result-subsequent}.}
\label{fig:example-id}
% \vspace{-0.6em}
\end{figure}

\begin{figure}[!t]
\centering
\includegraphics[width=0.9\columnwidth]{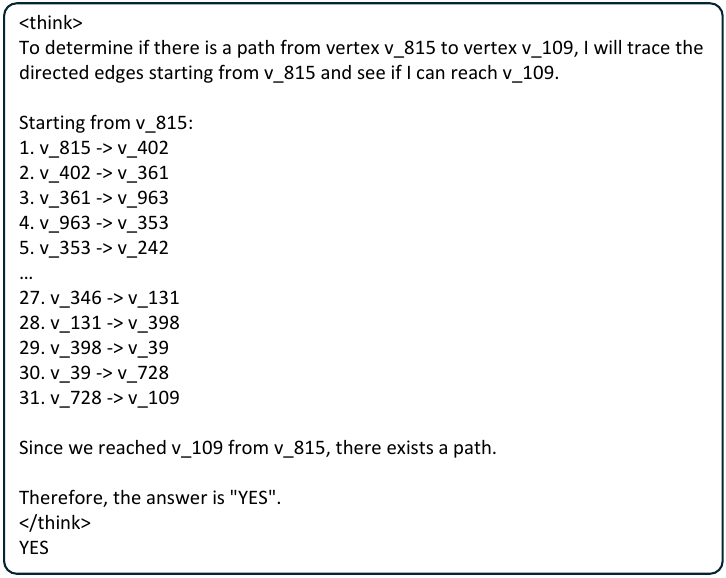}
\caption{Example of Cycle Response with our In-Domain LLM. The length of the reasoning trajectory is in proportional to the length of the path/cycle between two vertices, but not to the number of input edges. Without the latter heuristics, LLM is not able to reliably predict the total reasoning length at the initial stage of CoT, as discussed in \cref{ssec:result-step}.}
\label{fig:example-cycle-id}
\vspace{-0.6em}
\end{figure}

As shown in \cref{tab:perf}, our In-Domain LLM produces much shorter CoT trajectories, indicating more stable and decisive reasoning behavior. For qualitative comparison, we provide examples for Parity using both Off-the-Shelf LLM (Qwen3-32B) and In-Domain LLM, shown in \cref{fig:example-32b} and \cref{fig:example-id}. Another example for Cycle is shown in \cref{fig:example-cycle-id}.

\begin{table*}[tbp!]
\centering
\caption{Accuracy of LLM backbones on the 12 tasks spanning three categories described in \cref{ssec:tasks}, along with CoT length (measured in number of characters), averaged from 5 repeated runs. For the off-the-shelf Qwen3 LLMs, we evaluate two settings, with thinking mode disabled (w/o CoT) or enabled (w/ CoT), respectively. Our In-Domain LLM is trained by GRPO upon Qwen2.5-7B-Instruct (details in \cref{ssec:llm}), which is one generation behind the Qwen3 series, thus its performance may lag behind on certain datasets. Despite this, it achieves the best performance on three compositional tasks while producing \textbf{substantially shorter CoT trajectories}. Note that in our evaluation, the maximum CoT length is capped at 32,768 tokens; any response exceeding this limit is considered incorrect. Result discussions are addressed in \cref{ssec:results}.}
\resizebox{\textwidth}{!}{
\begin{tabular}{l|cccccccccccccccc}
\toprule
& \multicolumn{3}{c}{\sl Explicit Comp.} && \multicolumn{5}{c}{\sl Implicit Comp.} && \multicolumn{4}{c}{\sl Knowledge \& Semantics} \\
\cmidrule{2-4} \cmidrule{6-10} \cmidrule{12-15} 
& Parity & Cycle & Subsum && GSM8K & MATH & AIME & MuSR & Zebra && CSQA & MMLU & QuALITY & GPQA && \bf Avg.\\
\midrule
Random & 50 & 50 & 10 && 20 & 20 & 20 & 20 & 16.7 && 20 & 25 & 25 & 20 && 24.7 \\
\midrule
Qwen3-8B (w/o CoT) & 49.6 & 45.4 & 0.6 && 28.0 & 26.9 & 3.3 & 58.4 & 43.8 && 76.8 & 70.6 & 87.0 & 35.0 && 43.8 \\
Qwen3-8B (w/ CoT) & 76.5 & 92.8 & 89.1 && 97.0 & \bf 98.3 & \bf 76.8 & \bf 65.6 & 87.9 && 79.2 & 85.3 & 91.5 & 55.5 && 83.0 \\
\rowcolor{teal!20} \it CoT Length & \it 10550 & \it 16215 & \it 23503 && \it 2848 & \it 9873 & \it 44022 & \it 7904 & \it 22714 && \it 3221 & \it 4658 & \it 3523 & \it 20348 && \it 14114.9 \\
\midrule
Qwen3-32B (w/o CoT) & 46.8 & 45.4 & 8.4 && 49.0 & 39.3 & 33.3 & 55.8 & 50.6 && 77.6 & 78.8 & 92.6 & 42.6 && 51.7 \\
Qwen3-32B (w/ CoT) & 80.9 & 96.6 & \bf 94.7 && \bf 97.9 & 97.7 & 76.7 & 65.5 & 90.2 && \bf 82.3 & \bf 91.4 & \bf 97.3 & \bf 63.8 && \bf 86.2 \\
\rowcolor{teal!20} \it CoT Length & \it 6390 & \it 11951 & \it 13758 && \it 1678 & \it 6865 & \it 39085 & \it 4835 & \it 21306 && \it 2090 & \it 2920 & \it 2089 & \it 14793 && \it 10646.6\\
\midrule
In-Domain LLM (7B) & \bf 98.3 & \bf 97.2 & 87.6 && 94.4 & 87.3 & 56.7 & 53.8 & \bf 94.9 && 78.4 & 78.5 & 89.9 & 36.4 && 79.4 \\
\rowcolor{teal!20} \it CoT Length & \it 1699 & \it 972 & \it 2718 && \it 966 & \it 1268 & \it 1521 & \it 1013 & \it 1479 && \it 749 & \it 1247 & \it 628 & \it 1513 && \it 1314.6 \\
\bottomrule
\end{tabular}}
\label{tab:perf}
% \vspace{-1ex}
\end{table*}

\section{Probing Results}
\label{app:probe-results}

\subsection{Collection of Full Results}
\label{app:full}

\begin{figure*}[!p]
\centering
\includegraphics[width=\textwidth]{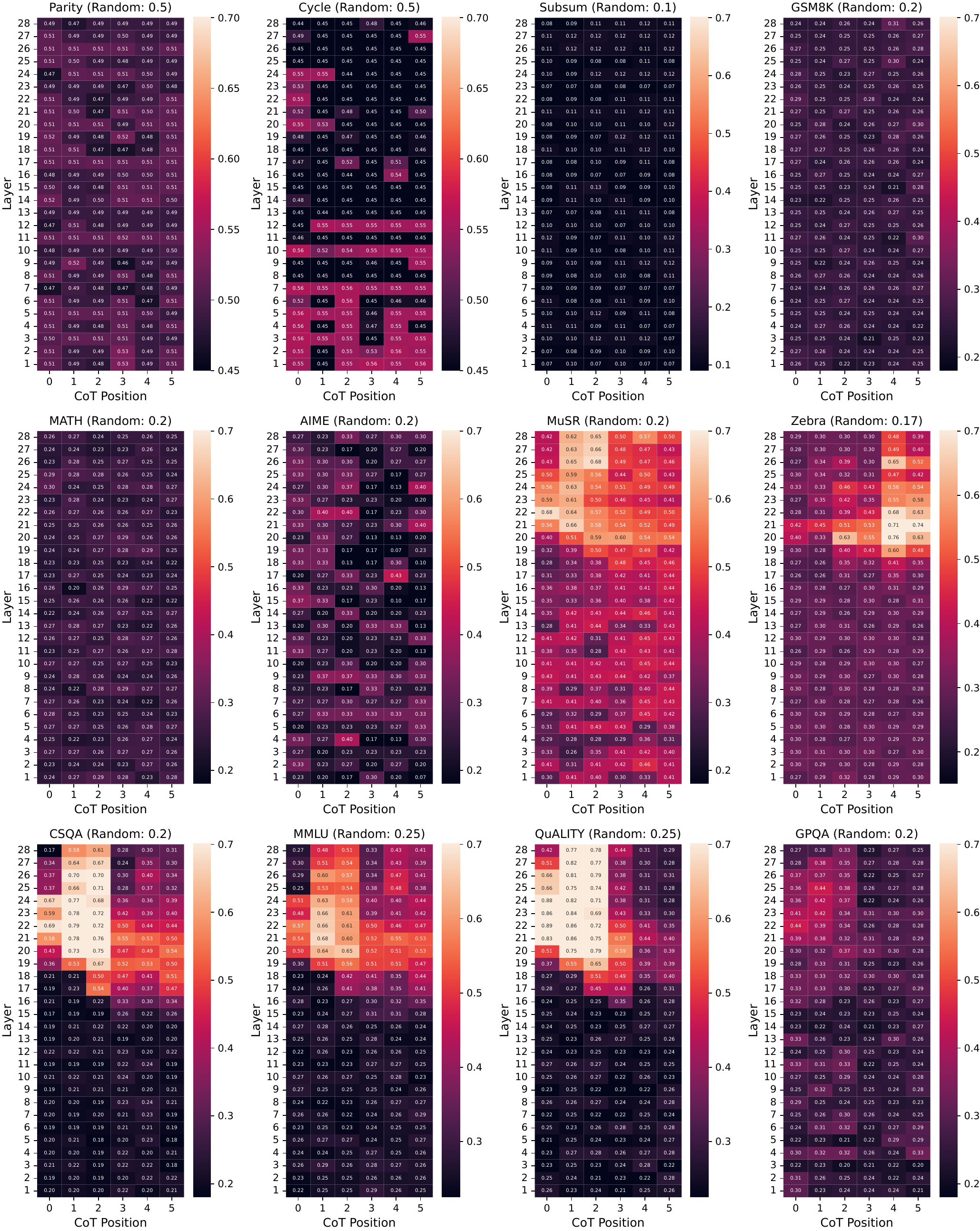}
\caption{Probing for final answers: averaged accuracy with In-Domain LLM for the first six tokens along CoT trajectories, measured across Transformers layers and tasks. Result discussions are addressed in \cref{ssec:result-answer}.}
\label{fig:probing-answer-by-layer-full}
% \vspace{-0.5em}
\end{figure*}

\begin{figure*}[!t]
\centering
\includegraphics[width=\textwidth]{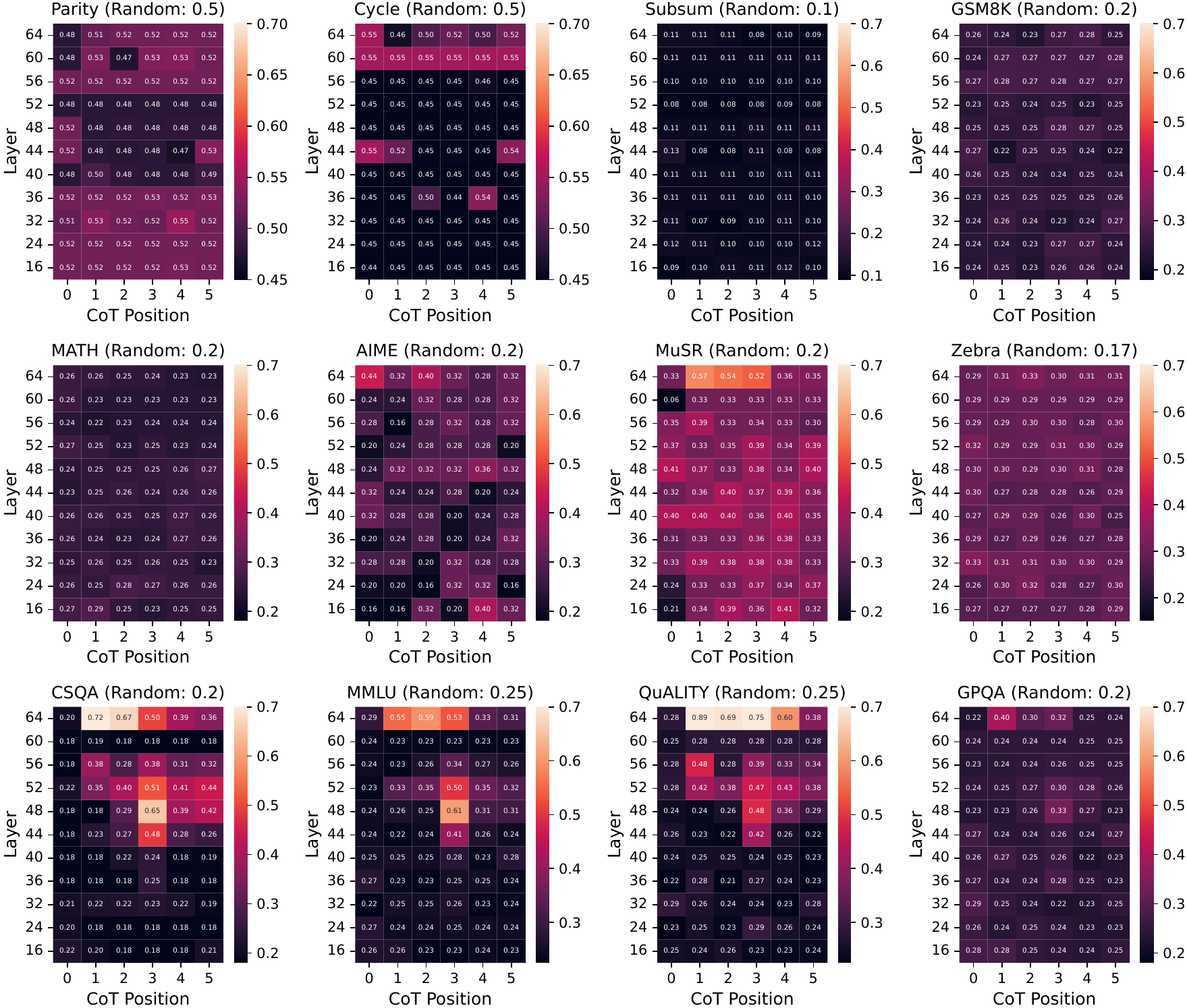}
\caption{Probing for final answers: averaged accuracy with Off-the-Shelf LLM (Qwen3-32B) for the first six tokens along CoT trajectories, measured across selected Transformers layers and tasks.}
\label{fig:probing-answer-by-layer-full-32b}
% \vspace{-0.5em}
\end{figure*}

\begin{figure*}[!th]
    \centering
    \begin{subfigure}{{0.48\textwidth}}
        \centering
        \includegraphics[width=\linewidth]{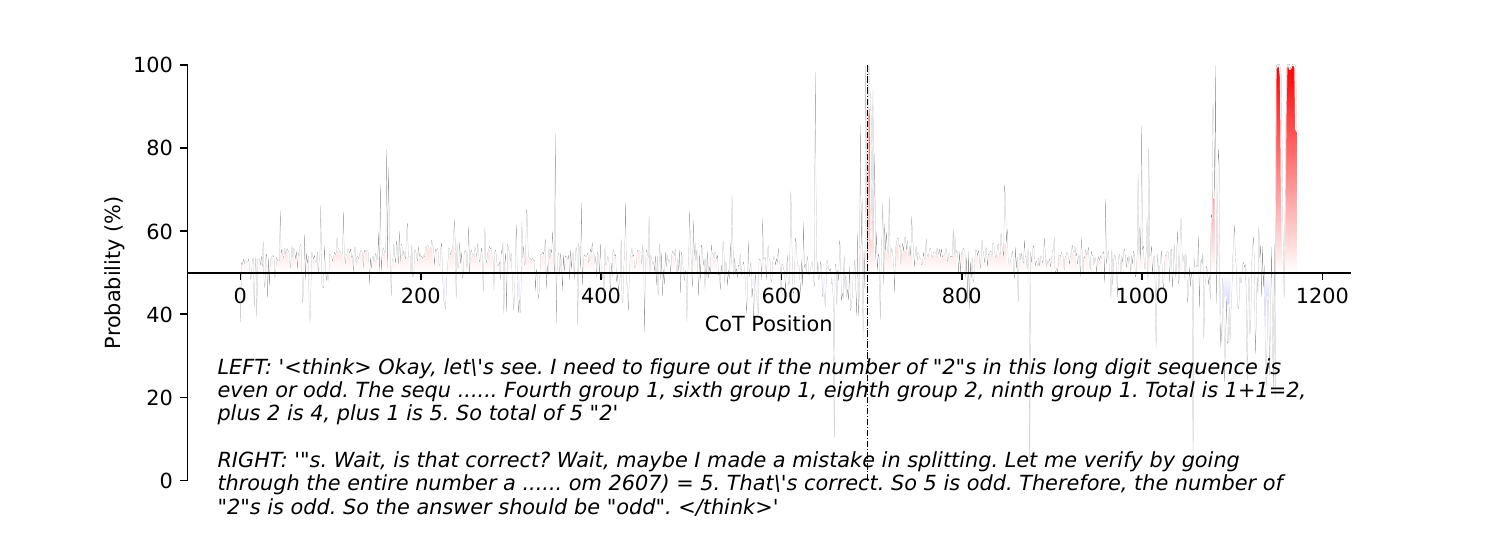}
        \caption{Parity example with Off-the-Shelf LLM.}
    \end{subfigure}
    \hspace{0.8em} % Horizontal space between images
    \begin{subfigure}{0.48\textwidth}
        \centering
        \includegraphics[width=\linewidth]{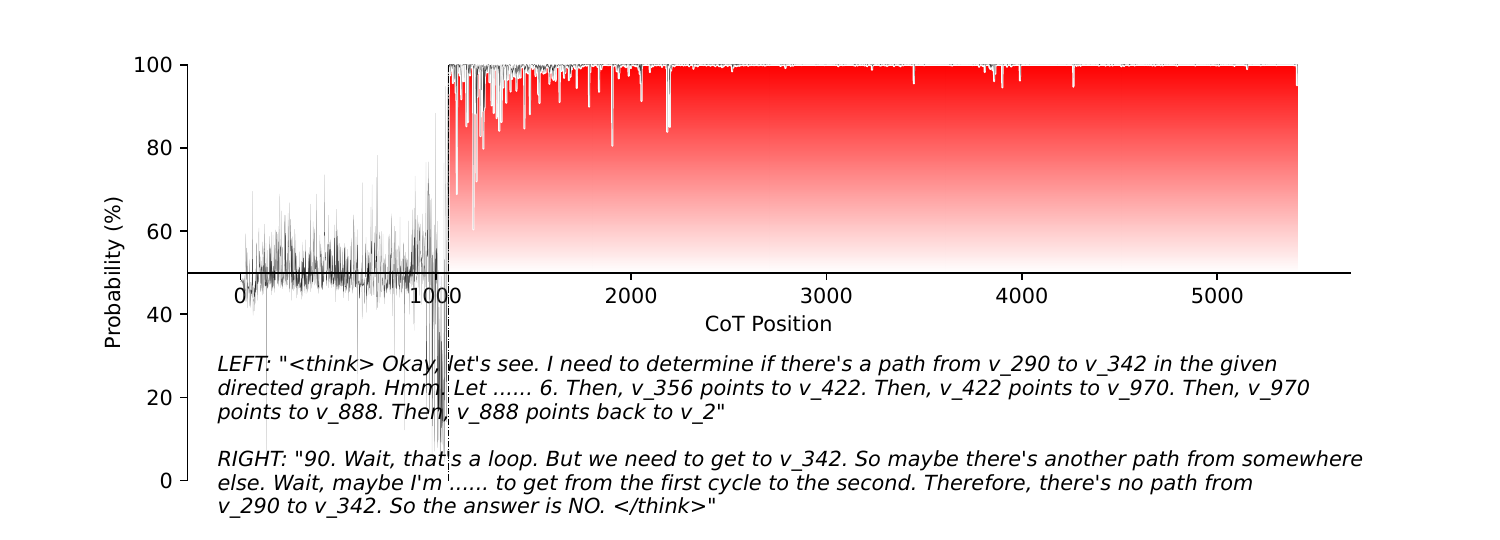}
        \caption{Cycle example with Off-the-Shelf LLM.}
    \end{subfigure}

    \caption{Examples of final-answer probing probabilities along CoT trajectories with Qwen3-32B (random guessing is at 50\%). The vertical dashed line indicates the position at which accuracy first spikes. ``\textsl{LEFT}'' and ``\textsl{RIGHT}'' at the bottom illustrate the reasoning details right before and after the accuracy spike, respectively. Examples with In-Domain LLM are addressed in \cref{fig:full}.}
    \label{fig:full-32b}
\end{figure*}

\begin{figure*}[!t]
\centering
\includegraphics[width=\textwidth]{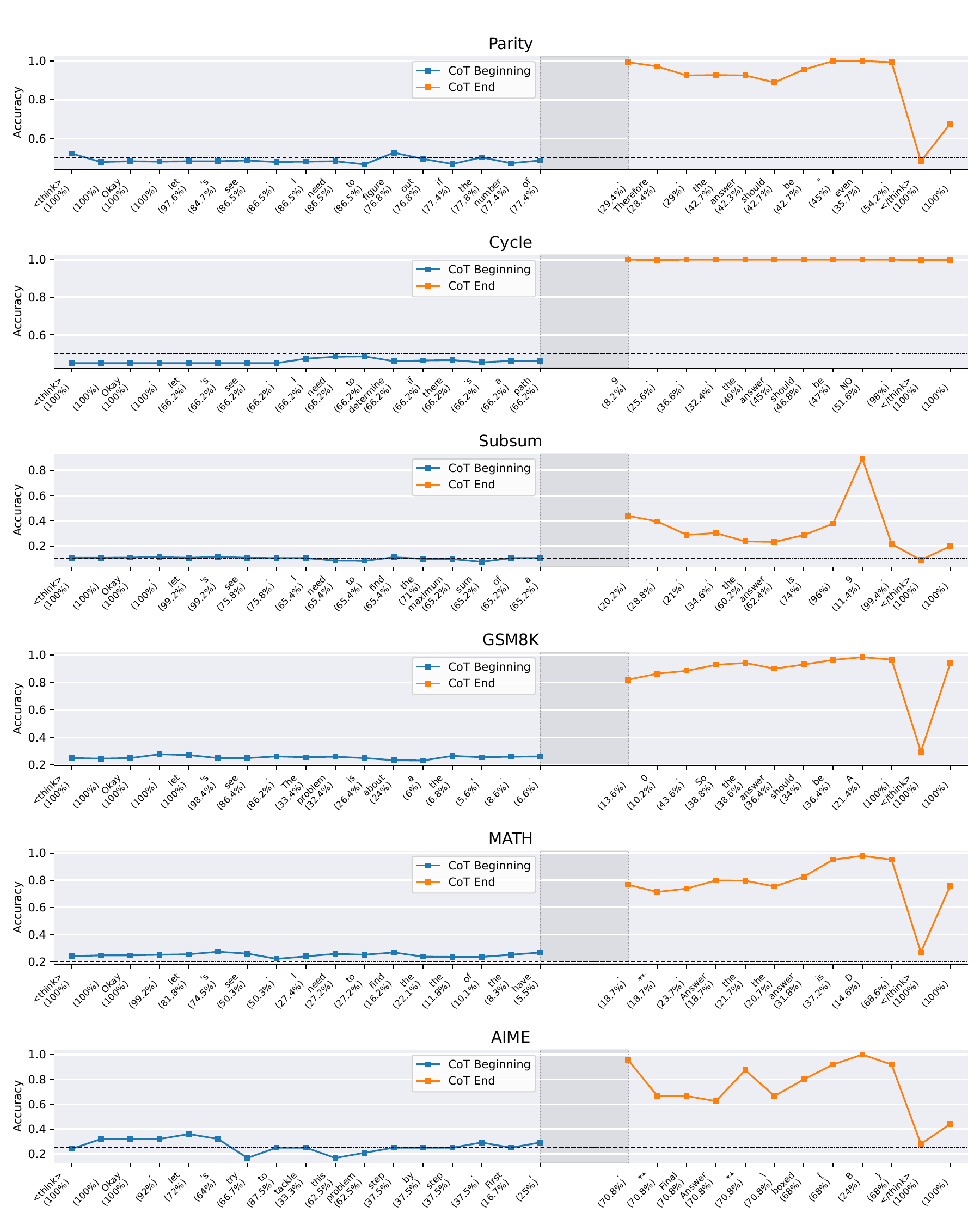}
\caption{Probing for final answers: averaged accuracy with Qwen3-32B along CoT positions. The most frequent token at each position is annotated with its occurrence frequency (the remaining 6 tasks are shown in \cref{fig:spike-32b-full-b}).}
\label{fig:spike-32b-full-a}
% \vspace{-0.5em}
\end{figure*}

\begin{figure*}[!t]
\centering
\includegraphics[width=\textwidth]{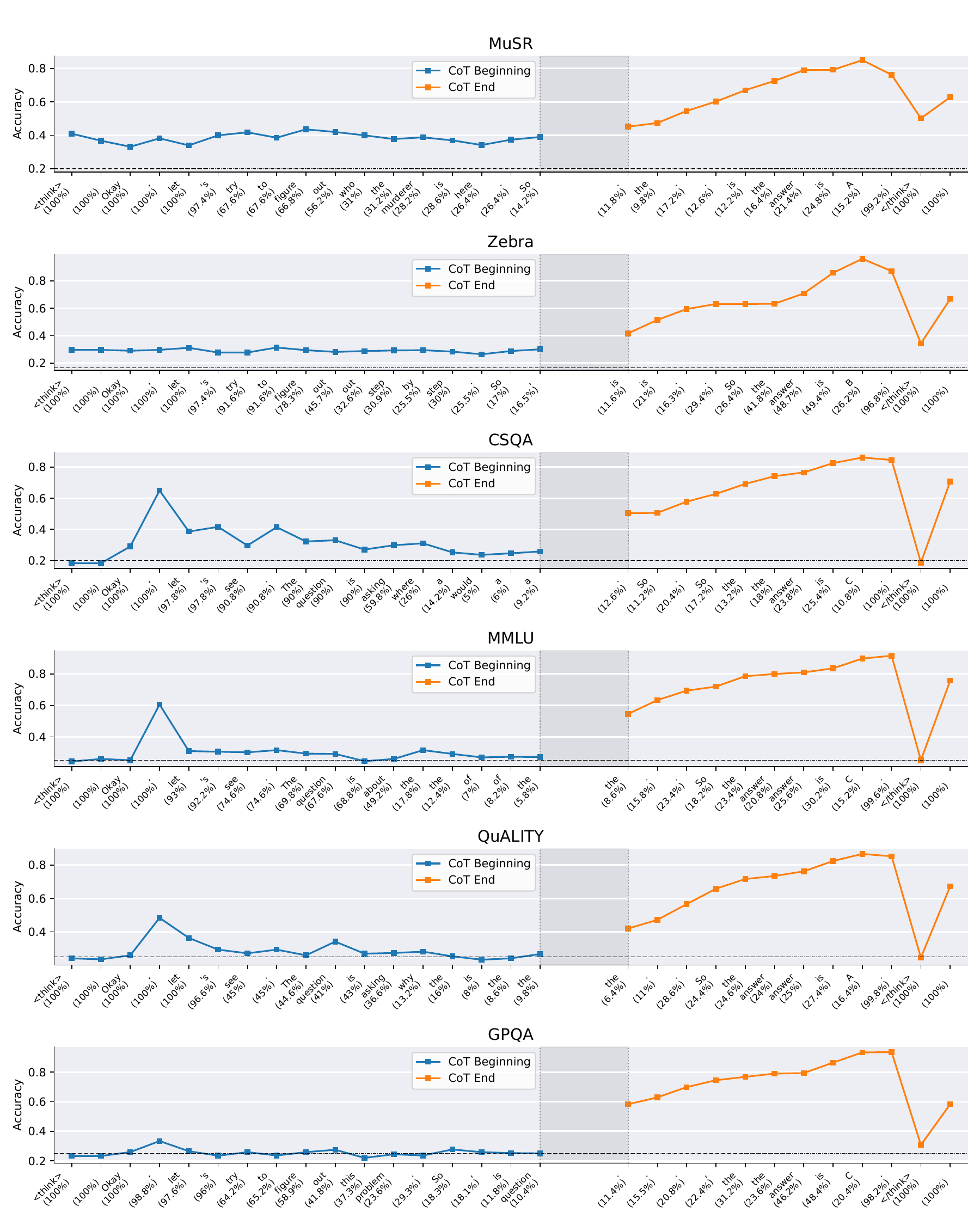}
\caption{Probing for final answers: averaged accuracy by Qwen3-32B along CoT positions. The most frequent token at each position is annotated with its occurrence frequency. Result discussions are addressed near \cref{fig:spike-32b}.}
\label{fig:spike-32b-full-b}
% \vspace{-0.5em}
\end{figure*}

\begin{figure*}[!t]
\centering
\includegraphics[width=\textwidth]{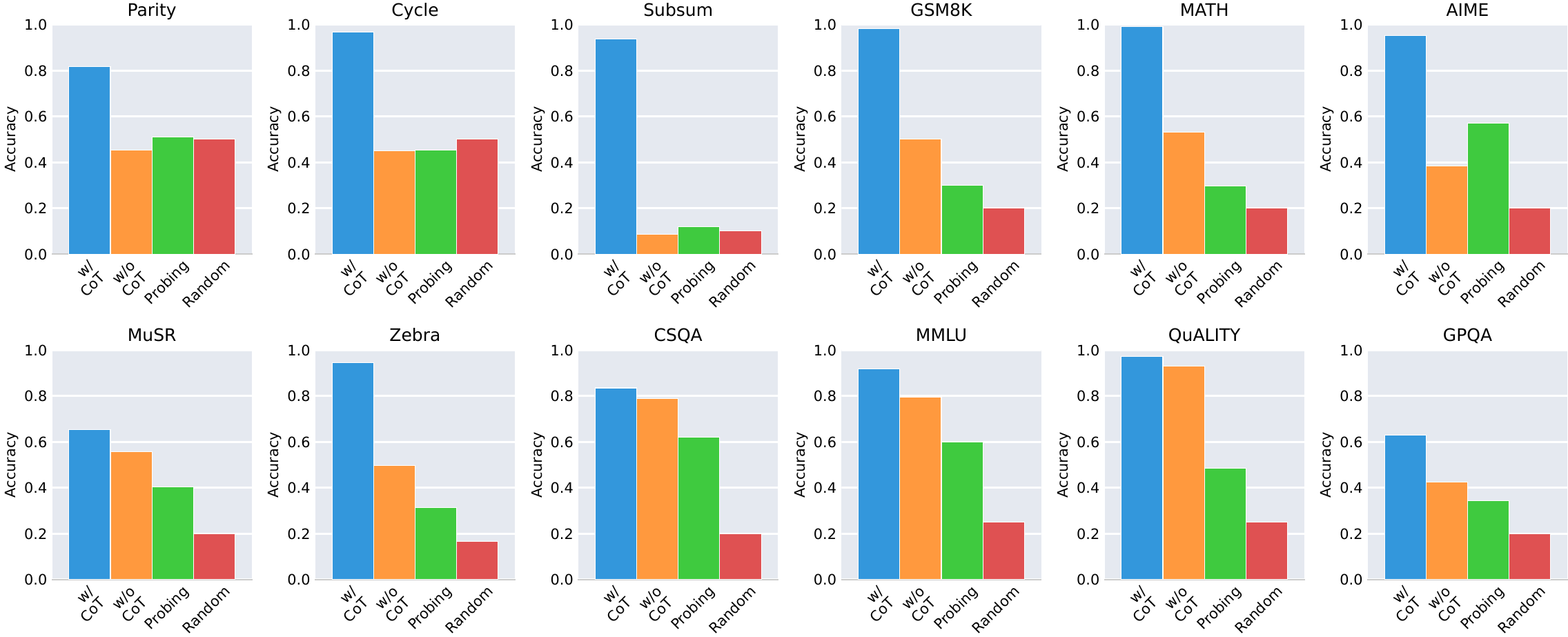}
\caption{Task accuracy comparison for Off-the-Shelf LLM (Qwen3-32B) under four settings: using thinking mode (\textbf{w/ CoT}); using non-thinking mode (\textbf{w/o CoT}); the best probing accuracy among initial CoT positions (\textbf{Probing}); the random-guess baseline (\textbf{Random}).}
\label{fig:acc-32b-full}
% \vspace{-0.5em}
\end{figure*}

\begin{figure*}[!t]
\centering
\includegraphics[width=\textwidth]{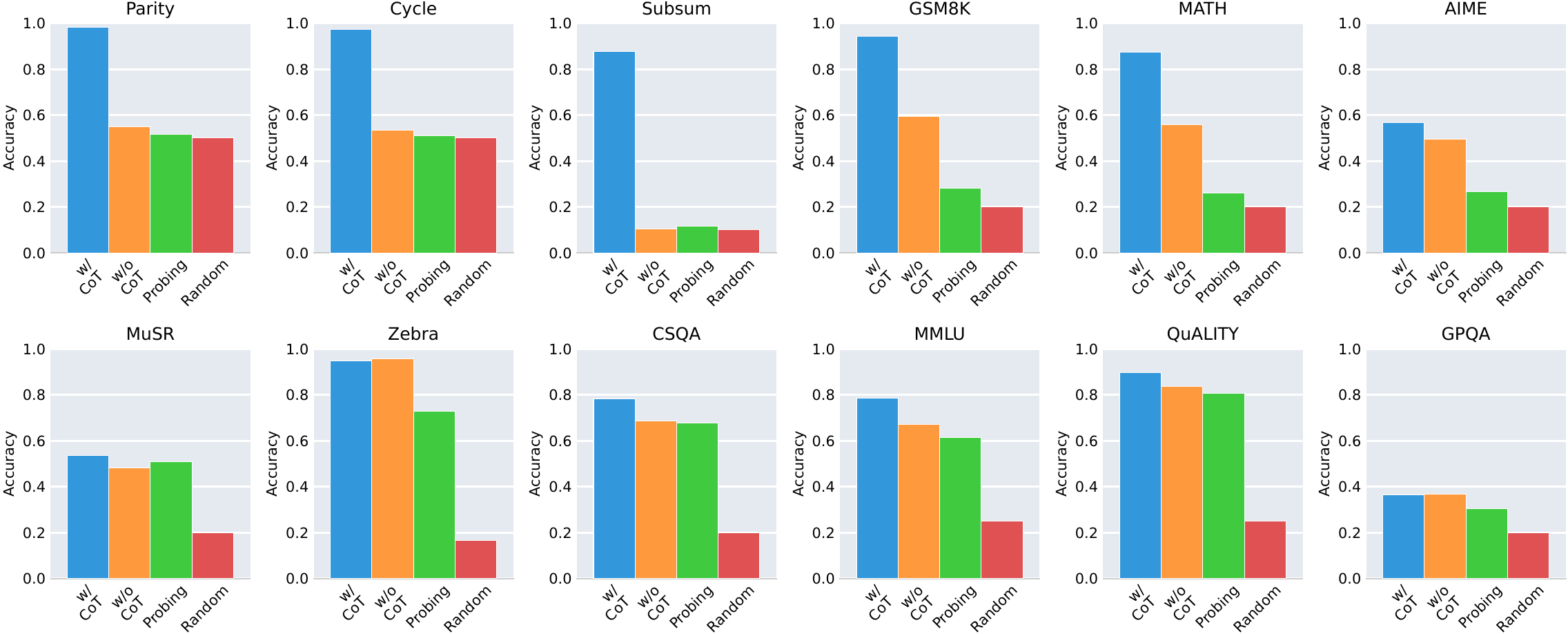}
\caption{Task accuracy comparison for In-Domain LLM under four settings: standard inference with learned CoT (\textbf{w/ CoT}); direct prediction by a separately trained model via naive supervised finetuning, without CoT learned (\textbf{w/o CoT}); the best probing accuracy among initial CoT positions (\textbf{Probing}); the random-guess baseline (\textbf{Random}). Result discussions are addressed near \cref{fig:acc-32b}.}
\label{fig:acc-full}
% \vspace{-0.5em}
\end{figure*}

\begin{figure*}[!t]
\centering
\includegraphics[width=\textwidth]{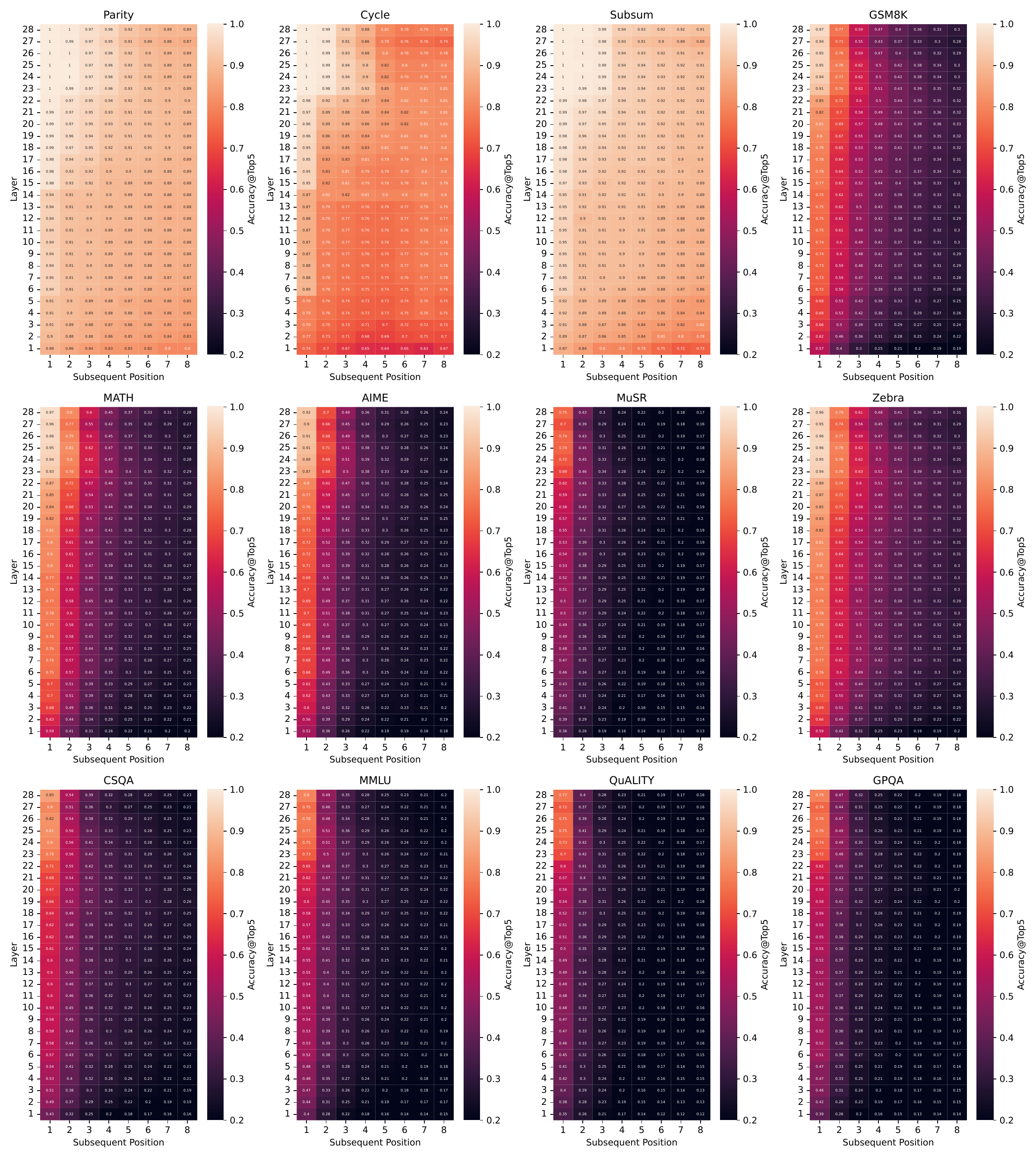}
\caption{Averaged Top-5 Accuracy of subsequent token prediction with In-Domain LLM, across Transformers layers and subsequent positions (up to the 8th following position). Result discussions are addressed near \cref{fig:hop-drop}.}
\label{fig:hop-full}
% \vspace{-0.5em}
\end{figure*}

\begin{figure*}[!t]
\centering
\includegraphics[width=\textwidth]{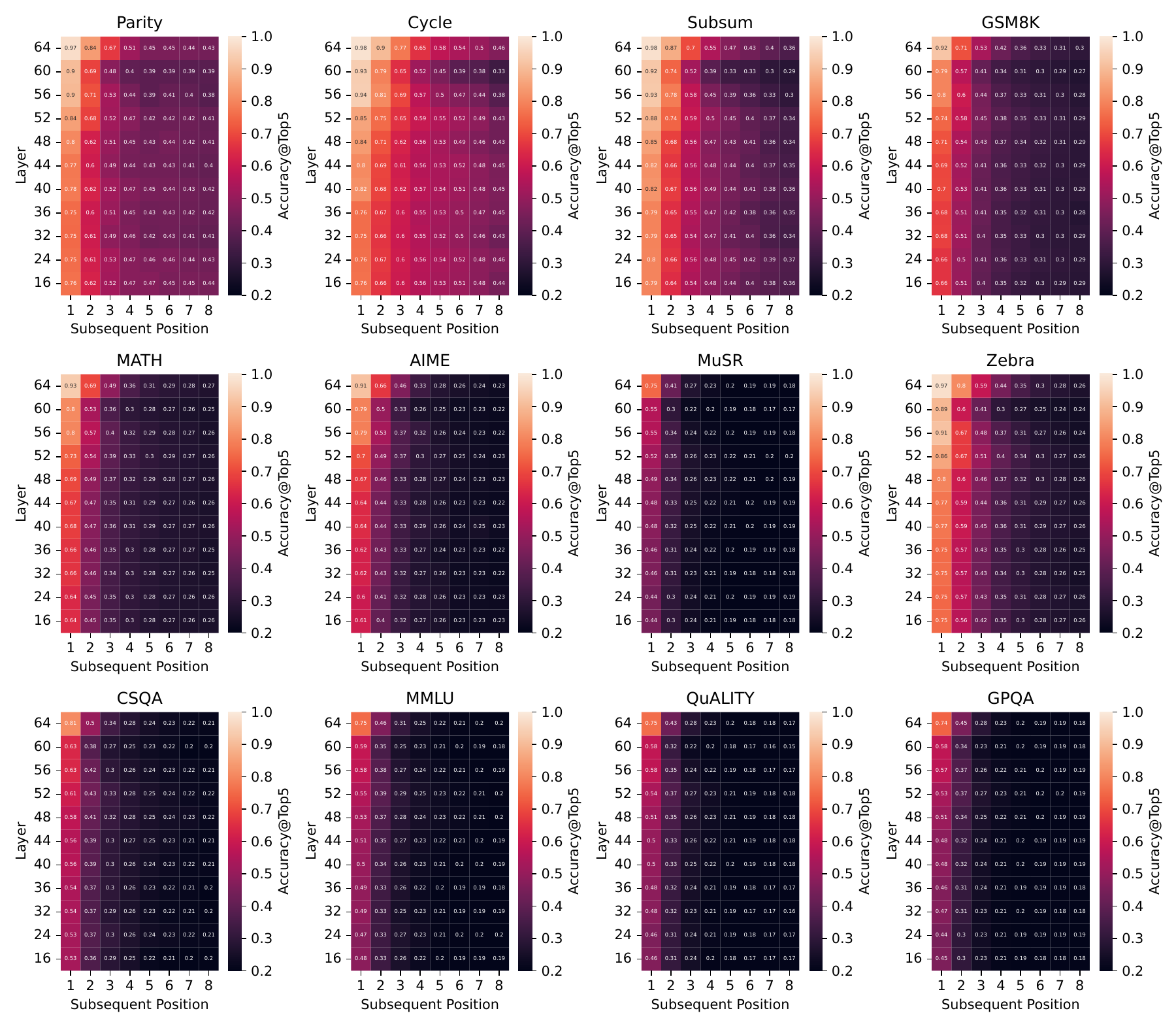}
\caption{Averaged Top-5 accuracy for subsequent token prediction with Off-the-Shelf LLM, across selected Transformers layers and subsequent positions (up to the 8th following position).}
\label{fig:hop-full-32b}
% \vspace{-0.5em}
\end{figure*}

\begin{figure*}[!t]
\centering
\includegraphics[width=0.95\textwidth]{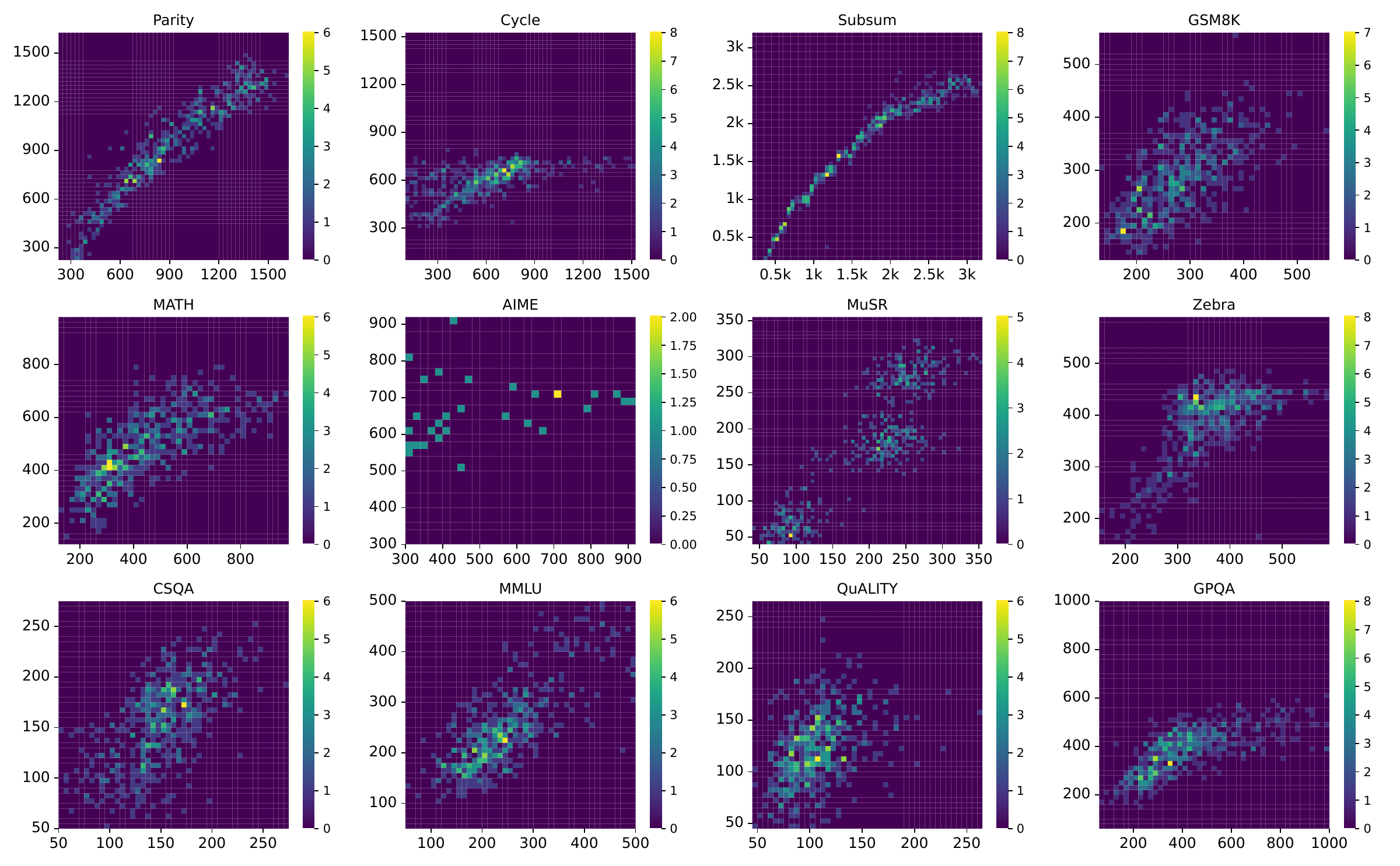}
\caption{Probing for reasoning length: heatmap of the predicted length (\emph{y}-axis) against the actual length (\emph{x}-axis) for In-Domain LLM.}
\label{fig:length-full}
% \vspace{-0.5em}
\end{figure*}

\begin{figure*}[!t]
\centering
\includegraphics[width=0.95\textwidth]{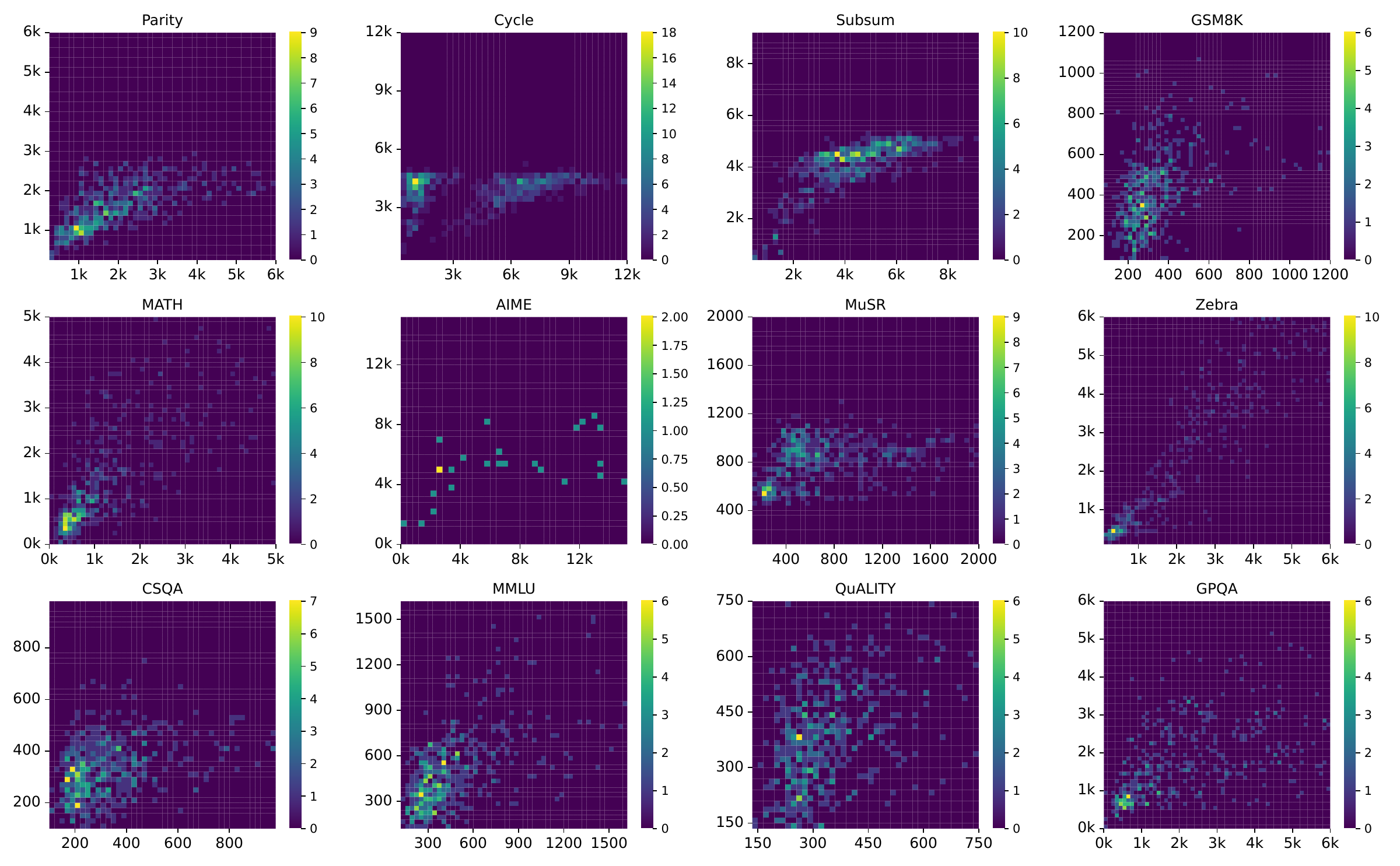}
\caption{Probing for reasoning length: heatmap of the predicted length (\emph{y}-axis) against the actual length (\emph{x}-axis) for Off-the-Shelf LLM. Result discussions are addressed near \cref{fig:length}.}
\label{fig:length-full-32b}
% \vspace{-0.5em}
\end{figure*}

\cref{ssec:results} reports the empirical results of our probing settings. Due to the limited space in the main pages, we present figures and tables for full results across layers and tasks, listed as below.
\begin{itemize}[noitemsep,nolistsep]
\item Probing for \textbf{Final Answers}
    \begin{itemize}[noitemsep,nolistsep]
    \item \cref{fig:probing-answer-by-layer-full}: the probing accuracy with In-Domain LLM at the beginning of CoT.
    \item \cref{fig:probing-answer-by-layer-full-32b}: the probing accuracy with Off-the-Shelf LLM at the beginning of CoT.
    \item \cref{fig:full-32b}: examples of probing accuracy dynamics along CoT trajectories with Off-the-Shelf LLM.
    \item \cref{fig:spike-32b-full-a}\&\ref{fig:spike-32b-full-b}: averaged probing accuracy along CoT trajectories with Off-the-Shelf LLM.
    \item \cref{fig:acc-32b-full}: task accuracy comparison for Off-the-Shelf LLM under settings that include thinking mode, non-thinking mode, early final-answer planning, random guess.
    \item  \cref{fig:acc-full}: comparison similar to \cref{fig:acc-32b-full} for In-Domain LLM.
    \end{itemize}
\item Probing for \textbf{Subsequent Tokens}
    \begin{itemize}[noitemsep,nolistsep]
    \item \cref{fig:hop-full}: Top-5 accuracy for subsequent token prediction with In-Domain LLM, up to the 8th following token.
    \item \cref{fig:hop-full-32b}: evaluation similar to \cref{fig:hop-full} with Off-the-Shelf LLM.
    \end{itemize}
\item Probing for \textbf{Global Steps}
    \begin{itemize}[noitemsep,nolistsep]
    \item \cref{fig:length-full}: heatmap of reasoning length probing with In-Domain LLM.
    \item \cref{fig:length-full-32b}: heatmap of reasoning length probing with Off-the-Shelf LLM.
    \end{itemize}
\end{itemize}

\subsection{More Myopic Planning Illustrations}
\label{app:more-myopic}

\begin{figure*}[!th]
    \centering
    \begin{subfigure}{{0.7\textwidth}}
        \centering
        \includegraphics[width=\linewidth]{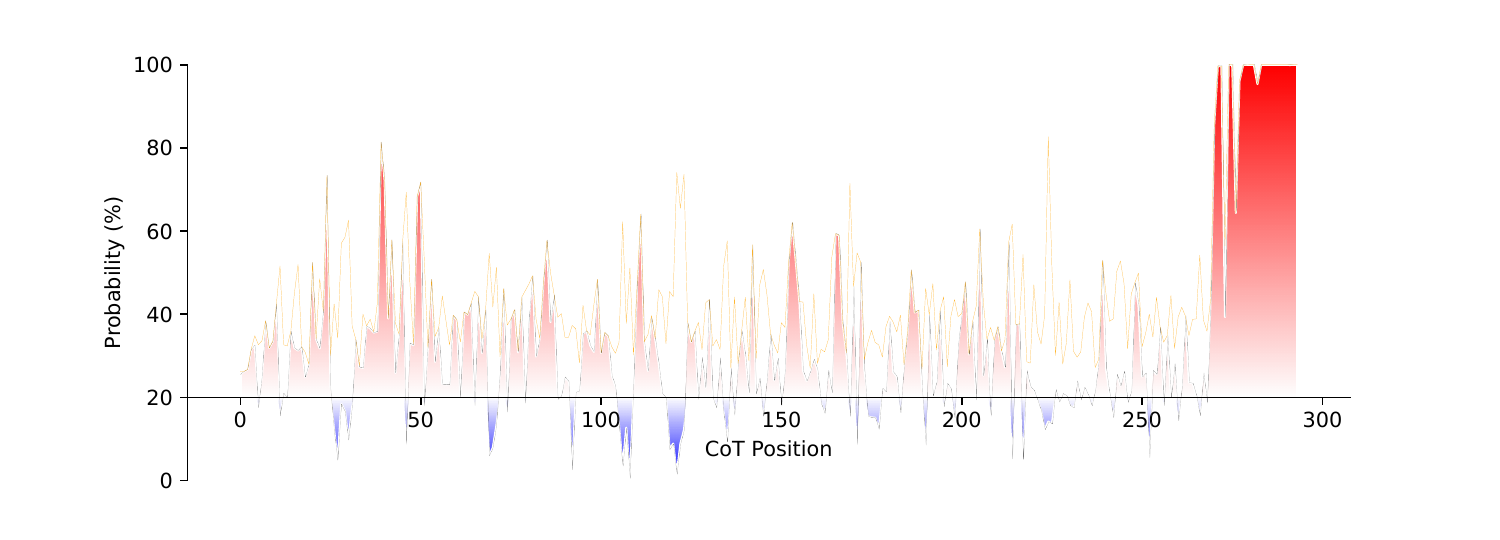}
        \caption{Trajectory example from GSM8K.}
    \end{subfigure}
    \begin{subfigure}{0.7\textwidth}
        \centering
        \includegraphics[width=\linewidth]{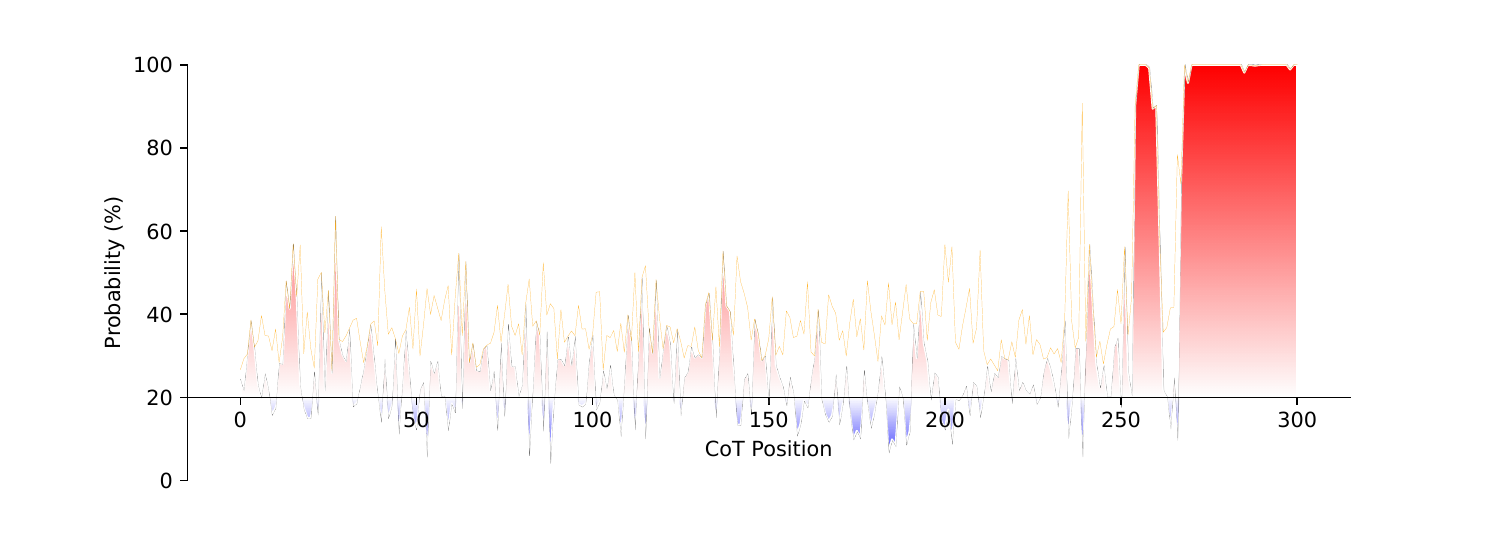}
        \caption{Trajectory example from MATH.}
    \end{subfigure}
    \begin{subfigure}{0.7\textwidth}
        \centering
        \includegraphics[width=\linewidth]{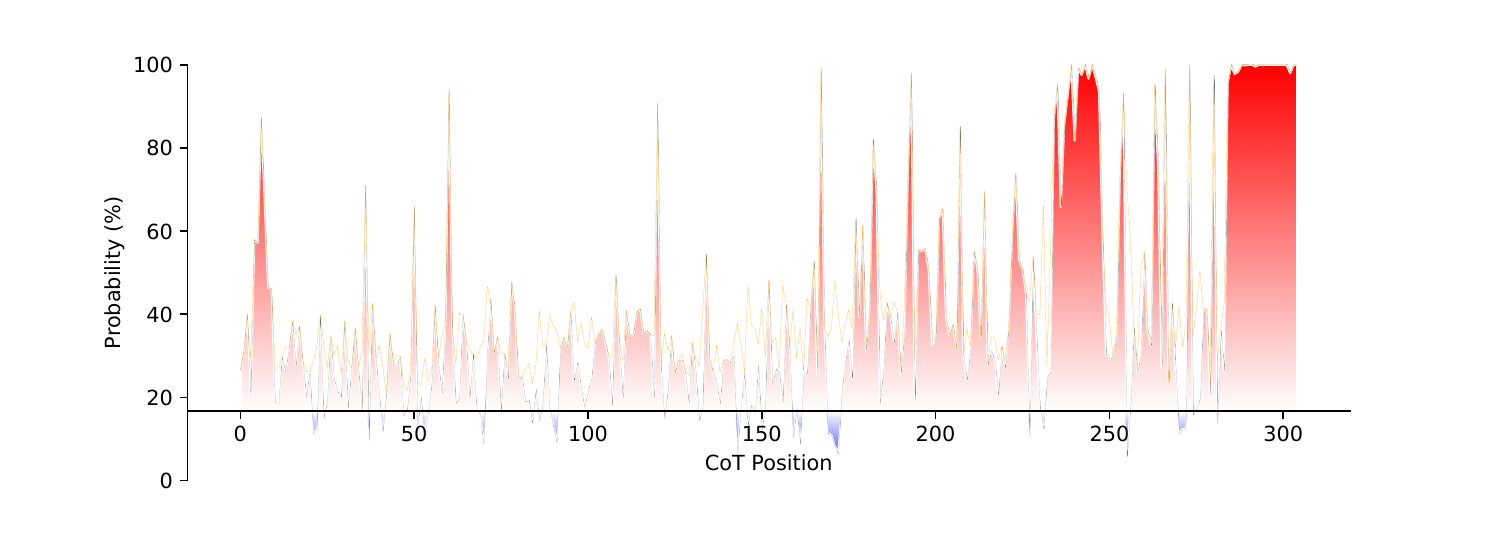}
        \caption{Trajectory example from Zebra.}
    \end{subfigure}
    \begin{subfigure}{0.7\textwidth}
        \centering
        \includegraphics[width=\linewidth]{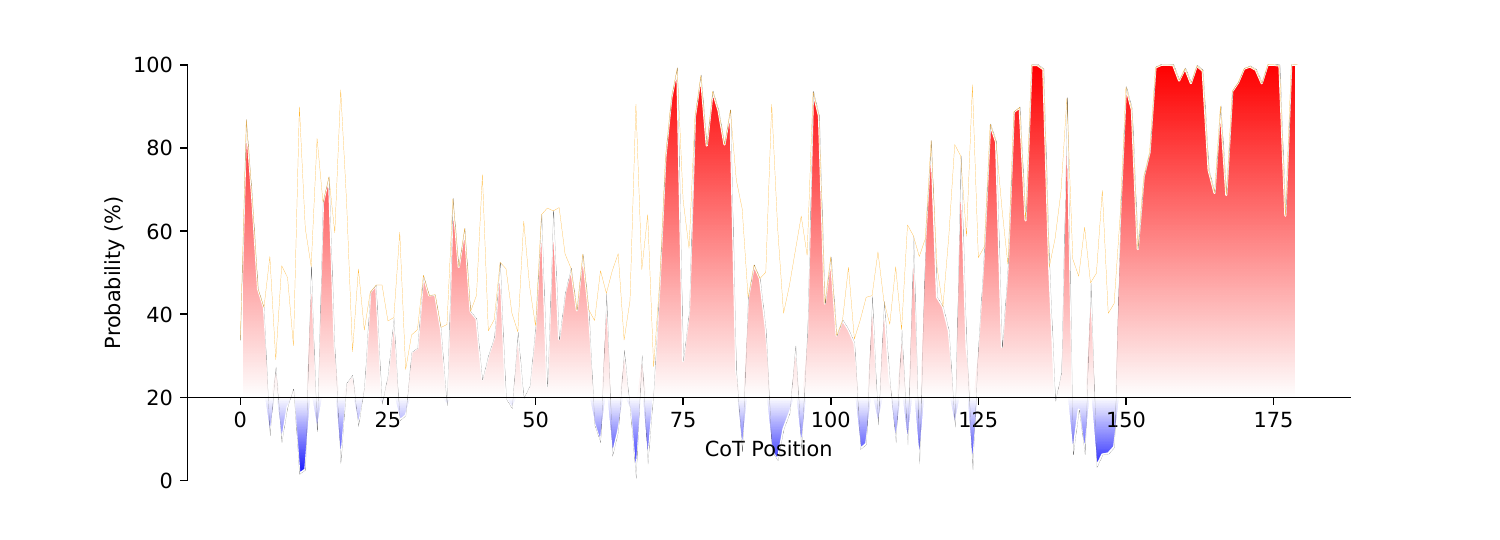}
        \caption{Trajectory example from CSQA.}
    \end{subfigure}

    \caption{Examples of final-answer probing probabilities along CoT trajectories with In-Domain LLM. The yellow line denotes the maximum probability over the answer space at each step. For tasks beyond explicit compositional reasoning, accuracy spikes also occur sparsely. Especially for mathematical and logical reasoning, the final answer emerges towards the end of the reasoning, indicating a myopic planning horizon. More discussions are addressed near \cref{fig:full}.}
    \label{fig:full-id-more}
\end{figure*}

\cref{fig:full-id-more} illustrates the final-answer planning dynamics on tasks beyond explicit compositional reasoning. They also exhibit a \emph{myopic} planning horizon, with high-confidence probing positions appearing sparsely. Similar to explicit compositional tasks, such positions tend to emerge near CoT completion in math and logical reasoning tasks as well.

\section{Leveraging CoT Dynamics}
\label{app:utility}

\begin{figure}[!t]
\centering
\includegraphics[width=0.86\columnwidth]{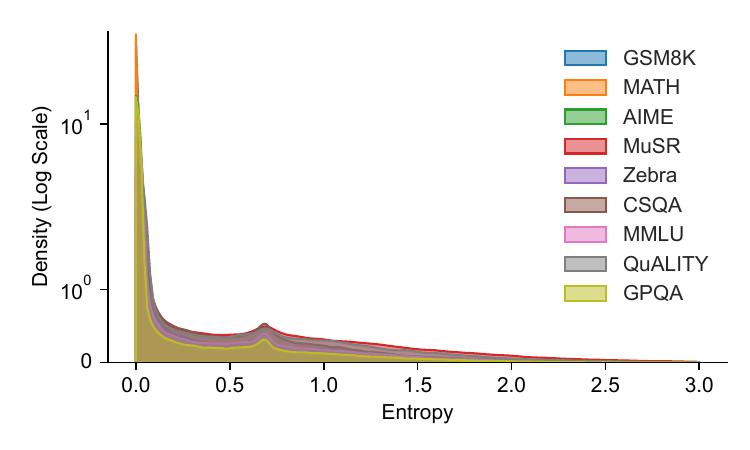}
\caption{Density distribution of LM entropy for next-token prediction steps per task. As depicted, most tokens exhibit low entropy, reflecting confident local transitions (\cref{ssec:uncertainty}).}
\label{fig:ent-density}
% \vspace{-0.5em}
\end{figure}

\paragraph{Results}
\begin{itemize}[noitemsep,nolistsep,leftmargin=1em]
\item \cref{fig:ent-density} presents the density distribution of LM entropy for next-token prediction.
\item \cref{tab:id-uncertainty-full} and \ref{tab:qwen-uncertainty-full} show the full evaluation results for uncertainty estimation, using our top-$k$ pivot selection strategy described in \cref{ssec:uncertainty}.
\item \cref{tab:reduction-full} presents the full evaluation results of CoT bypass described in \cref{ssec:necessity}.
\end{itemize}

\section{Related Works and Discussions}
\label{app:related}

There can be many implications brought by the \emph{myopic} planning horizon uncovered in this work. Since the model cannot plan the end from the beginning, it must initiate the dynamic reasoning as an necessary act of state searching and exploration. Therefore, explicit planning within CoT can be important, as empirically validated by recent works \cite{wang-etal-2023-plan,wang2025planning,wang2026latent}.
The exploitation of latent signals from CoT dynamics can be significant to various LLM characteristics and applications. For instance, recent works have investigated to utilize latent signals to compress CoT \cite{zhang2025reasoningmeetslaws,li2026making}, steer model behavior \cite{sheng2026on,li2026efficient}, perform early stop of CoT \cite{afzal-etal-2025-knowing,singh2026llmsencodefunctionalimportance,li2026making}, and improve model training \cite{huang-etal-2025-enhancing}. Thus, we believe understanding the underlying CoT dynamics is significant to address in research.

To understand the internal states of LLMs, prior works have conducted probing studies on Transformers' hidden states to address truthful responses \cite{azaria-mitchell-2023-internal, liu-etal-2024-enhancing-language,gottesman-geva-2024-estimating,chen2025deephiddencognitionfacilitates}, assess world knowledge representation \cite{patel2022mapping,li2023emergent} or the global planning prior to CoT generation \cite{dong2025emergent}. In terms of probing methods, in addition to training probers on LLM hidden states as in \cite{dong2025emergent}, other methods have been proposed as well, such as Patchscopes \cite{ghandeharioun2024patchscopes}. In this work, we opt in probers for the flexibility predicting various targets, as we have three probing dimensions described in \cref{sec:horizon}.

Apart from probing, prior works have studied CoT dynamics from different aspects. \citet{wang2025beyond} finds that only about 20\% tokens are of high entropy. \citet{bigelow2025forking} proposes a sampling-based method for pivot token identification. \citet{ton2025understanding} proposes a methodology to quantify information gain at each CoT step. \cite{shao2025continuous} proposes CoT 
Several works have also identified that CoT could bring negative impact in certain scenarios \cite{sprague2025to,liu2025mind}.

In additional to analyses related to CoT dynamics, studies on Transformers' learnability and expressibility have highlighted the functional necessity of CoT on certain problems. 
Several works have focused on the theoretical limitation of Transformers, where it fails to perform soft multi-step reasoning within one step \cite{bhattamishra-etal-2023-simplicity,merrill-sabharwal-2023-parallelism,li2024chain}, and only intermediate CoT steps can derive length generalization \cite{anil2022exploring,xiao2025generalizing} and compositional reasoning \cite{wies2023subtask,abbe2024how,zubic2025limits}, making CoT indispensable especially for compositional problems. Our experiments in this work generally align with those findings.

To the best of our knowledge, this work is the first to focus explicitly on the latent planning horizon and its effective utilization, offering a unified perspective on prior work from complementary angles. We also call for attention on the identification and exploitation of more such hidden yet valuable latent signals to further deepen our understanding of CoT synergy.

\begin{table*}[tbp!]
\centering
\caption{Uncertainty estimation results (AUROC) with In-Domain LLM, using latent signals from final-answer probing via \probe. Note that we exclude auto-generated tasks (Parity, Cycle, Subsum), as their uncertainty is already artificially correlated with input length. Result discussions are provided near \cref{tab:id-uncertainty}.}
\resizebox{0.94\textwidth}{!}{
\begin{tabular}{l|ccccccccc|c}
\toprule
& GSM8K & MATH & AIME & MuSR & Zebra & CSQA & MMLU & QuALITY & GPQA & \bf Avg.\\
\midrule
Perplexity & 0.70 & 0.64 & 0.49 & 0.50 & 0.58 & 0.57 & 0.53 & 0.64 & 0.50 & 0.57 \\
Entropy & 0.72 & 0.65 & 0.49 & 0.51 & 0.60 & 0.58 & 0.52 & 0.63 & 0.50 & 0.58 \\
Self-Certainty & 0.76 & 0.71 & 0.51 & \bf 0.52 & 0.67 & 0.58 & 0.53 & 0.64 & 0.51 & 0.60 \\
\midrule
\probe (Top-5) & \bf 0.87 & \bf 0.81 & \bf 0.65 & 0.49 & \bf 0.77 & \bf 0.64 & \bf 0.73 & \bf 0.72 & \bf 0.56 & \bf 0.69 \\
\probe (Top-10) & 0.81 & \bf 0.81 & 0.63 & 0.49 & 0.75 & 0.63 & 0.72 & 0.70 & 0.56 & 0.68 \\
\probe (Top-20) & 0.82 & 0.79 & 0.56 & 0.49 & 0.67 & 0.52 & 0.65 & 0.66 & 0.51 & 0.63 \\
\probe (Top-50) & 0.78 & 0.73 & 0.49 & 0.50 & 0.69 & 0.40 & 0.56 & 0.56 & 0.47 & 0.64 \\
\bottomrule
\end{tabular}}
\label{tab:id-uncertainty-full}
% \vspace{-1ex}
\end{table*}

\begin{table*}[tbp!]
\centering
\caption{Uncertainty estimation results (AUROC) with Off-the-Shelf LLMs, using our top-$k$ strategy upon each general metric. Note that we exclude auto-generated tasks (Parity, Cycle, Subsum), as their uncertainty is already artificially correlated with input length. We also exclude AIME, as there are not enough negative instances from both models. Our top-$k$ strategy brings no negative impact, and particularly drives consistent improvement with Qwen3-32B. Result discussions are provided near \cref{tab:qwen-uncertainty}.}
\resizebox{0.95\textwidth}{!}{
\begin{tabular}{ll|cccccccc|c}
\toprule
&& GSM8K & MATH & MuSR & Zebra & CSQA & MMLU & QuALITY & GPQA & \bf Avg.\\
\midrule
\multirow{12}{*}{Qwen3-8B} & Perplexity & 0.85 & 0.84 & 0.54 & 0.73 & \bf 0.77 & \bf 0.84 & 0.85 & \bf 0.72 & 0.77 \\
& \quad + 10 Pivots & 0.82 & 0.82 & 0.52 & 0.79 & 0.73 & 0.79 & 0.81 & 0.61 & 0.74 \\
& \quad + 100 Pivots & 0.87 & \bf 0.87 & 0.58 & 0.89 & 0.76 & 0.82 & 0.85 & 0.67 & 0.79 \\
& \quad + 1000 Pivots & \bf 0.88 & \bf 0.87 & \bf 0.59 & \bf 0.94 & \bf 0.77 & 0.82 & \bf 0.86 & 0.70 & \bf 0.80 \\
\cmidrule{2-11}
& Entropy & 0.85 & 0.83 & 0.55 & 0.76 & \bf 0.76 & 0.85 & 0.85 & \bf 0.73 & 0.77 \\
& \quad + 10 Pivots & 0.87 & 0.84 & 0.54 & 0.93 & \bf 0.76 & 0.82 & 0.84 & 0.70 & 0.79 \\
& \quad + 100 Pivots & \bf 0.88 & 0.85 & 0.57 & \bf 0.94 & \bf 0.76 & 0.82 & \bf 0.86 & 0.72 & \bf 0.80 \\
& \quad + 1000 Pivots & \bf 0.88 & \bf 0.86 & \bf 0.59 & \bf 0.94 & \bf 0.76 & 0.82 & \bf 0.86 & 0.71 & \bf 0.80 \\
\cmidrule{2-11}
& Self-Certainty & 0.87 & \bf 0.92 & 0.58 & \bf 0.96 & \bf 0.77 & \bf 0.84 & 0.86 & 0.72 & \bf 0.82 \\
& \quad + 10 Pivots & 0.86 & 0.91 & 0.55 & 0.95 & \bf 0.77 & 0.83 & 0.85 & \bf 0.73 & 0.81 \\
& \quad + 100 Pivots & \bf 0.88 & 0.90 & 0.58 & \bf 0.96 & 0.76 & 0.83 & \bf 0.87 & \bf 0.73 & \bf 0.82 \\
& \quad + 1000 Pivots & \bf 0.88 & 0.91 & \bf 0.59 & \bf 0.96 & 0.76 & 0.83 & \bf 0.87 & 0.72 & \bf 0.82 \\
\midrule
\multirow{12}{*}{Qwen3-32B} & Perplexity & 0.71 & 0.93 & 0.48 & 0.74 & 0.68 & 0.76 & 0.78 & 0.69 & 0.72 \\
& \quad + 10 Pivots & 0.74 & 0.79 & 0.48 & 0.75 & 0.69 & 0.76 & 0.79 & 0.71 & 0.71 \\
& \quad + 100 Pivots & \bf 0.81 & 0.92 & \bf 0.50 & 0.90 & \bf 0.74 & \bf 0.81 & \bf 0.82 & 0.73 & \bf 0.78 \\
& \quad + 1000 Pivots & 0.72 & \bf 0.95 & 0.49 & \bf 0.91 & 0.71 & 0.80 & 0.80 & \bf 0.74 & 0.76 \\
\cmidrule{2-11}
& Entropy & 0.71 & \bf 0.92 & 0.47 & 0.77 & 0.68 & 0.77 & 0.77 & 0.68 & 0.72 \\
& \quad + 10 Pivots & 0.78 & 0.69 & 0.48 & 0.87 & 0.71 & 0.79 & 0.80 & 0.73 & 0.73 \\
& \quad + 100 Pivots & \bf 0.81 & 0.70 & \bf 0.49 & 0.90 & \bf 0.74 & \bf 0.83 & \bf 0.82 & \bf 0.74 & \bf 0.75 \\
& \quad + 1000 Pivots & 0.71 & 0.87 & 0.48 & \bf 0.91 & 0.71 & 0.81 & 0.79 & 0.73 & \bf 0.75 \\
\cmidrule{2-11}
& Self-Certainty & 0.45 & 0.82 & 0.47 & 0.92 & 0.51 & 0.67 & 0.64 & 0.68 & 0.65 \\
& \quad + 10 Pivots & 0.53 & 0.89 & 0.47 & 0.91 & 0.57 & 0.71 & 0.67 & 0.69 & 0.68 \\
& \quad + 100 Pivots & \bf 0.55 & 0.90 & 0.47 & \bf 0.93 & \bf 0.59 & \bf 0.74 & \bf 0.70 & 0.70 & \bf 0.70 \\
& \quad + 1000 Pivots & 0.52 & \bf 0.91 & \bf 0.48 & \bf 0.93 & 0.54 & \bf 0.74 & 0.68 & \bf 0.73 & 0.69 \\
\bottomrule
\end{tabular}}
\label{tab:qwen-uncertainty-full}
% \vspace{-1ex}
\end{table*}

\begin{table*}[tbp!]
\centering
\caption{Evaluation results for our CoT bypass described in \cref{ssec:necessity}, varying thresholds of normalized entropy obtained from final-answer probing at early CoT positions. The CoT bypass ratio for each task is reported. \textbf{Avg.} denotes the average bypass ratio, and \textbf{Perf.} indicates the average change in task performance measured by absolute accuracy. Result discussions are addressed near \cref{tab:reduction}.}
\resizebox{\textwidth}{!}{
\begin{tabular}{l|cccccccccccc|cc}
\toprule
& Parity & Cycle & Subsum & GSM8K & MATH & AIME & MuSR & Zebra & CSQA & MMLU & QuALITY & GPQA & \bf Avg. & \bf Perf.\\
\midrule
\multicolumn{15}{c}{In-Domain LLM}\\
\cmidrule{1-15}
Th=0.02 & 0\% & 0\% & 0\% & 0\% & 0\% & 0\% & 1.2\% & 3.6\% & 16.6\% & 11.2\% & 26.2\% & 2\% & 5.07\% & -0.15 \\
Th=0.05 & 0\% & 0\% & 0.2\% & 0\% & 0\% & 0\% & 2.8\% & 11.2\% & 27.4\% & 22\% & 41.4\% & 4.4\% & 9.12\% & -0.32 \\
Th=0.1 & 0\% & 0\% & 0.2\% & 0\% & 0\% & 0\% & 8.2\% & 18.4\% & 40.2\% & 30.4\% & 55.6\% & 7\% & 13.33\% & -0.47 \\
Th=0.2 & 0\% & 0\% & 0.2\% & 0\% & 0\% & 0\% & 27\% & 34.6\% & 65\% & 45\% & 75.8\% & 12\% & 21.63\% & -1.42 \\
\midrule
\multicolumn{15}{c}{Off-the-Shelf LLM (Qwen3-32B)}\\
\cmidrule{1-15}
Th=0.02 & 0\% & 0\% & 0\% & 0\% & 0\% & 0\% & 0\% & 0\% & 6.8\% & 5\% & 0.6\% & 0.2\% & 1.05\% & -0.03 \\
Th=0.05 & 0\% & 0\% & 0\% & 0\% & 0\% & 0\% & 0.2\% & 0\% & 10.4\% & 8.4\% & 2\% & 0.4\% & 1.78\% & -0.03 \\
Th=0.1 & 0\% & 0\% & 0\% & 0\% & 0\% & 0\% & 0.2\% & 0\% & 16.2\% & 12.4\% & 3.2\% & 1.2\% & 2.77\% & -0.03 \\
Th=0.2 & 0\% & 0\% & 0\% & 0\% & 0\% & 0\% & 14.4\% & 0\% & 28.8\% & 20.2\% & 7.6\% & 3.2\% & 6.19\% & -0.37 \\
\bottomrule
\end{tabular}}
\label{tab:reduction-full}
% \vspace{-1ex}
\end{table*}

%===== Figures for Full Results
\clearpage

\end{document}